%% file: main-arxiv.tex
\documentclass[12pt]{article}
\usepackage[margin=1in]{geometry}

\usepackage[colorlinks,allcolors=darkgray,pagebackref]{hyperref}
\renewcommand*{\backrefalt}[4]{%
\ifcase #1 %
No citations.%
\or
Cited on page #2.%
\else
Cited on pages #2.%
\fi
}

\usepackage[tiny,notocpart*]{titlesec}
\titleformat{\part}[display]
  {\normalfont\itshape\filcenter}
  {}
  {0em}
  {\titlerule\vspace{6pt}}[\vspace{6pt}\titlerule]

\usepackage{multirow}
\usepackage{algorithmic}
\usepackage{algorithm}
\usepackage{url}
\usepackage{booktabs}
\usepackage{enumitem}
\usepackage{makecell}
\usepackage{amsthm}
\usepackage[normalem]{ulem}
\usepackage{color}
\usepackage{xcolor}
\usepackage{tocloft}
\definecolor{darkgray}{gray}{0.35}
\definecolor{lightgray}{gray}{0.85}
\usepackage{colortbl}
\usepackage[labelfont={color=darkgray,sf},font={color=darkgray,sf,footnotesize},justification=raggedright,singlelinecheck=false]{caption}

\usepackage{notes2bib}
\newcommand{\SM}{}
\newcommand{\fullsection}{\paragraph*}
\newcommand{\myfootnote}{\footnote}

\newcommand{\fullonly}[1]{#1}
\newcommand{\scienceonly}[1]{}

\title{Topological structure of complex predictions}
\author{Meng Liu \and Tamal K.~Dey \and David F.~Gleich}
\input{preamble.tex}

\usepackage{graphicx}
\graphicspath{{figures_low_res/}}

\usepackage{cleveref}
\begin{document}

\makeatletter 
\noindent\begin{minipage}{\textwidth}
\centering 
{\large\itshape\bfseries\@title}
\bigskip

{Meng Liu, Tamal K.~Dey, David F.~Gleich}\\
{\footnotesize Purdue University, Computer Science}\\
\end{minipage}
\makeatother

\theoremstyle{definition}
\newtheorem{definition}{Definition}[section]
\DeclareRobustCommand{\rchi}{{\mathpalette\irchi\relax}}
\newcommand{\irchi}[2]{\raisebox{\depth}{$#1\chi$}} 

\begin{abstract}
    \input{abstract}
\end{abstract}

\part*{Overview and Central Results}
\input{main-findings}



\scienceonly{\bibliography{refs}\bibliographystyle{Science}}

\newpage

\scienceonly{\baselineskip14pt}
\part*{Detailed Discussion And Description of GTDA}

\vspace{-2\baselineskip}
\setcounter{tocdepth}{1}
\tableofcontents

\scienceonly{\input{scimain-supplementary-summary}}
\renewcommand{\thesection}{\S\arabic{section}}
\setcounter{section}{0}
\renewcommand{\thefigure}{\arabic{figure}}
\renewcommand{\thetable}{\arabic{table}}
\renewcommand{\thealgorithm}{\arabic{algorithm}}

\section{Our GTDA method for Reeb nets \emph{\&} prediction functions}
\label{sec:method}
\input{method.tex}
\section{Demonstration in graph based prediction}

\input{exp_graph.tex}
\section{Understanding image predictions}\label{sec:imagenette}
\input{exp_image.tex}
\section{Comparing models on ImageNet-1k predictions}
\label{sec:imagenet-1k}
\input{exp_image_comparison.tex}

\clearpage

\input{exp_mutation.tex}
\clearpage

\input{exp_chest.tex}
\input{parameters.tex}
\input{scaling.tex}
\input{exp_tsne_umap.tex}
\section{Code availability}
The implementation of GTDA framework we developed is available at 
\begin{center}
    \url{https://github.com/MengLiuPurdue/Graph-Topological-Data-Analysis}
\end{center}
along with all supporting code for the results in this paper.  Demos are found at 
\begin{center}
    \url{https://mengliupurdue.github.io/Graph-Topological-Data-Analysis/}
\end{center}
and show the Reeb networks from the first four figures. High resolution figures as well as an animated demo are found at
\begin{center}
    \url{https://github.com/MengLiuPurdue/Graph-Topological-Data-Analysis/tree/main/high-res-figures}
\end{center}

\bibliographystyle{plain}
\bibliography{refs}

\end{document}

%% file: preamble.tex
\usepackage{geometry}
\usepackage{amsmath}
\usepackage{amssymb}
\usepackage{textcase}
\usepackage{soul}
\usepackage{indentfirst}

\newcommand{\mat}[1]{\boldsymbol{#1}}
\renewcommand{\vec}[1]{\boldsymbol{\mathrm{#1}}}

\providecommand{\mA}{\ensuremath{\mat{A}}}

\providecommand{\mD}{\ensuremath{\mat{D}}}

\providecommand{\mP}{\ensuremath{\mat{P}}}

\providecommand{\ve}{\ensuremath{\vec{e}}}

%% file: abstract.tex
Complex prediction models such as deep learning are the output from fitting machine learning, neural networks, or AI models to a set of training data. These are now standard tools in science. A key challenge with the current generation of models is that they are highly parameterized, which makes describing and interpreting the prediction strategies difficult. 
    We use topological data analysis to transform these complex prediction models into pictures representing a topological view. The result is a map of the predictions that enables inspection. The methods scale up to large datasets across different domains and enable us to detect labeling errors in training data, understand generalization in image classification, and inspect predictions of likely pathogenic mutations in the BRCA1 gene. 

%% file: main-findings.tex
\fullsection{Introduction}
Deep learning is a successful strategy where a highly parameterized model makes human-like predictions across many fields~\cite{avsec2021effective,Esteva2017,Reichstein2019,Townshend2021}. 
Yet challenges in generalization may keep deep learning from use in practice~\cite{Zech2018,OakdenRayner2022}. Detailed prediction mechanisms are also difficult to assess directly due to the large collection of model parameters. 
Topological methods of data analysis excel at distilling representation invariant information from large datasets~\cite{singh2007topological,Nicolau2011,lum2013extracting}. 
However, topological data analysis (TDA) of these complex predictive models such as deep learning  remains in its infancy~\cite{topology-deep-learning,bodnar2021deep}. 

\begin{figure}[p]
    \centering
      \includegraphics[width=\linewidth]{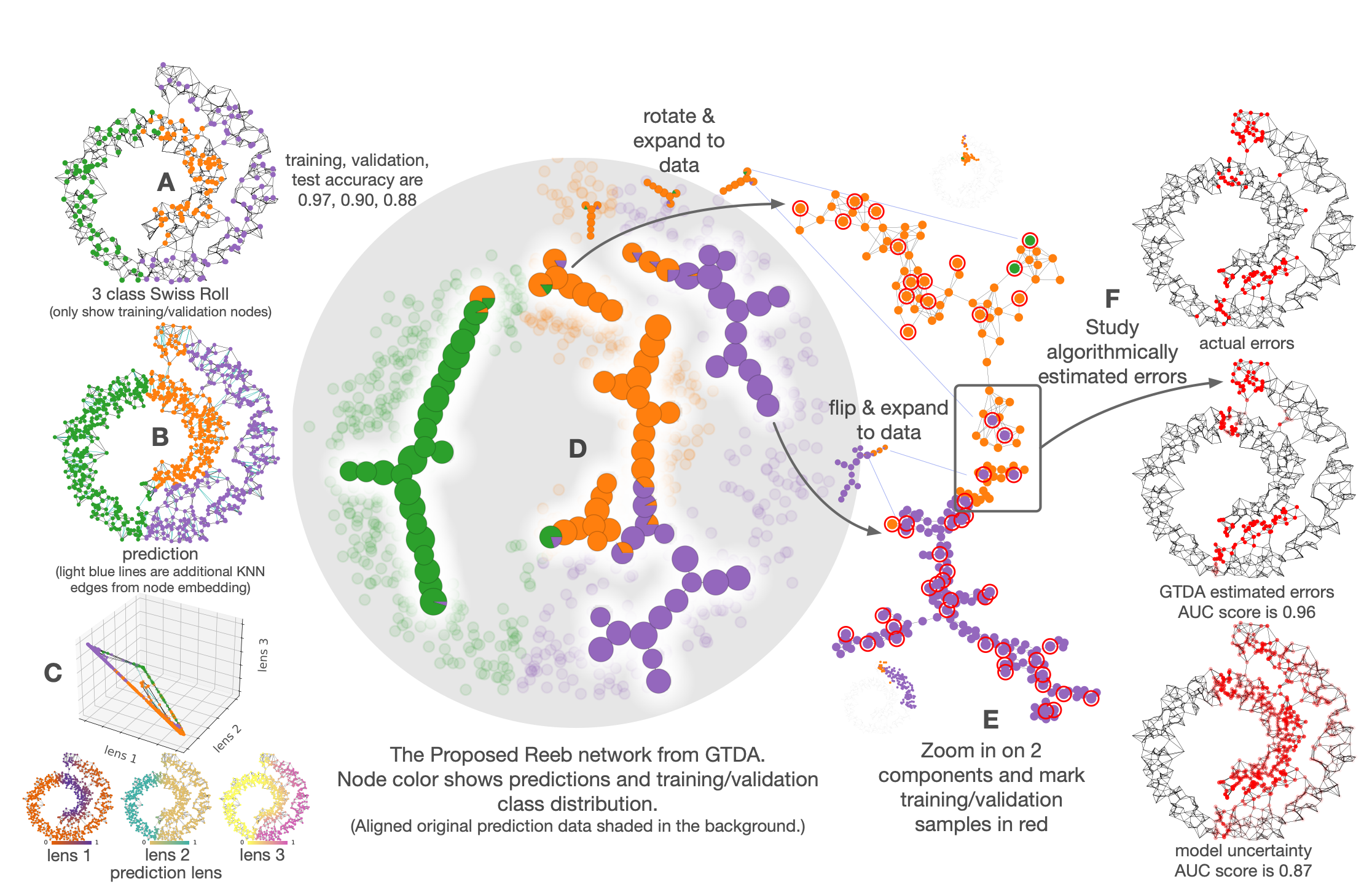}
     \caption{
     Consider a prediction scenario with three classes in a Swiss roll structure and an underlying graph (A). Graph neural network predictions  show reasonable accuracy (B). The 3-dimensional prediction lens from the neural network is shown in (C) and gives a guide to class predictions. The Reeb network is shown in (D). Each node maps to a small cluster of similar values of the lens. Nodes are colored by the fraction of points in each predicted class. 
     Regions with mixed predictions indicate potential ambiguities in the results. Further study of two such connected regions (E) shows many datapoints where there are training points with different labels closer to the given test points.   This situation motivates an algorithmic error estimate for each datapoint without ground truth (F). This estimate is nevertheless highly correlated with true errors and better than class uncertainty estimates. The topological simplification highlights datapoints with confusing or ambiguous predictions given the totality of predictions. 
     }
  \label{swiss_roll_combined}
  \label{fig:swiss_roll_combined}
  \end{figure}

We construct a Reeb network to assess the prediction landscape of a neural network-like prediction method (Figure~\ref{fig:swiss_roll_combined} shows an example). \emph{Reeb networks} are discretizations of topological structures called Reeb spaces,
which generalize Reeb graphs~\cite{dey2016multiscale,lum2013extracting}.\myfootnote{The term network or net is often used to mean a graph abstraction of a complex system. A Reeb graph is a topological structure that gives univariate topological information and produces a graph. A Reeb space is a more complicated multidimensional structure. A Reeb network is an undirected graph like a Reeb graph, and a Reeb network shows multidimensional topological information like a Reeb space. See \SM Section~\ref{sec:reeb-space}.}
Each node of the Reeb network (Figure~\ref{fig:swiss_roll_combined}D) is a local simplification of the prediction space and is computed as a cluster of datapoints with similar predictions.
Reeb nodes are linked if they share any datapoint or, in some cases, represent nearby datapoints. 

This construction suggests that datapoints within a Reeb network node should have the same prediction. Also, connected neighborhoods of the Reeb network should share predicted values. When this scenario is violated (Figure~\ref{fig:swiss_roll_combined}E), such as at a prediction boundary or ambiguous region in prediction space, it suggests a small set of points for further investigation. Additional algorithmic analysis of the Reeb network and relationships between predictions and training data can be used to diagnose \emph{estimated errors} in the prediction function without any access to ground truth information (Figure~\ref{fig:swiss_roll_combined}F, additional information in Section~\ref{sec:error-estimation}). Existing topological techniques are limited in analyzing predictions.\myfootnote{\SM Section~\ref{sec:standard-tda}, Figure~\ref{fig:tda_img_pred}, Figure~\ref{swiss_roll_steps}D.}

These Reeb networks identify and display information about the prediction function and the associated datapoints relevant to understand those predictions. This makes it comparable to other widely used visualization techniques such as tSNE~\cite{tSNE}, UMAP~\cite{Becht2018,umap}, and nonlinear dimension reduction~\cite{Tenenbaum-2000-isomap}, while providing greater information about the boundaries between prediction and localizing informative relationships between training data and prediction.









\fullsection{The Reeb network construction on a prediction function}
Computing a Reeb network for a complex prediction function or deep learning method has a few prerequisites.
There must be a large set of datapoints with unknown labels beyond those used for training and validating the prediction model, which is common when gathering data is easy. There must be known relationships among all datapoints such as (i) a given graph of relationships among all points, (ii) a nearest neighbor computation to create such a graph, or (iii) a domain-relevant means of clustering related points. All of our examples use (i) and (ii) and in this case, nodes of the Reeb network are created via a parameter free connected components analysis on subsets of the graph. Finally, there must be a real-valued guide to each predicted value, such as the last layer of a neural network. Following the terminology from Lum~\cite{lum2013extracting}, these are called \emph{lenses} (Figure 1C). 

Constructing the Reeb net from these prerequisites involves two main choices: the maximum size of a Reeb node or cluster and the amount of overlap in Reeb nodes.~\myfootnote{Other parameters are less influential. See \SM Table~\ref{tab:params}.
} 
We employ a recursive splitting and merging procedure called GTDA (graph based TDA) to build the Reeb net from the original datapoints and graph. This uses the lenses to identify datapoints with similar predictions and splits them into overlapping regions.~\myfootnote{We found it helpful to first smooth the information from the lenses over the relationship graph to avoid sharp gradients using 5 or 10 steps of an iterative smoothing procedure related to a diffusion.} Overlap is needed to use the connection strategy among nodes of the Reeb network. At each step, the data are clustered by computing connected components of the graph of relationships and the analysis proceeds iteratively for newly created components until they are smaller than the selected maximum size. Each remaining cluster constitutes a node of the Reeb net. Nodes of the Reeb network are connected if they share any datapoint. This connection strategy may leave many nodes isolated, which is not helpful to understand predictions. We reduce this isolation by adding edges from a minimum spanning tree.\myfootnote{Full algorithm in \SM Section~\ref{sec:construction}.} Useful results arise from a wide range of parameters.~\myfootnote{Further discussion of parameter sensitivity is in \SM Section~\ref{sec:sensitivity}.} 

Constructing a Reeb net is a scalable operation: analyzing 1.3 million images in ImageNet~\cite{ILSVRC15} with 2000 lenses for 1000 classes in a comparison of ResNet~\cite{He2016} and AlexNet~\cite{AlexNet} takes 7.24 hours.\myfootnote{\SM Table~\ref{tab:running_time} for additional runtime information.} 

\fullsection{Demonstration in Graph-based prediction}

\begin{figure}[p]
    \centering
      \includegraphics[width=\linewidth]{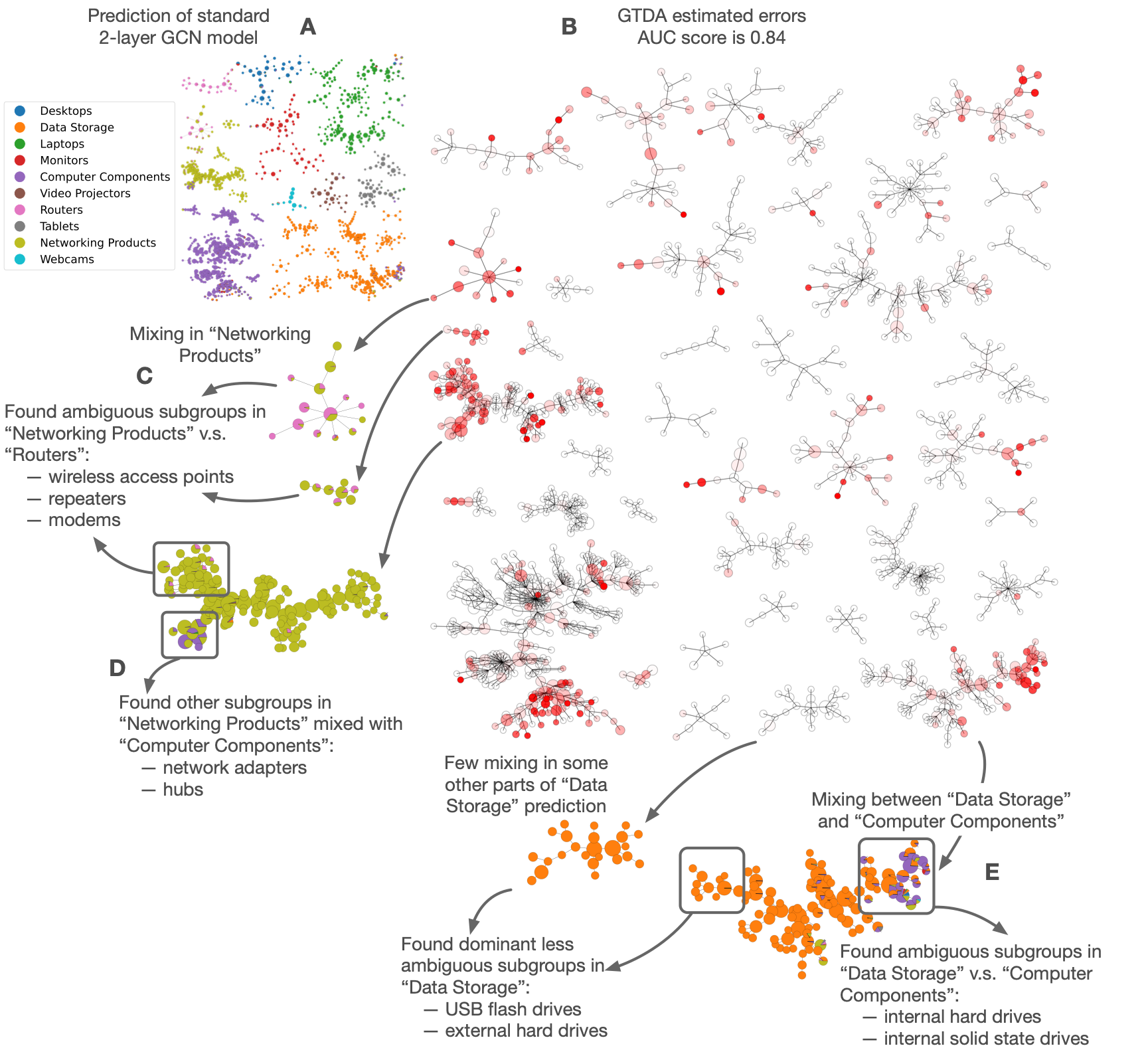}
     \caption{Reeb network of a standard 2-layer graph convolutional network model trained and validated on 10\% labels of an Amazon co-purchase dataset (A) and estimated errors shown in red (B). The map highlights ambiguity between ``Networking Products'' and ``Routers''. Checking these products shows wireless access points, repeaters or modems as likely ambiguities (C). Additional label ambiguities involve ``Networking Products'' and  ``Computer Components'' regarding  network adapters (D); likewise ``Data Storage'' and ``Computer Components'' are ambiguous for  internal hard drives (E). These findings suggest that the prediction quality is limited by arbitrary subgroups in the data, which Reeb networks helped  locate quickly. }
  \label{amazon_analysis}
  \label{fig:amazon}
  \end{figure}

We apply the Reeb net framework to a graph neural network that predicts the type of product on Amazon based on reviews. This framework identifies a key ambiguity in product categories that limits prediction accuracy (Figure~\ref{amazon_analysis}). Specifically, ``Networking Products'' and ``Routers'' overlap (a Router is a specific type of Network Product) and show high levels of confusion as do ``Data Storage'' and ``Computer Components'' (an internal data storage drive is a computer component). These results suggest that large improvements are unlikely with better algorithms and would require label improvements to differentiate categories or other divisions in a hierarchy~\cite{NIPS2011_d5cfead9}. 
This was verified by checking another graph neural network~\cite{chien2021adaptive} with similar behavior.\myfootnote{\SM Section~\ref{sec:amazon-advanced}.}

\fullsection{Understanding image predictions}

\begin{figure}[p]
    \centering
      \includegraphics[width=\linewidth]{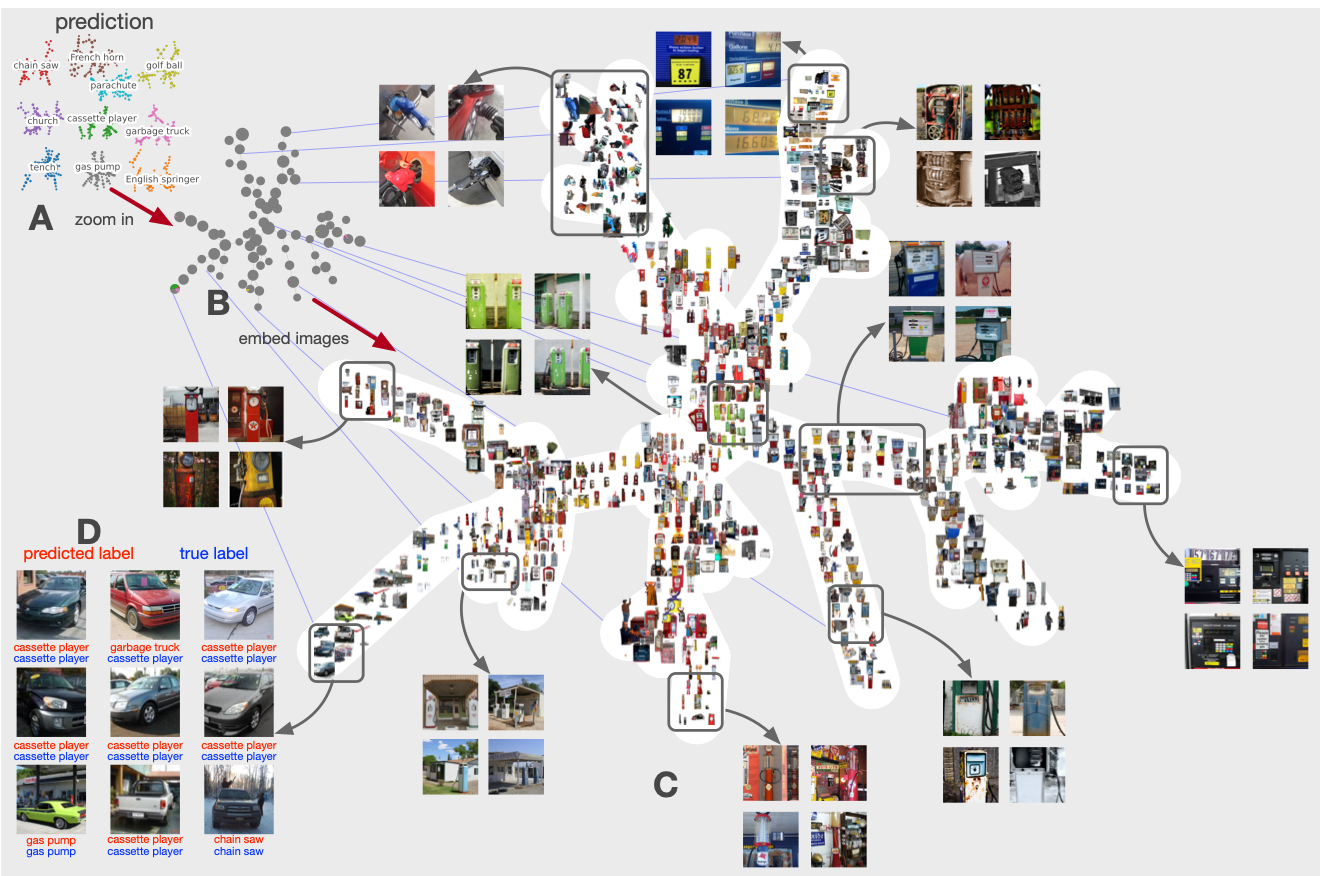}
     \caption{We take a pretrained ResNet50 model and retrain the last layer to predict 10 classes in Imagnette (A). In (B), we zoom into the Reeb network group of ``gas pump'' predictions and display images at different local regions (C). This shows gas pump images with distinct visual features. Examining these subgroups can provide a general idea on how the model will behave when predicting future images with similar features as well as help us quickly identify potential labeling issues in the dataset. For instance, we find a group of images in (D) whose true labels are ``cassette player'' even though they are really images of ``cars''.}
  \label{gas_pump}
  \end{figure}

When the framework is applied to a pretrained ResNet50 model~\cite{He2016} on the Imagnette dataset~\cite{imagenette}, then it produces a visual taxonomy of images suggesting \emph{what} ResNet50 is using to categorize the images (Figure~\ref{gas_pump}). This example also highlights a region where the ground truth labels of the datapoints are incorrect and cars are erroneously labeled as ``cassette player''.\fullonly{\myfootnote{We conjecture a car labeled as ``cassette tape'' may be due to images of cars listed for sale including the string ``cassette tape player.''}}

\fullsection{Understanding Malignant Gene Mutation Predictions}

Reeb networks from a proposed DNA prediction method~\cite{avsec2021effective}, when applied to the BRCA1 gene, show Reeb components that are localized in the DNA sequence and also that these components map to secondary structures, e.g., in the 1JNX repeat region, that aid interpretation. For one of the helix structures, this analysis shows regions where insertions and deletions are harmful (pathogenic) and  single nucleotide variants lack evidence of harm. In an analysis of a component with many harmful predictions, these results show that places where the framework incorrectly predicts errors are strongly associated with insignificant results in the underlying data.

\begin{figure}[p]
    \centering
      \includegraphics[width=\linewidth]{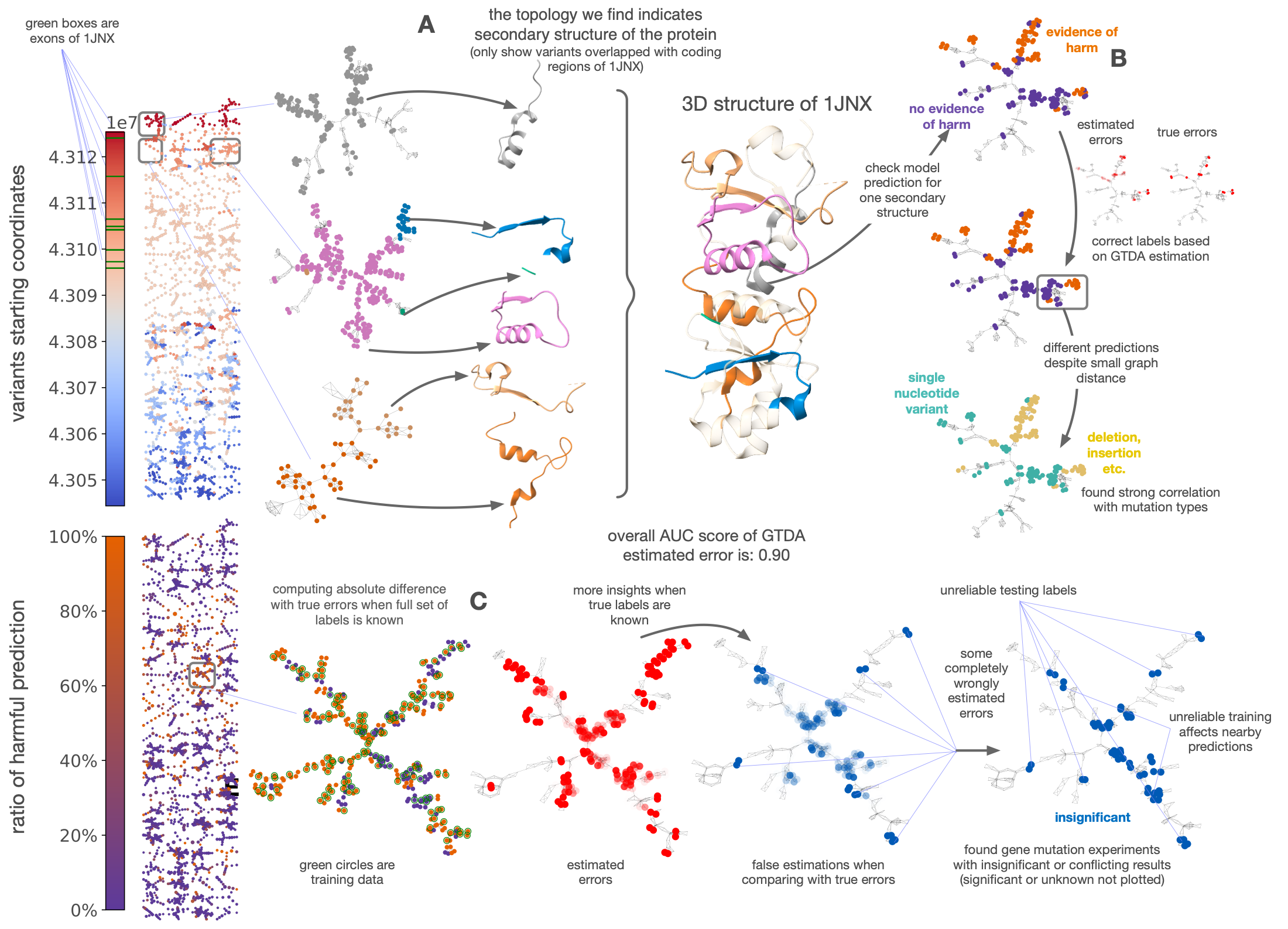}
     \caption{We use Reeb networks to visualize harmful (likely pathogenic) and potentially non-harmful (no evidence of pathogenicity) predictions of gene variants in BRCA1. The topology  indicates several secondary structures on part of the protein (1JNX) as shown in (A). We further check the model predictions on variants targeting one secondary structure (B). Our error estimate shows a number of likely erroneous predictions, and we flip these expected errors (a final analysis showed these errors were correctly identified). 
     We continue to see variants with distinct prediction in a small region of a few amino acids. Close examination shows a strong association between mutation types and model predictions where deletion or insertion is more likely to be harmful than a single nucleotide variant. Additional insights when using the full label set show some estimated errors are completely wrong (C). These prediction mistakes involve gene mutation experiments with insignificant or conflicting results and indicate underlying uncertainty.}
  \label{mutation}
  \end{figure}
  
\fullsection{Additional examples}
Two additional studies show how the Reeb nets find incorrect diagnostic annotations in chest x-rays datasets used for deep learning~\cite{wang2017chestx} (AUC 0.75, details in \SM Section~\ref{sec:chest-xray}) and help compare deep learning predictions from  ResNet~\cite{He2016}, AlexNet~\cite{AlexNet}, and VOLO-50~\cite{yuan2021volo}, showing the importance of image backgrounds or partial objects to improvements over AlexNet (\SM Section~\ref{sec:comparison}). 

\fullsection{Related work}
Our Reeb network construction extends recent analytical methods from topology~\cite{lum2013extracting,singh2007topological} to facilitate applications to the topology of complex prediction. 
Prior work on analyzing deep learning methods for errors focuses on a single result list~\cite{koh2017understanding}, without the associated topological structure provided by Reeb nets. Our work relates to \emph{interpretability}~\cite{Murdoch2019} and seeks to produce a comprehensible \emph{map} of the prediction structure to aid navigation of a large space of prediction to those most interesting areas. 
Beyond identifying that there is a problem, the insights from the topology suggest relationships to nearby data and thereby suggest mechanisms that could be addressed through future improvements. 


\fullsection{Conclusion}
Considering the ability of these topological inspection techniques to translate prediction models into actionable human level insights -- from label issues to protein structure -- we expect them to be applicable to new models and predictions, broadly, as they are created and to be a critical early diagnostic of prediction models.  The interaction of topology and prediction may provide a fertile ground for future improvements in prediction methods. 

%% file: method.tex
In this paper, we developed a framework to inspect the predictions of complex models by visualizing the interactions between predictions and data. The framework has the following properties:
\begin{itemize}
\item it can produce a topological view of the original dataset through pictures
\item the visualization can provide clues for any sample of interest to be inspected
\item it is highly scalable and can process large datasets with thousands of classes
\item it can provide intuitive insights and suggest places that are worth a further study for users without any prior knowledge on the model or the data
\end{itemize}

\subsection{Background: Topological Data Analysis and the Mapper Algorithm}

Our method is rooted in the growing field of computational topology and topological data analysis and the framework is closely related to the \emph{mapper} algorithm~\cite{singh2007topological} for topological data analysis (TDA). Mapper builds a discrete approximation of a Reeb graph or Reeb space (see Section~\ref{sec:reeb-space}, Figure~\ref{fig:reeb-space}). It begins with a set of datapoints $(x_1, \ldots, x_n)$, along with a single or multi-valued
function sampled at each datapoint. The set of all these values $\{f_1, \ldots, f_n\}$ samples a map $f: X\rightarrow \mathbb{R}^k$ on a topological space $X$. The map $f$ is called a \emph{filter} or \emph{lens}. The idea is that when $f$ is single valued, a Reeb graph shows a quotient topology of $X$ with respect to $f$ and mapper discretizes this Reeb graph using the sampled values of $f$ on points $x_1, \ldots, x_n$.  Algorithmically, \emph{mapper} consists of the steps:
\begin{itemize} 
\item Sort the values $f_i$ and split them into overlapping bins $B_1, \ldots, B_{r}$ of the same size. 
\item For each bin of values $B_j$, let $S_j$ denote the set of datapoints with that same value and \emph{cluster} the datapoints in each $S_j$ independently. (That is, we run a clustering algorithm on each $S_j$ as if it was the entire dataset.)
\item For each cluster found in the previous step, create a node in the Reeb graph.
\item Connect nodes of the Reeb graph if the clusters they represent share a common point. 
\end{itemize}
The resulting graph \textbf{is a discrete
approximation of the} \emph{Reeb graph} and represents a compressed view of the shape underlying the original dataset.  

Our goal is to extract a similar type of topological description for lenses that are multi-valued, which we interpret as a collection of single-valued lenses.

\subsection{Rationale for a graph-based method}

The input format for \emph{mapper} is usually a point cloud in a high dimensional space where the point coordinates are used only in the clustering step. 

In our methodology, we are interested in datasets that are even more general. Graph inputs provide this generality. Datasets not in graph format like images or DNA sequences can be easily transformed into graphs by first extracting intermediate outputs of the model as embeddings and then building a nearest neighbor graph from the embedding matrix. Then the resulting graph facilitates easy clustering: for each subset of points, we extract the subgraph induced by those points and then use a parameter-free connected components analysis to generate clusters.

Our method could also work with point cloud data and clustering directly through standard relationships between graph-based algorithms and point cloud-based algorithms. We focus on the graph-based approach for simplicity and because we found it the most helpful for these applications. 

\subsection{The Reeb network construction on a prediction function using a graph (GTDA)}
\label{sec:gtda-method}
\label{sec:construction}
We take as input:
\begin{enumerate}
    \item an $n$-node graph $G$
    \item a set of $m$ lenses based on a prediction model as an $n \times m$ matrix $\mP$
\end{enumerate} 
The lenses we use are the prediction matrix $\mP$ of a model where $P_{ij}$ is the probability that sample $i$ belongs to class $j$. Key differences from existing studies of TDA frameworks on graphs include using the connected components of each bin~\cite{bodnar2021deep,hajij2018mapper} as clusters and also additional steps to improve the analysis of prediction functions by adding weak connections from a minimum spanning tree. 

\paragraph{Problems with straightforward algorithmic adaptation.}
Mapper does extend to multidimensional lens functions by using a tensor product bin construction. We found issues with a straightforward adaptation of \emph{mapper} to such multidimensional input for prediction functions. In our extensive trials, we found that most of the resulting Reeb networks end up with too many tiny components or even singletons where no prediction-specific insights were possible. This is especially so when the dataset has many classes, most multi-dimensional bins will just contain very few samples because the space grows exponentially, limiting the potential of overlap to find relationships. Simply reducing the dimension of $\mP$ with PCA will lose the interpretability of the lens. Moreover, classic \emph{mapper} uses a fixed bin size and density-based or multi-scale alternatives~\cite{dey2016multiscale} were unsuccessful in our investigations although they solve this problem from a theoretical perspective. (We note this is a potential area for followup work to better understand why.) 


\paragraph{Preprocessing to smooth the predictions.}
As a preprocessing step, we apply a few steps (usually four or five) of the smoothing iteration: $\mP^{(i+1)}=(1-\alpha)\mP+\alpha\mD^{-1}\mA\mP^{(i)}$. Here $\mP^{(0)}=\mP$, $\mA$ is the adjacency matrix of the input graph, $\mD$ is the diagonal degree matrix and $0<\alpha<1$. This helps to prevent prevent sharp changes between adjacent nodes. This equation is a diffusion-like equation closely related to the PageRank vector that is known to smooth data over graphs and has many uses~\cite{gleich2015pagerank}.  The iteration keeps all the prediction data non-negative and the smoothed $\mP$ will also be min-max column normalized so that each value is between 0 and 1. As is standard, this setup can use any weights associated with the adjacency matrix, or remove them and use an unweighted graph. 

\paragraph{Our graph-based construction for a prediction function.} 
The following approach was used for datasets in the main paper. We call this a graph-based topological data analysis framework (GTDA). It uses a recursive splitting strategy to build the bins in the multidimensional space. For each subgroup of data, the idea is that we find the lens that has the maximum difference on those data. Then split the component by putting nodes into two approximately equal sized overlapping bins based on the values in this lens. Then if the resulting groups are big enough, we add them back as sets to consider splitting. 

Detailed pseudo code can be found in Algorithm~\ref{GTDA}.\scienceonly{ An animation of the method can be found in the supplemental video.} We give a quick outline here. The recursive splitting starts with the set of connected components in the input graph. This is a set of sets: $\mathbb{S}$. The key step is when the algorithm takes a set $\mathbb{S}_i$ from $\mathbb{S}$, it splits $\mathbb{S}_i$ into new (possibly) overlapping sets $\mathbb{T}_1, \ldots, \mathbb{T}_{h}$ based on the lens with maximum difference in value on $\mathbb{S}_i$ and also connected components. Each $\mathbb{T}_i$ is then either added $\mathbb{S}$ if it is large enough (with more than $K$ vertices) and where there exists a lens with maximum difference larger than $d$. Otherwise, $\mathbb{T}_i$ goes into the set of finalized sets $\mathbb{F}$. 


Once we have the final set of sets, $\mathbb{F}$, then we do have two final merging steps, along with building the Reeb net. The first is to merge sets in $\mathbb{F}$ if they are too small (\Cref{node_merging}). The second is to add edges to the Reeb net to promote more connectivity (\Cref{component_merging}). 

In the first merging (\Cref{node_merging}), which occurs before the Reeb net is constructed, we check and see if any set in $\mathbb{F}$ is too small (smaller or equal to $s_1$). If so, then we find nearby nodes based on the input graph $G$ and based on a user-provided distance measure $f$ and merge the small component with the closest component (giving preference to the smallest possible set to merge into). This could be a simple graph-distance measure (e.g. shortest path), something suggested by the domain, or a weight based on the prediction values (what we use). The algorithm is closely related to Bor\r{u}vka's algorithm for a minimum spanning tree. 

Next, we build the Reeb net $\hat{G}$ from this set of sets $\mathbb{F}$. Each group $\mathbb{F}_i$ becomes a node, and nodes are connected if they share any vertex. 

In the second merging (\Cref{component_merging}) we seek to improve the overall connectivity of the Reeb net by connecting small disconnected pieces of the Reeb net $\hat{G}$. This step is designed to enhance the ability to work with predictions by adding weaker connections to the more strongly connected topological pieces. It uses the same distance measure $f$ to find components and uses a similar Bor\r{u}vka-like strategy. We save the set of edges added at this step to study in the error estimation procedures noted below.


\paragraph{Choices for the parameters.}
As a result, GTDA has 8 parameters as in \Cref{tab:params}. Tuning of the parameters is straightforward, and we often use the default choice or values from a small set.  The values $K$, $d$ and $s_1$ provide direct control about the number of nodes in the final group visualization, while $r$ and $s_2$ control how connected we want the visualization to be. In practice, we could first tune $K$ and $d$ to determine the number of nodes, then tune $r$ so that no component in the Reeb net is too large and finally tune $s_1$, $s_2$ to remove any tiny nodes or components. We leave the smoothing parameters fixed at $\alpha = 0.5$ and $S = 5$ or $10$ (very smooth). A detailed discussion on these parameters can be found in \Cref{sec:params}.

\paragraph{Choice of distance function for merging }
Possibly the hardest parameter to pick is the merging function $f$.
The user can choose any distance metric $\emph{f}$ in the merging step, in our experiments, we use $\ell^\infty$ norm of the difference between rows of the preprocessed $\mP$ as the distance between 2 samples, which roughly means how much larger the bin containing one of those 2 samples should be in order to include the other sample. Put another way, this choice makes us less sensitive to the exact choice for $r$ because we will add small connections that would have been included in a slightly larger bin.

\paragraph{Drawing the graph.}
Unless otherwise specified, all coordinates of any layout we show are computed with Kamada-Kawai algorithm~\cite{kamada1989algorithm}.

\paragraph{Showing the Reeb network and explorations.}
In the Reeb net of a prediction function, we draw each node as a small pie-chart. The size of the pie-chart represents the number of nodes. The pieces of the pie show the local prediction distribution. In some cases, we find it useful to study the predicted labels directly, such as when studying mechanisms underlying the prediction. In other cases, we find it useful to study predictions and training data, such as when studying possible errors. 
These visualizations facilitate exploring regions of the prediction landscape based on interactions among predicted values and training data. By mapping these \emph{small} regions back to the original data, it suggests what the model is utilizing to make the predictions. Examples on this can be found in the experiments in the main text as well as in the supplemental information. 

\begin{table}[t]
  \centering
  \begin{tabular}{rll}
  \toprule
  parameter & description & suggested choices\\
  \midrule
  $K$ & component size threshold to stop splitting & 5\% of smallest class size\\
  $d$ & lens difference threshold to stop splitting & 0 or 0.001 \\
  $r$ & overlapping ratio & 0.01 \\
  $s_1$ & Reeb node size threshold & 5 \\
  $s_2$ & Reeb component size threshold & 5\\
  $\alpha$ & lens smoothing parameter & 0.5 (used in all experiments)\\
  $S$ & lens smoothing steps & 5 or 10 \\
  $\emph{f}$ & distance function in the merging step & $\ell^{\infty}$ difference of row $i$, $j$ of $\mP^{(S)}$\\
  \bottomrule
  \end{tabular}
  \caption{List of parameters in GTDA.}
  \label{tab:params}
  \end{table}

\subsection{Demonstration of GTDA} 
\begin{figure}[tp]
  \centering
    \includegraphics[width=\linewidth]{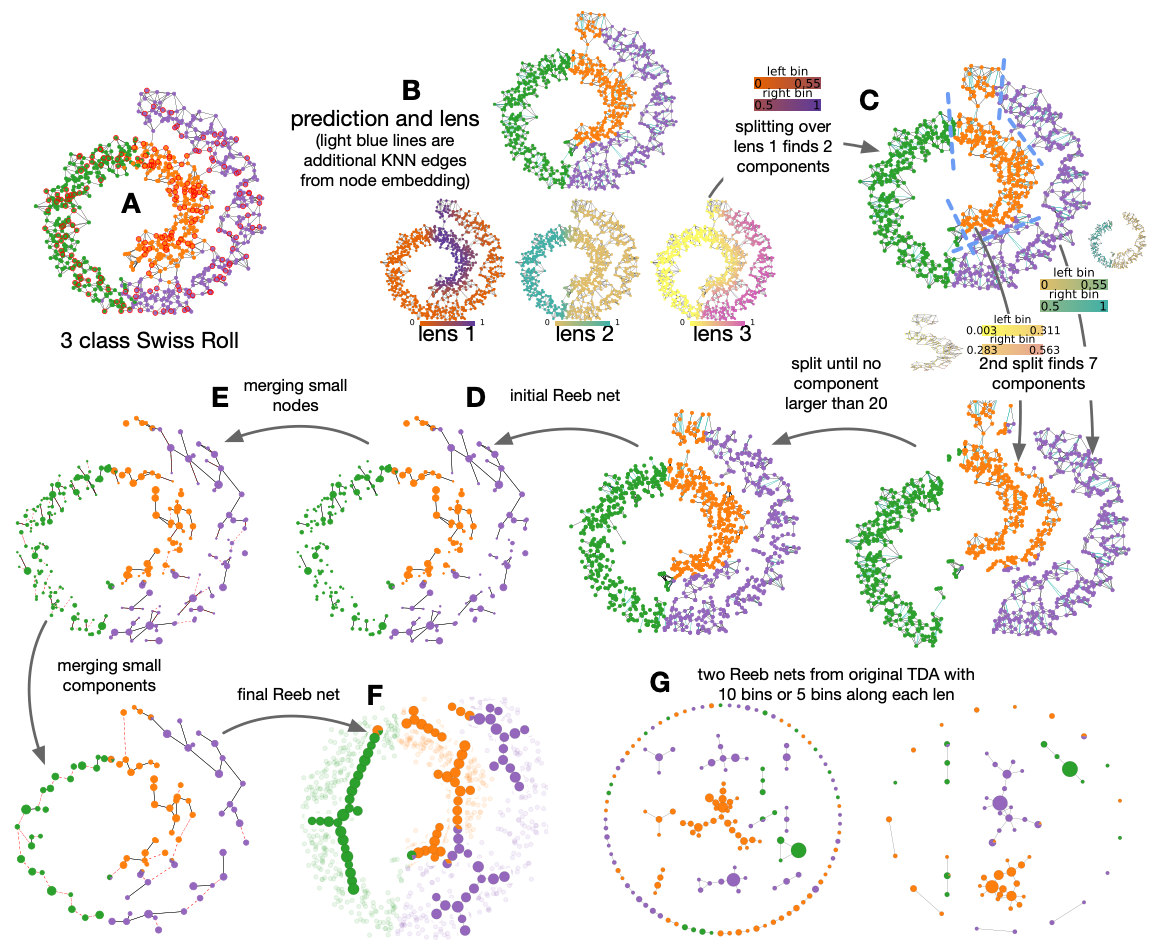}
   \caption{A detailed illustration of applying GTDA to build a Reeb net on a 3-class Swiss roll dataset. The original data graph and ``ground truth'' values are in (A). We show the model prediction for a simple GCN and the three prediction lenses (after smoothing) in (B).\scienceonly{ See also the supplementary video illustrating the process.} The first splitting iteration over lens 1 finds 2 components, (C). At the second split, for each component, we choose the lens with the largest difference, which means the outer ring is split over lens 2 and the inner ring is split over lens 3. The second splitting finds 7 components in total. We continue to split until no more components larger than 20 and get the initial Reeb net, (D). Then small nodes are merged to neighbors iteratively as shown by the red dashed lines in (E). Similarly, small components in the Reeb net are iteratively connected to get the final Reeb net in (F). As a comparison, two Reeb nets from the original \emph{mapper} using 10 lens or 5 lens have many isolated nodes or components and are not suitable for the subsequent inspection. The figure (F) uses predicted classes for training and validation points instead of the actual training and validation classes as in \cref{fig:swiss_roll_combined}(D).}
\label{swiss_roll_steps}
\end{figure}

We use a 3 class Swiss roll dataset to demonstrate each step of our GTDA framework (plot (A) of figure~\ref{swiss_roll_steps}). For the GTDA parameters, we set $K=20$, $d=0$, $r=0.1$, $s_1=5$, $s_2=5$, $\alpha=0.5$, and $S=5$. In (B), we show the three prediction lenses we use in the top plot as well as the predicted labels of the model we use. We also add additional edges based on nearest neighbors from node embeddings to take node features into account. This is standard practice in graph neural network methods. Details on the dataset and the model can be found in \Cref{sec:intro_details}. Each lens is just the prediction probability of a class after smoothing and normalization. In (C), we pick the lens with the largest min-max difference and split it into 2 bins with 10\% overlap (we pick the one with smaller index to break ties). This round of splitting finds 2 components. For each component found in the first iteration, we pick the lens with the largest min-max difference and split it again. In this case, the inner component is split along lens 3 while the outer component is split along lens 2. This round of splitting further divides the graph into 7 components. We repeat the splitting until no component has more than 20 vertices of the original graph. 

In the end, we find 247 unique components. As noted above, we use a pie chart to represent each Reeb node and connect Reeb nodes with black lines if they have any samples in common to get the initial Reeb net, (D). Node size is proportional to the number of samples it represents, the pie chart shows the distribution over predicted values. This initial Reeb net has many tiny components or even singletons that are a barrier to deeper insights; the merging steps address this issue. In (E), we use red dashed lines to mark how we will merge those small Reeb nodes so that all nodes will contain more than 5 samples. Similarly, we use red dashed lines to mark extra edges that will be added so that each connected component in the Reeb net will contain more than $5$ Reeb nodes. The final Reeb net is shown in (F) with the original graph embedded in the background. We can see that all important structures found in (D) are also preserved in (F) such as the mixing of nodes from different classes. And what merging does is to estimate how the tiny nodes and components are connected in the original graph or via the prediction lens so that we have a clear view of predictions over the entire dataset.  This supports an inspection of the model's prediction on any sample we want.

As a comparison, in plot (G), we show two Reeb nets that are constructed by the original \emph{mapper} algorithm with different number of bins along each lens. These Reeb nets are not useful to understand the prediction structure. Most samples from the green class are grouped into a few nodes because prediction probability distribution on this class is more skewed, which makes the inspection hard.

\begin{algorithm}[h!]
  \caption{$\texttt{GTDA}(G,\mP,K,d,r,s_1,s_2,\alpha,S,\emph{f})$ See \Cref{tab:params} for parameters description}
  \begin{algorithmic}[1]
  \label{GTDA}
  \STATE Smooth $\mP$ for $S$ steps with $\mP^{(i+1)}=(1-\alpha)\mP+\alpha\mD^{-1}\mA\mP^{(i)}$ and $\mP^{(0)}=\mP$
  \STATE Perform a min-max normalization along each column of $\mP$
  \STATE Find connected components in $G$ and put all components with size larger than $K$ and maximum lens difference larger than $d$ in $\mathbb{S}$, otherwise in $\mathbb{F}$
  \WHILE {$\mathbb{S}$ is not empty}
    \STATE Let $\mathbb{S}^{(\text{iter})}$ be a copy of $\mathbb{S}$
    \FOR {each $\mathbb{S}_i$ in $\mathbb{S}^{(\text{iter})}$}
      \STATE Remove $\mathbb{S}_i$ from $\mathbb{S}$
      \STATE Find column $c_i$ (for a lens) such that $\mP^{(S)}[\mathbb{S}_i,c_i]$ has the largest min-max difference
      \STATE Split interval $[\min(\mP[\mathbb{S}_i,c_i]),\max(\mP^{(S)}[\mathbb{S}_i,c_i])]$ into two halves of the same length and extend the left half by a ratio of $r$ to give overlapping groups $\mathbb{R}_1$ and $\mathbb{R}_2$ based on which vertices had values in the left and right parts of the interval. 
      \STATE Create sets $\mathbb{T}_1, \ldots, \mathbb{T}_h$ based on the connected components in $\mathbb{R}_1, \mathbb{R}_2$. 
      \FOR {each $\mathbb{T}_i$}
        \STATE If there are more than $K$ vertices in $\mathbb{T}_i$ and if there is a lens with a maximum difference larger than $d$, then add $\mathbb{T}_i$ to $\mathbb{S}$. Otherwise, add $\mathbb{T}_i$ to  $\mathbb{F}$.
      \ENDFOR
    \ENDFOR
  \ENDWHILE
  \STATE Run $\texttt{node-merging}(\mathbb{F},G,s_1,\emph{f})$ to get the updated $\mathbb{F}$
  \STATE Generate Reeb net $\hat{G}$ by considering each component of $\mathbb{F}$ as a Reeb net node and connecting two Reeb net nodes if their corresponding components have overlap
  \STATE Run $\texttt{component-merging}(\mathbb{F},G,\hat{G},s_2,\emph{f})$ to get the updated $\hat{G}$ and the extra set of edges $\mathbb{E}$
  \STATE Return $\hat{G}$, $\mathbb{E}$
  \end{algorithmic}
  %
  \end{algorithm}
  
  \begin{algorithm}[h!]
    \caption{$\texttt{node-merging}(\mathbb{F},G,s_1,\emph{f})$}
    \begin{algorithmic}[1]
    \label{node_merging}
    \WHILE{there exists components in $\mathbb{F}$ with at most $s_1$ vertices }
    \STATE Set $\mathbb{C}$ to be empty. 
    \FOR{each component $\mathbb{F}_i$ in $\mathbb{F}$ where $|\mathbb{F}_i| \le s_1$}
      \STATE for each edge $(v_i, v_j)$ in $G$ where $v_i \in \mathbb{F}_i$ and $v_j \in \mathbb{F}_j \not= \mathbb{F}_i$, compute $\emph{f}(v_i,v_j)$
      \STATE Select the pair of nodes $v_i$, $v_j$ with the smallest $\emph{f}(v_i,v_j)$. Let $\mathbb{F}_j$ be the set associated with $v_j$ and choose the smallest size $F_j$ if $v_j$ is in multiple such sets. 
      Add $(\mathbb{F}_i, \mathbb{F}_j)$ to $\mathbb{C}$. 
    \ENDFOR
    \STATE View the choices in $\mathbb{C}$ as edges of an undirected graph $H$ where vertices are $\mathbb{F}_i$.
    \STATE Compute connected components of $H$. 
    \FOR{each connected component $H_i$ of $H$ of size larger than 1}
    \STATE Let $\mathbb{F}_1, \ldots, \mathbb{F}_h$ be the underlying sets of $H_i$ from $\mathbb{F}$. Remove each $\mathbb{F}_i$ from $\mathbb{F}$. Then add $\mathbb{F}_1 \cup \ldots \cup \mathbb{F}_h$ to $\mathbb{F}$.
    \ENDFOR 
    \ENDWHILE
    \STATE Return the updated $\mathbb{F}$
  \end{algorithmic}
  \end{algorithm}
  
    \begin{algorithm}[h!]
    \caption{$\texttt{component-merging}(\mathbb{F},G,\hat{G},s_2,\emph{f})$}
    \begin{algorithmic}[1]
    \label{component_merging}
    \STATE Initialize the set of extra edges $\mathbb{E}$ to be empty
    \STATE Compute connected components of Reeb net $\hat{G}$
    \STATE Let $\mathbb{D}$ be the set of connected components of $\hat{G}$.
    \WHILE{there exists any $\mathbb{D}_i \in \mathbb{D}$ where $\mathbb{D}_i$ has at most $s_2$ Reeb nodes}
    \FOR{each $\mathbb{D}_i \in \mathbb{D}$ where $\mathbb{D}_i$ has at most $s_2$ Reeb nodes}
    \STATE Let $\mathbb{C}$ be the union of vertices in $G$ (not $\hat{G}$) for Reeb nodes in $\mathbb{D}_i$.
    \STATE For each edge $(v_i,v_j) \in G$ where $v_i \in \mathbb{C}$ and $v_j \not\in \mathbb{C}$, compute $f(v_i,v_j)$.
    \STATE Select the pair of nodes $v_i$, $v_j$ with the smallest $\emph{f}(v_i,v_j)$. Let $\mathbb{F}_i$ and $\mathbb{F}_j$ be the Reeb nodes associated with $v_i$ and $v_j$ and choose the smallest size $\mathbb{F}_j$ if $v_j$ is in multiple such sets. Pick an arbitrary $\mathbb{F}_i$ (we used smallest index in our data structure) if $\mathbb{F}_i$ in multiple such sets.
     \STATE Add $(\mathbb{F}_i, \mathbb{F}_j)$ to $\mathbb{E}$. 
     \STATE Connect the Reeb nodes for $\mathbb{F}_i, \mathbb{F}_j$ in $\hat{G}$
    \ENDFOR 
    \STATE Recompute connected components analysis of $\hat{G}$ and update $\mathbb{D}$
    \ENDWHILE
    \STATE Return $\hat{G}$ and $\mathbb{E}$
  \end{algorithmic}
  \end{algorithm}

\subsection{Error estimation of GTDA} 
\label{sec:error-estimation}
The model often highlights places where there is no reasonable basis for a prediction, e.g.~where there is training data with a different label closer to a prediction. This scenario admits an estimate where the model will likely make mistakes by checking the relative locations between predictions and training data. 

\begin{figure}[tp]
  \centering
  \includegraphics[width=\linewidth]{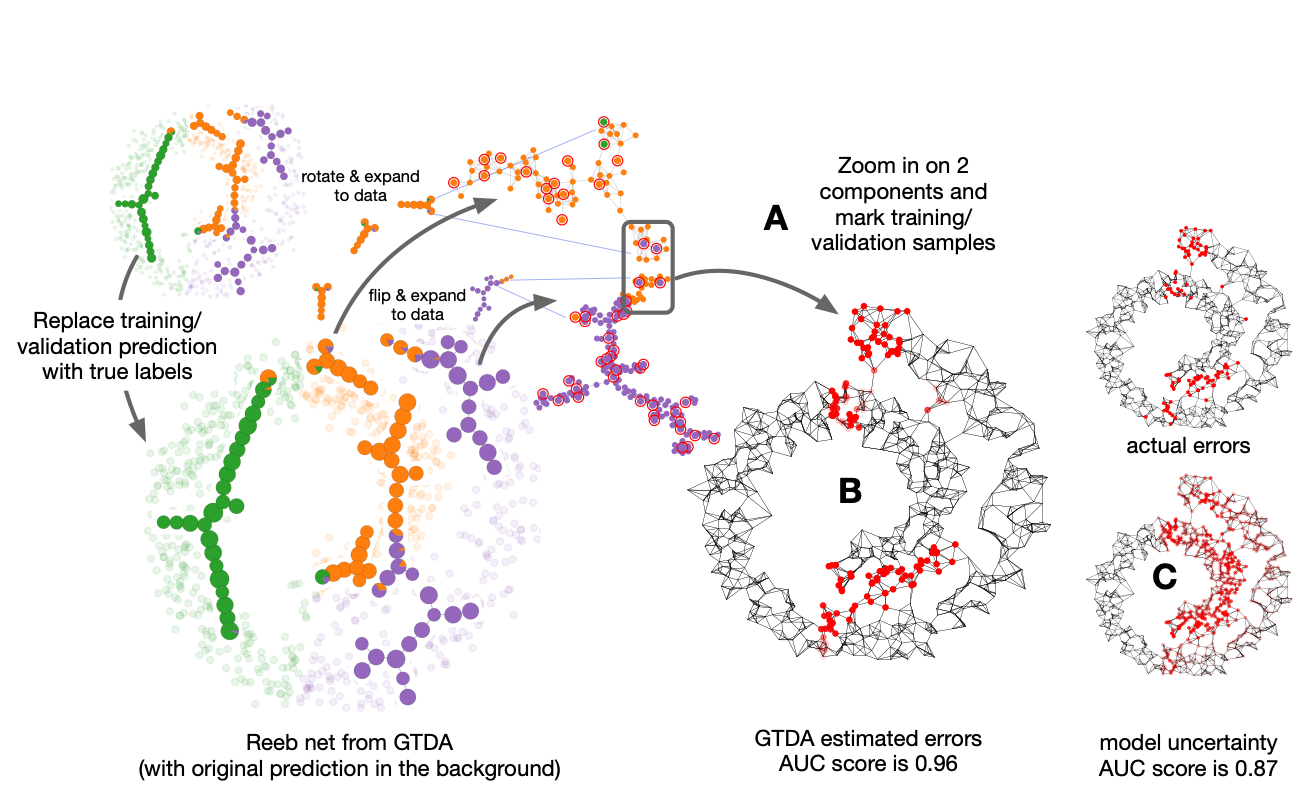}
  \caption{
  This figure demonstrates the procedure of estimating errors from the Reeb net produced by GTDA. In comparison with \Cref{swiss_roll_steps}, we show the training data labels in the pie charts instead of the predicted values. If we zoom in on two components and mark training and validation samples (red circles) with true labels, we see many orange predictions without any training or validation data nearby to support them (inset box nearby) (A), which suggests potential errors -- note that the model may be using additional features to predict these values, but these examples do merit closer inspection. We develop an error estimation procedure in \Cref{error_est} to automate this inspection. Overall, GTDA estimated errors have a AUC score of 0.95 with true errors (B), while using model uncertainty (one minus prediction probability) only has a AUC score of 0.87 (C).}
  \label{swiss_roll_errors}
\end{figure}

Using the Swiss roll example, in plot (A) of figure~\ref{swiss_roll_errors}, we zoom in on two components GTDA.  We then look at the induced subgraph of this region in a projection of the Reeb network. The Reeb network projection expands each Reeb node into the original set of input vertices with duplicated nodes merged and also adds in edges that we put into study predictions (the extra set $\mathbb{E}$ in the algorithms). A detailed projection procedure can be found in Algorithm~\ref{Reeb_projection}.  

Put formally: Given a set of Reeb network nodes in $\hat{G}$, find the union of all vertices in $G$ these nodes represent and call that $T$. We look at the induced subgraph of $T$ in the projection of the $\hat{G}$ from Algorithm~\ref{Reeb_projection}. 

To show these induced subgraphs, we can either use Kamada Kawai layout or, as an alternative to Kamada Kawai, we can also compute coordinates for each projected Reeb node and then combine different layouts using their relative coordinates in Reeb net. 

Then we use red circles to mark training and validation data and color them with the true labels. Unknown data points are still colored with predicted labels. 

One can immediately notice the problem: \emph{There are some orange predictions in the grey box, but there is no orange training or validation data nearby to support them.} Thus, either the model or the dataset itself have issues with these prediction and merit a second look. In this case, it is just the model that cannot classify some parts of the graph correctly due to noisy links. 

We  developed an intuitive algorithm to automatically highlights which parts of the visualization will likely contain prediction errors, \Cref{error_est}. The core part of this algorithm is to perform a few steps of random walk starting from nodes with known labels. Predictions that are close to training data with the same labels in the Reeb net will have higher scores in the column of predicted labels and hence have smaller error estimates. 

Applying this algorithm can successfully find other places where mistakes will happen (see plot (B) of figure~\ref{swiss_roll_errors}). As a simple comparison, we also include another plot where we directly use model uncertainty (i.e. 1 minus model prediction probability) to estimate errors (see plot (C) of figure~\ref{swiss_roll_errors}). This metric has been previously used to estimate uncertainty of dataset labels~\cite{northcutt2021confidentlearning}. Clearly, GTDA is able to localize model errors much better and has a higher AUC score (0.95 vs 0.87). There always exists other methods~\cite{huang2021combining} that can also give similar error estimations or even correct predicted labels. But explaining why those methods should work or be trusted to a user without background knowledge is a challenge, while our method offers a map-like justification that can give a rough rationale. Moreover, any results from \Cref{error_est} can always be validated and supported through pictures similar to plot (A) of figure~\ref{swiss_roll_errors}. Also, other than finding possible errors, as shown in the following experiments sections, we can often get many other insights about the model and the dataset by checking abnormal areas of GTDA visualization, ranging from labeling issues to strong correlation between model predictions and a particular dataset property. These are explored in subsequent case studies.

\begin{algorithm}[t]
  \caption{$\texttt{error\_estimation}(\hat{G},\mathbb{E},\ell,n,\alpha)$ where $\hat{G}$ and $\mathbb{E}$ is the Reeb net and extra set of edges from algorithm~\ref{GTDA}, $\ell$ are the original predicted labels, $S$ is an integer for the number of steps (10, or 20 were used), and $0<\alpha<1$ (we use $\alpha = 0.5$ in all experiments).}
  \begin{algorithmic}[1]
    \label{error_est}
    \STATE Compute $G^{(R)}$, the projection of the Reeb net back to a graph from \Cref{Reeb_projection}. 
    \STATE Let $\mA^{(R)}$ be the adjacency matrix of $G^{(R)}$
    \STATE Compute a diagonal matrix $\mD^{(R)}$ where ${\mD}^{(R)}_{ii}$ is the degree of node $i$ in ${G}^{(R)}$ and $0$ elsewhere.
    \STATE Initialize matrix $\hat{P}^{(0)}$ where $\hat{P}^{(0)}_{ij}=1$ iff node $i$ is a training node with label $j$, otherwise $\hat{P}^{(0)}_{ij}=0$.
    \FOR {$i=1...S$}
    \STATE $\hat{\mP}^{(i)}=(1-\alpha)\hat{\mP}^{(0)}+\alpha{\mD^{(R)}}^{-1} \mA^{(R)} \hat{\mP}^{(i-1)}$
    \ENDFOR
    \STATE Row normalize $\hat{\mP}^{(S)}$ so that each row sums to $1$.
    \STATE Compute estimated prediction error for node $i$ to be $e_i=1-\hat{\mP}^{(S)}[i,\ell_i]$
    \STATE Return estimated errors $\ve$.
  \end{algorithmic}
\end{algorithm}

\begin{algorithm}[h!]
  \caption{$\texttt{Reeb-graph-projection}(\mathbb{F},\mathbb{E},G)$ where $\mathbb{F}$, $\mathbb{E}$ is the final set of components and extra set of edges from \Cref{GTDA} and $G$ is the original graph} 
  \begin{algorithmic}[1]
  \label{Reeb_projection}
  \STATE Initialize $G^{(R)}$ with the same dimension of $G$ and no edges
  \FOR{Each $\mathbb{F}_i$ of $\mathbb{F}$}
  \STATE Add the set of edges of $\mathbb{F}_i$ from $G$ to $G^{(R)}$
  \ENDFOR
  \STATE Add edges in $\mathbb{E}$ to $G^{(R)}$
  \STATE Return $G^{(R)}$
\end{algorithmic}
\end{algorithm}

\subsection{Reeb graph vs.~Reeb space vs.~Reeb network}
\label{sec:reeb-space}
\begin{figure}[tp]
    \centering
    \includegraphics[width=\linewidth]{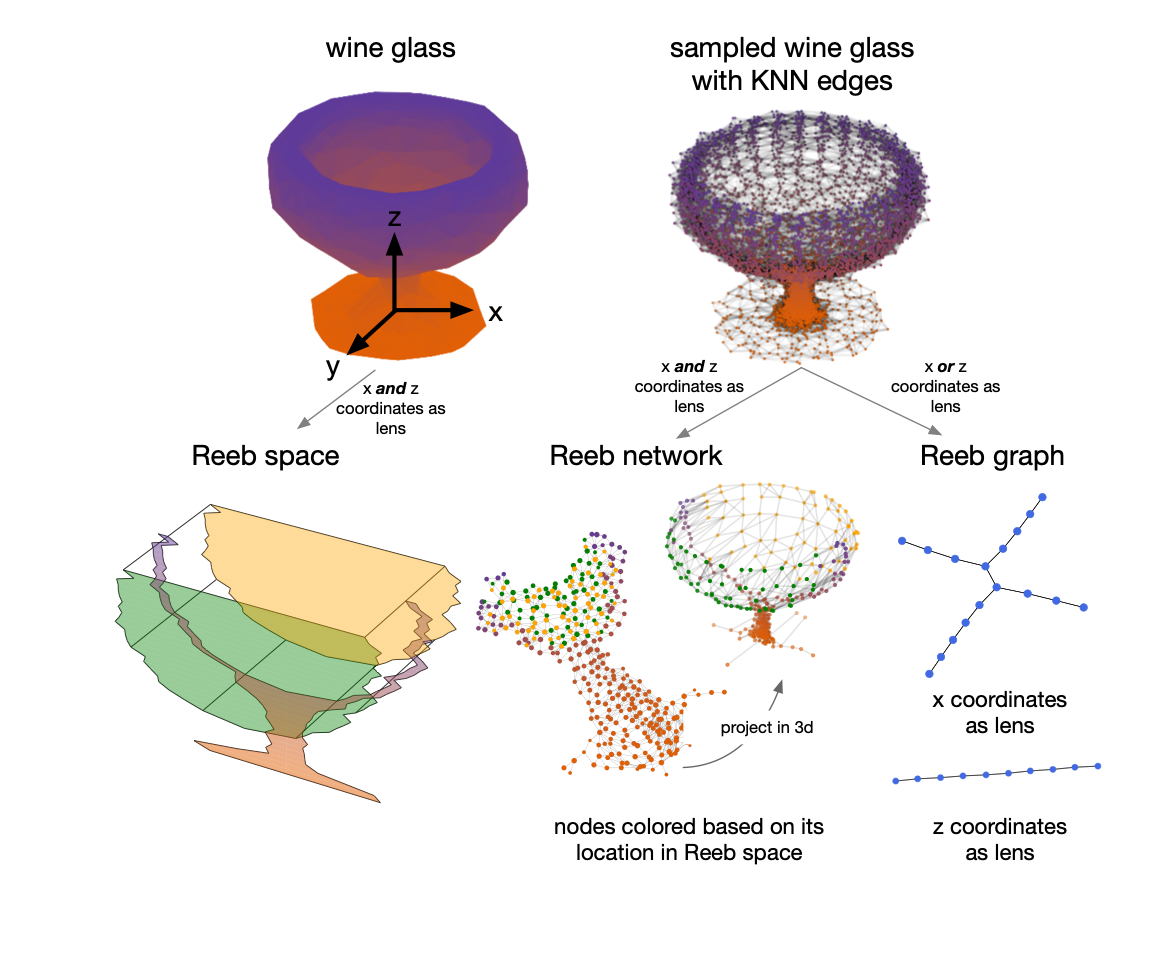}
    \caption{This illustrates the difference between a Reeb graph and a Reeb network on a topologically interesting object. The lenses we use here are the $x$ and $z$ coordinates. The inspiration for the object is~\cite{STRODTHOFF2015186}.  }
    \label{fig:reeb-space}
\end{figure}

The main difference between a Reeb graph and Reeb network is the number of lenses used because the Reeb net involves a multivalued map which can be thought of as a collection of single valued maps. A demonstration to show this difference can be found in \Cref{fig:reeb-space}.

Formally, let $f: X\rightarrow \mathbb{R}^k$ map a topological space $X$ to a $k$-dimensional real space. Two points $x,y\in X$ are called equivalent if (i) $f(x)=f(y)$ and (ii) they belong to the same connected component of the level set $f^{-1}(f(x))$. Denoting this equivalence relation with $\sim$, we obtain the quotient space $R_X^f=X/\sim$. When the range of $f$ is $\mathbb{R}$,  $R_X^f$ is a one-dimensional space called the Reeb graph of $f$. When $f$ is multi-valued, that is, $k>1$, $R_X^f$ becomes a space called Reeb space. By choosing the bins in $\mathbb{R}^k$, we discretize this Reeb space with a graph which we call the \emph{Reeb net} here. We choose the term Reeb net to distinguish it from discretized Reeb graph because both are graphs but one discretizes a one-dimensional space (a graph) and the other discretizes a quotient space that need not be one-dimensional.

\subsection{Opportunities and extensions of the method}
We presented the GTDA framework for the main methods we used. In the following case studies and demonstrations, we show there are multiple variations that would be easy to adapt. For instance, we could easily combine multiple graphs from different sources to reveal potential errors that might  hidden in a single source. 

\paragraph{Areas for future improvement.} Our current GTDA framework does rely on some tuning of parameters and manually finding any interesting local structures in the visualization, especially the component size threshold, which behaves similarly to bin size in the original TDA algorithm. While we designed the algorithm to be as robust as possible, it remains an open question on whether we can automatically select a good set of parameters and identify structures worth looking at. Existing work selects parameters for the original TDA framework based on statistical analysis~\cite{carriere2018statistical}. But it is not clear how to extend such technique to our GTDA framework. 

\paragraph{Areas for additional topology.} Another direction is to study the outputs of GTDA under perturbations or filtrations over parameters. Alternatively, there are opportunities to utilize additional topological insights to improve the graph drawing. Consider that a study of persistence of structures in the graph should suggest their placement, i.e.~two components that will be connected more easily by perturbing parameters should be put closer. This can then lead to a better overall view of the entire dataset.

\subsection{Other details}\label{sec:intro_details}
\paragraph{Swiss Roll dataset construction} We use \emph{scikit-learn} package to build the Swiss Roll dataset. We use 1000 samples in total and the noise parameter is set to be 1.2. The initial Swiss Roll dataset is a 1000-by-3 matrix $X$ and a vector $y$ which represents the position of each sample in the main dimension of the manifold. We only keep the first and the third columns of $X$ and use them as node features. And we sort samples based on $y$ and consider the first $33\%$ samples as the first class, the second $33\%$ samples as the second class and all the other samples as the third class. The graph is a nearest neighbor graph with each node connecting to its 5 closest neighbors using Euclidean metric on $X$. We use a random set of use 10\% samples as training, another 10\% samples as validation and all the other points as testing. 

\paragraph{Model and parameters} We use a standard 2-layer GCN model to predict labels of testing samples. The dimension of the hidden layer is 64, learning rate is 0.01 and weight decay is $10^{-5}$. Once the model is trained, we use outputs of the first layer as node embeddings. The embedding matrix is reduced to 16 dimension using PCA with whitening and then each row is $\ell_2$ normalized. We build another 2-NN graph using the preprocessed embedding matrix and cosine similarity to encode any information from node features. This graph is combined with the original graph. GTDA framework is then applied on the combined graph. For GTDA parameters, we set $K=20$, $d=0$, $r=0.1$, $s_1=5$, $s_2=5$, $\alpha=0.5$ and $S=5$. We use 10 steps of iterations for GTDA error estimation.

\paragraph{Alternative neural networks.}
We note that we saw similar results using the GNN methods from~\cite{chien2021adaptive}. We  include discussions and images with this alternative method for the Amazon dataset (next section) to evaluate our statement from the main text about the taxonomy.

%% file: exp_graph.tex
\label{sec:amazon-advanced} 

In this section, we provide more details for the application of our GTDA framework on an Amazon co-purchase graph~\cite{shchur2018pitfalls} constructed from Amazon reviews data~\cite{mcauley2015image}. Each node in this graph is a product, edges connect products that are purchased together and node features are bag-of-words from product reviews. The goal is to predict product category. The original dataset~\cite{shchur2018pitfalls} that has been used in several GNN papers does not have information for each specific product. To better understand the visualization from GTDA, we build a similar dataset directly from the Amazon reviews data~\cite{mcauley2015image}. We use the 2014 year version of reviews data and extract products with the same set of labels as in the original~\cite{shchur2018pitfalls}. In the remainder of this section, we will work through how the dataset is constructed and the GCN model parameters that are used in the main text. We will also provide GTDA results on another more recent GNN model, GPRGNN~\cite{chien2021adaptive}, that is based on spectral theory. We will inspect this model's prediction on both the customized dataset and the original dataset. We will see later in this section that the same conclusions as Figure 2 of the main text still hold even after switching to the new model.

\subsection{Dataset and GNN model} 
Our own version of the Amazon co-purchase graph has the same set of the labels as the original one~\cite{shchur2018pitfalls}. We download all products and reviews in the category of ``Electronics'' by following the link provided in~\cite{mcauley2015image}. We use the 2014 version as we can find the exact same set of labels in this version. In the Amazon reviews dataset, each product is associated with a list of categories. To assign labels, for each product, we check from the most general category (i.e.\ Electronics) to the most specific one (i.e.\ Routers). And if we find a match to the set of labels we choose, we directly assign the matched label to that product and ignore the other categories in the list. Two products will be connected if they are marked as ``also bought'', ``bought together'' or ``buy after viewing''. After we get the initial graph, we first make the graph undirected and then filter out components that are smaller than 100. We use bag-of-words node features with TF-IDF term weighting constructed from each product's review text. The final graph we get has 39,747 products and 798,820 edges. The number of products for each category is listed in table~\ref{tab:electronics_stats}.

\begin{table}[t]
\centering
\begin{tabular}{cc}
\toprule
category & number \\
\midrule
Desktops & 1,757 \\
Data Storage & 7,297 \\
Laptops & 4,590 \\
Monitors & 1,710 \\
Computer Components & 15,167 \\
Video Projectors & 804 \\
Routers & 1,086 \\
Tablets & 1,919 \\
Networking Products & 4,869 \\
Webcams & 548 \\
\bottomrule
\end{tabular}
\caption{Number of products for each category in our own version of Amazon Computers dataset.}
\label{tab:electronics_stats}
\end{table}

To get the prediction results used in Figure 2, we use the same 2-layer GCN model as the Swiss Roll experiment to predict product categories (Section~\ref{sec:intro_details}). The dimension of the hidden layer is 64, learning rate is 0.01 and weight decay is $10^{-5}$. We randomly use 10\% samples as training, another 10\% samples as validation and all the other samples as testing. We extract the output of the first layer as node embeddings and we also build a 2-NN graph using cosine similarity to combine with the original graph. This will let GTDA show the impact of the feature similarity on the GNN.  For GTDA parameters, we set $K=100$, $d=0$, $r=0.01$, $s_1=5$, $s_2=5$, $\alpha=0.5$ and $S=5$. We use 20 steps of iterations for GTDA error estimation. For the more advanced GPRGNN model used below, we use the same set of parameters as suggested by its authors~\cite{chien2021adaptive} and node embeddings are extracted from the first layer output as well. We also use the same GTDA parameters as GCN.

\subsection{Inspecting model predictions with GTDA} 
\begin{figure}[p]
  \centering
  \includegraphics[width=\linewidth]{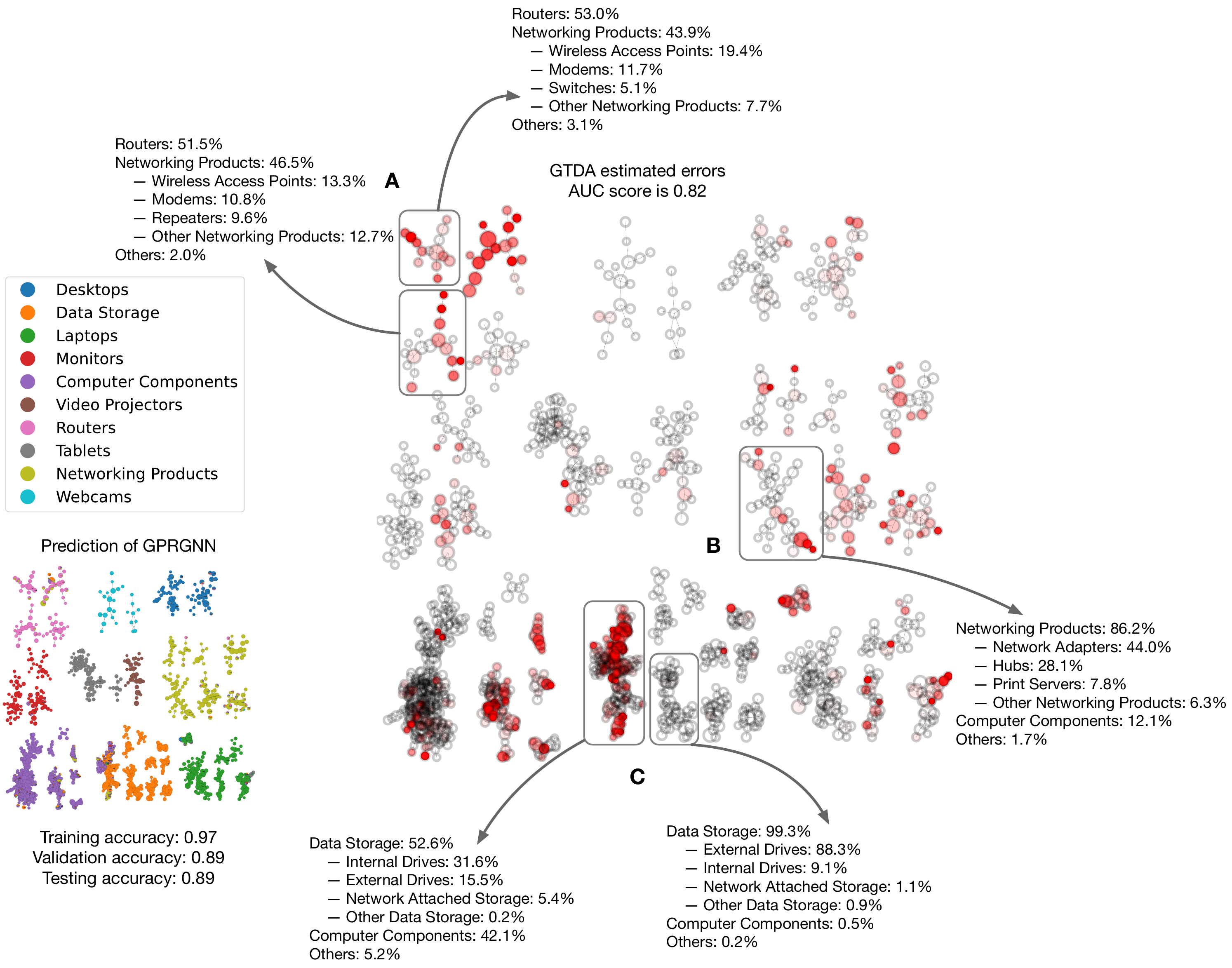}
  \caption{We provide GTDA results on inspecting the prediction on the GPRGNN method instead of the GCN used in Figure~\ref{fig:amazon} in the main text. We list a detailed breakdown of categories and subcategories for a few components. For the two ``Routers'' components in (A), there are many estimated errors because of ambiguous subgroups of ``Networking Products'' like ``Wireless Access Points'', ``Modems'' or ``Repeaters''. The estimated errors are much less in (B) because ``Networking Products'' has dominant less ambiguous subgroups. Similarly, for two ``Data Storage'' components in (C), the one with more estimated errors has dominant ambiguous subgraphs like ``Internal Drives'' or ``Network Attached Storage'' which is confusing with ``Computer Components'' or ``Networking Products''.
  }
\label{amazon_analysis_detailed}
\end{figure}

In Figure 2 of the main text, we found ambiguous subgroups inside ``Data Storage'' and ``Networking Products'' with the help of GTDA visualization. Similar ambiguities persist after switching to the more advanced GPRGNN model as shown in \Cref{amazon_analysis_detailed}. Here, we also notice many estimated errors in ``Routers'' and ``Data Storage'' as before. 
We show a detailed breakdown of products true categories for some components. 
For each component highlighted, we list top 2 most common categories. 
The other categories are put in ``Others''. 
For ``Networking Products'' and ``Data Storage'', we also list the top 3 most common subcategories. 
For the two ``Routers'' components in (A), we see many ``Modems'' or ``Wireless Access Points'' from ``Networking Products''. These should be frequently bought together, and ``Routers'' should be considered as another subcategory of ``Networking Products''. 
As a comparison, for the other ``Networking Products'' component that is less mixed (B), the most common subcategories are ``Network Adapters'' and ``Hubs'', which are more precise than the more ambiguous ``Routers''. 
Similarly, for the two ``Data Storage'' components in (C), the mixed one has many ``Internal Drives'' such as solid state drives (SSDs). These are essential parts of a PC and should be considered as a part of ``Computer Components'' as well. 
There are also a small portion of ``Network Attached Storage'', which may be confused with ``Networking Products''. 
On the contrary, the less mixed one mostly contains ``External Drives'' like USB drives, which are common additions to an already built PC. These results suggest that for this dataset, no matter which model we choose, the performance on some portion of the dataset will always be limited by the same type of underlying labeling issues. 
GTDA helps capture those issues in both cases.

\subsection{GTDA visualization on the original Amazon dataset} 
\begin{figure}[t]
  \centering
  \begin{minipage}{0.26\linewidth}
    \includegraphics[width=\linewidth]{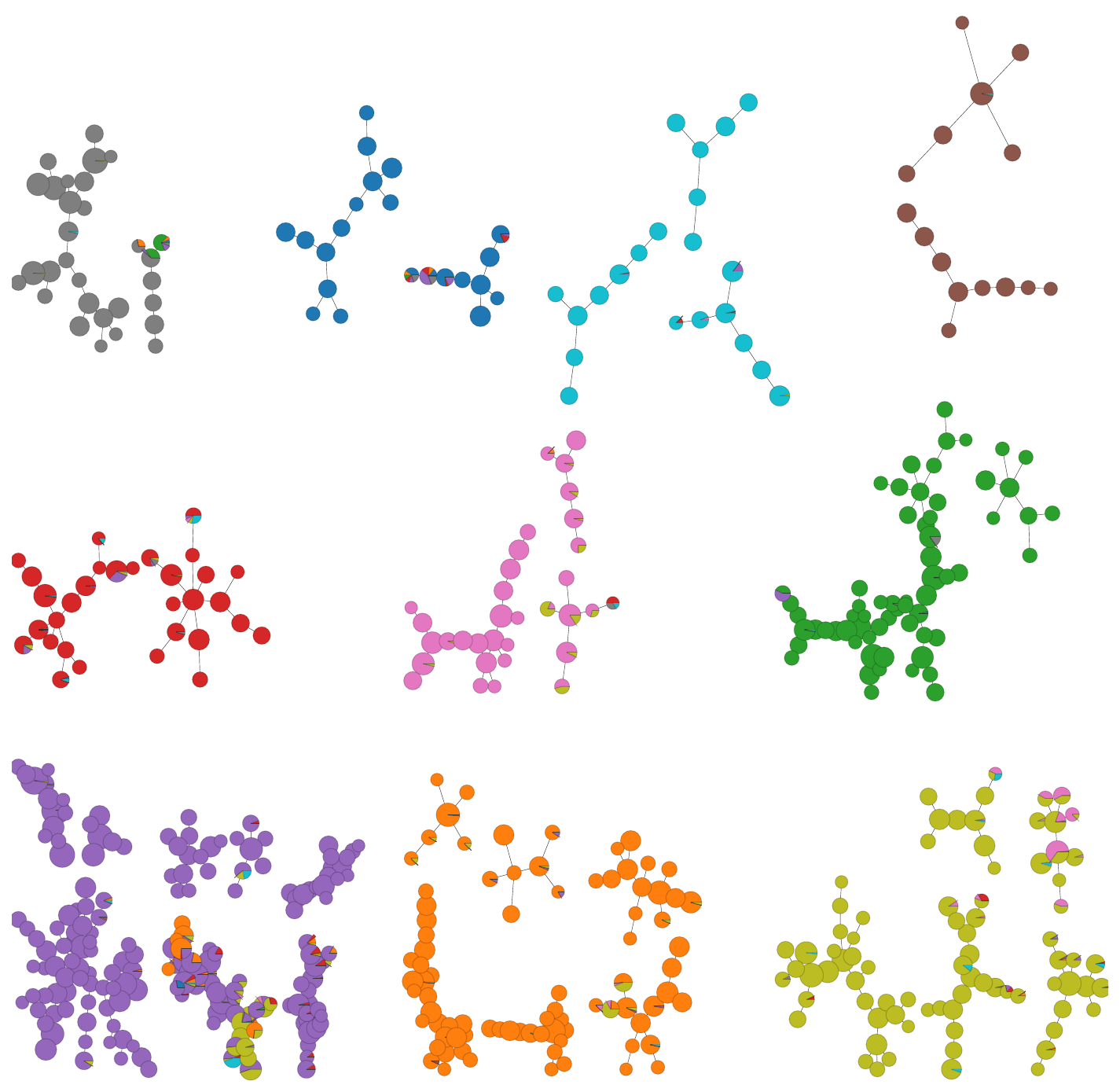}
    \footnotesize (1) predicted labels
  \end{minipage}\hfill
  \begin{minipage}{0.26\linewidth}
    \includegraphics[width=\linewidth]{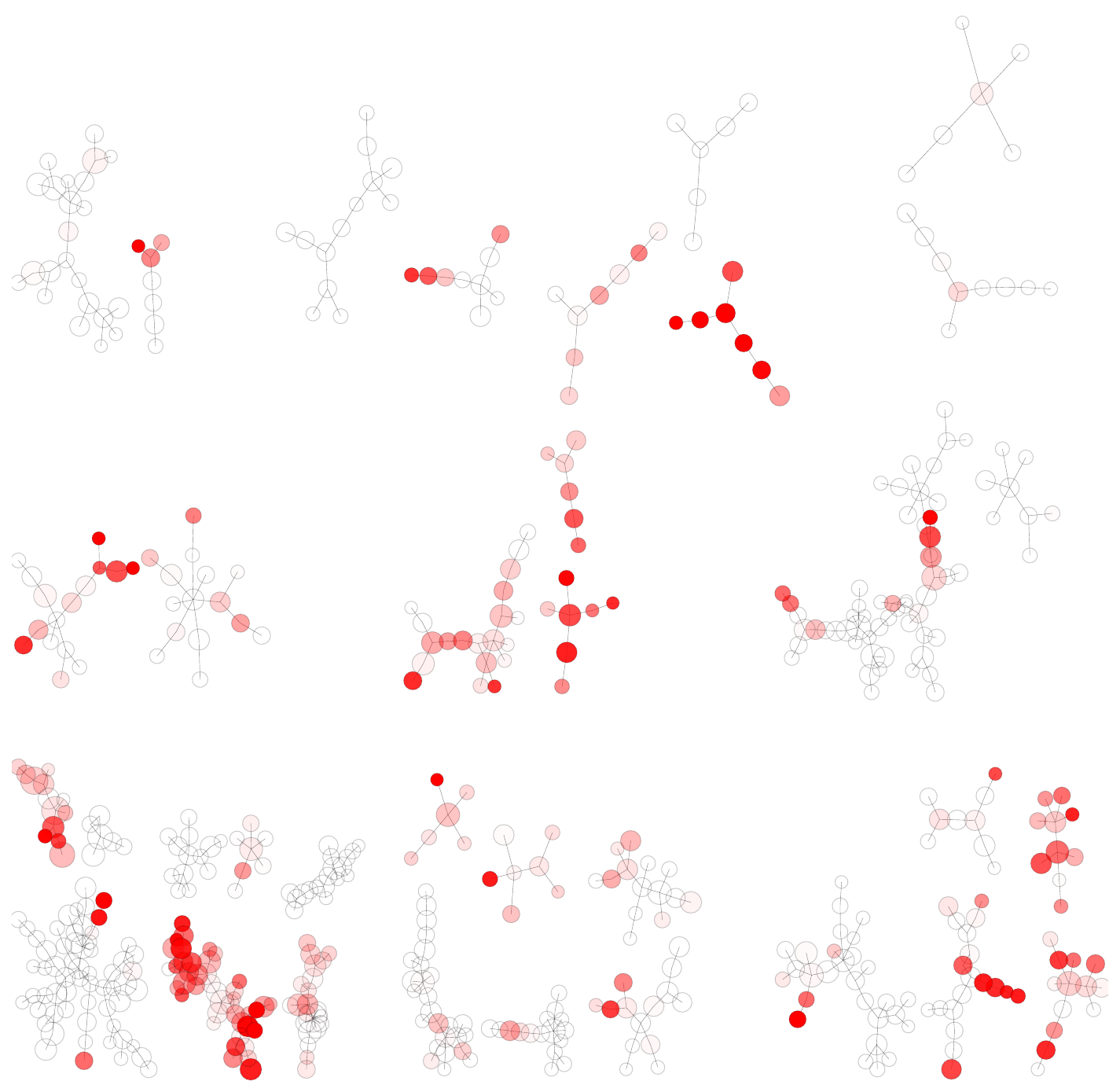}
    \footnotesize (2) GTDA estimated errors
  \end{minipage}
  \begin{minipage}{0.26\linewidth}
    \includegraphics[width=\linewidth]{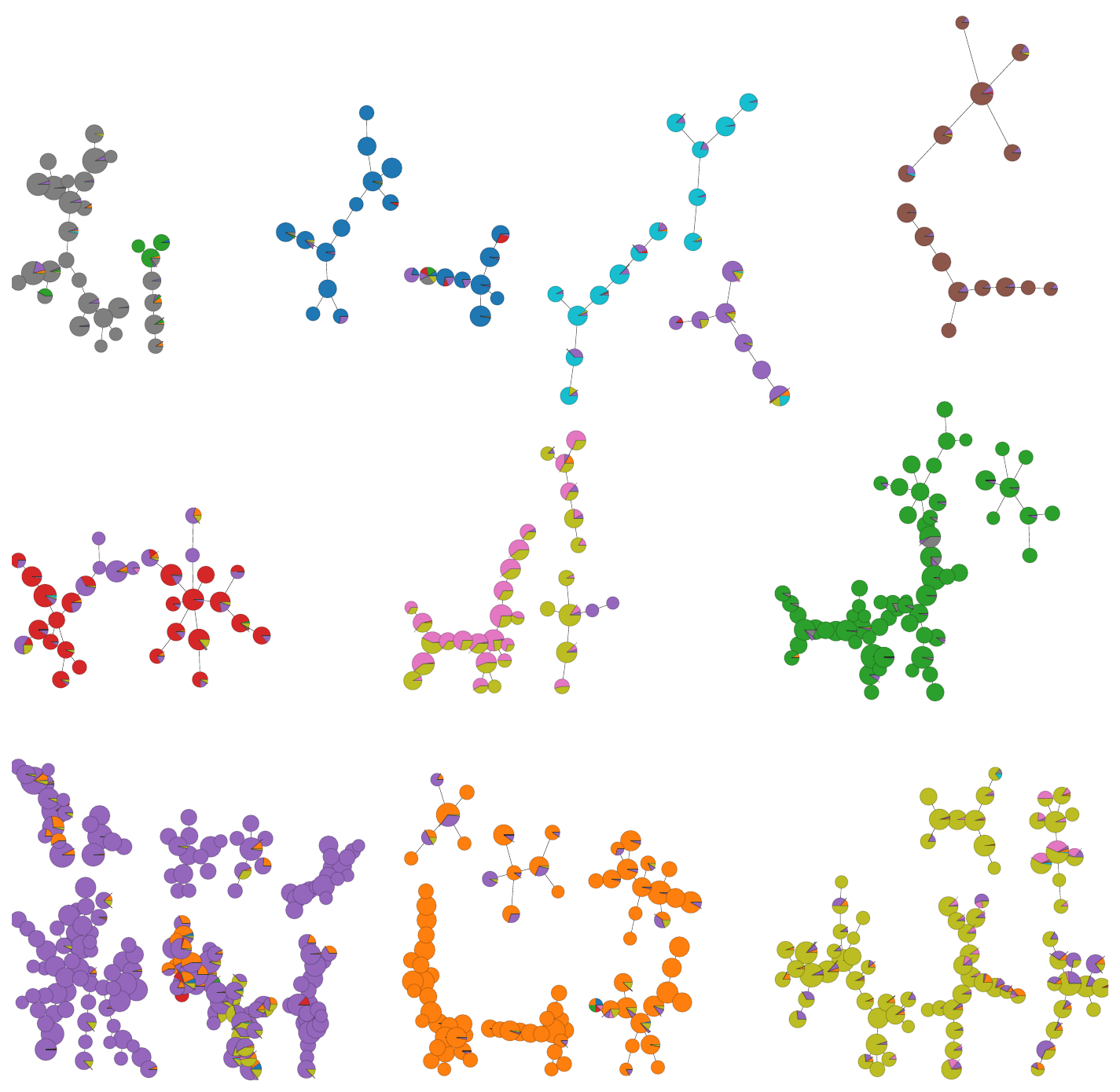}
    \footnotesize (3) true labels
  \end{minipage}
  \begin{minipage}{0.18\linewidth}
    \includegraphics[width=\linewidth]{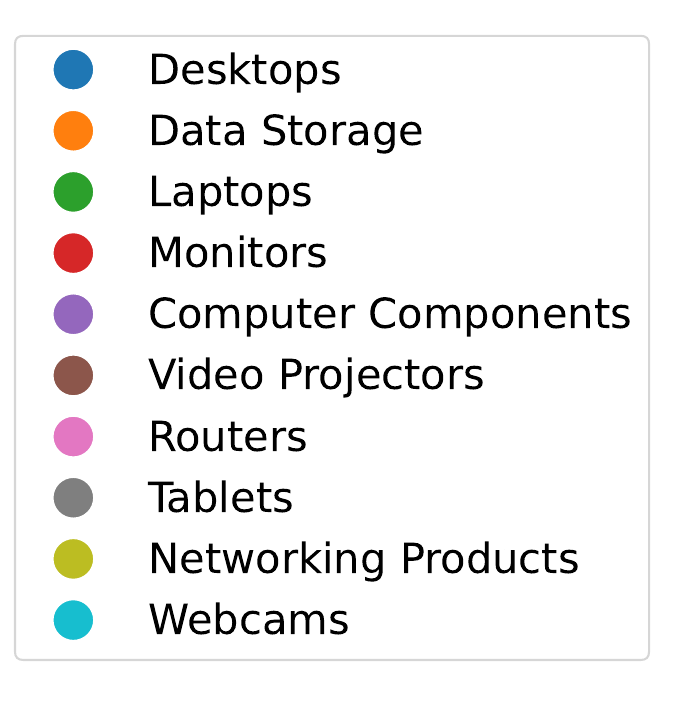}
  \end{minipage}
  \caption{GTDA visualization of GPRGNN's prediction on the original Amazon Computers dataset~\cite{shchur2018pitfalls}. Similar to Figure~\ref{amazon_analysis_detailed}, ``Routers'' is mixed with ``Networking Products'' and some components of ``Data Storage'' are mixed with ``Computer Components''.}
\label{orig_amazon_results}
\end{figure}

As a final check on our results, in \Cref{orig_amazon_results}, we apply GTDA to inspect GPRGNN's prediction on the original Amazon dataset built by~\cite{shchur2018pitfalls} with the same setting. We can observe similar behavior to \Cref{amazon_analysis_detailed}, that is ``Routers'' is mixed with ``Networking Products'' and components of ``Data Storage'' are mixed with ``Computer Components''. 

%% file: exp_image.tex
One of the most successful applications for complex neural network models is detecting objects in images. Image classifiers based on convolutional neural networks (CNN) can achieve extremely high accuracy, sometimes even higher than humans. What remains not entirely understood is how to explain a model's prediction and whether it will generalize well beyond the training scenario. 

\paragraph{Summary of GTDA results}
In Figure 3 of the main text, we have shown how we can use GTDA visualization to study predictions made by a pretrained ResNet50 classifier on a subset of ImageNet called Imagenette. The Reeb net from GTDA highlights ``cassette players'' images that are really pictures of cars inside the ``gas pump'' group. This is a key difference from the Amazon experiment in the previous section, because only a few samples inside ``cassette player'' have labeling issue. 
In the following, we will provide more details on the dataset and the CNN model. Then we will use a random experiments to show that GTDA is stable in detecting this issue and the criteria we use to find such issue cannot be easily satisfied in random set of images. We will show the ResNet50 model's generalization ability by embedding images on some other components of GTDA results. Finally, we compare with results of the original \emph{mapper}.

\paragraph{Displaying Reeb networks for images.}
Because each image can be displayed, we customize a display of a Reeb net (which is simply a graph) to show the results of a Reeb net analysis by placing images directly on the layout. This involves a few relevant details that may assist others in the future so we detail our methodology here. It was inspired by Tufte's work on image quilts and small multiples~\cite{tufte}.

\paragraph{Prior work on understanding image predictions.}
Existing research seeks to explain model predictions by computing activation maps or saliency maps~\cite{springenberg2014striving,zhou2016learning,selvaraju2017grad,simonyan2013deep}. In these maps, areas that contribute to the final prediction will be highlighted and the user can justify model predictions by checking whether the areas highlighted make sense. Some other studies take a different approach by training a simple and explainable model (i.e.~a linear classifier) to mimic the prediction functions of the original model~\cite{ribeiro2016should}. However, all these approaches can only explain the model's prediction on a single sample each time instead of model's prediction ability in the entire dataset. The training and testing datasets can contain hundreds of thousands of images. So examining the explanation for all images is not straightforward. Finding representative samples is another alternative~\cite{ribeiro2016should}, but checking explanation on each selected sample is still required. We note that our GTDA analysis could assist such efforts by studying the topology of the saliency maps, along with the predictions, although we have not pursued this direction.

\subsection{Dataset and CNN model} The dataset we use is Imagenette~\cite{imagenette}, which is a subset of the entire ImageNet containing 10 easily classified classes, ``tench'' (a type of fish), ``English springer'' (a type of dog), ``cassette player'', ``chain saw'', ``church'', ``French horn'', ``garbage truck'', ``gas pump'', ``golf ball'' and ``parachute''. This dataset can be directly downloaded from a Github repository~\cite{imagenette}. The author uses a different training and testing split from the original ImageNet dataset so we first restore the original split before model training. This choice is because the pretrained model from the full ImageNet dataset may have had access to images in the Imagenette test set.  The number of training and testing images for each class is shown in table~\ref{tab:imagenet_stats}.
\begin{table}[tp]
\centering
\begin{tabular}{ccc}
\toprule
label & training & testing \\
\midrule
tench & 1,300 & 50 \\
English springer & 1,300 & 50 \\
cassette player & 1,300 & 50 \\
chain saw & 1,194 & 50 \\
church & 1,300 & 50 \\
French horn & 1,300 & 50 \\
garbage truck & 1,300 & 50 \\
gas pump & 1,300 & 50 \\
golf ball & 1,300 & 50 \\
parachute & 1,300 & 50 \\
\bottomrule
\end{tabular}
\caption{Number of training and testing images for each label.}
\label{tab:imagenet_stats}
\end{table}

We use a pretrained ResNet50 model that is included in the PyTorch package and retrain the last fully connected layer to make predictions on these 10 classes only. We use a batch size of 128, learning rate of 0.01 and run for 5 epochs. We also use the common image transform during training and testing. That is, each training image will be randomly cropped into 224-by-224, randomly horizontally flipped and normalized by the mean and standard deviation computed over the entire ImageNet dataset, while each testing image will be resized to 256 along the shorter edge, center cropped to 224-by-224 and then normalized. We modify the pooling of the last convolutional layer from average pooling to maximum pooling and extract its output as node embeddings. Similar techniques are used in the context of image retrivial~\cite{razavian2016visual}. Initially, the embedding dimension is 2048. We first PCA reduce the dimension to 128 with PCA whitening. Then each row is $\ell_2$ noramlized. A 5-NN graph is constructed on the preprocessed embedding matrix with cosine similarity. For GTDA parameters, we set $K=25$, $d=0.001$, $r=0.01$, $s_1=5$, $s_2=5$, $\alpha=0.5$ and $S=10$. We use 10 steps of iterations for GTDA error estimation.

\subsection{Details on selecting images to embed} 
\begin{figure}[tp]
  \centering
    \includegraphics[width=\linewidth]{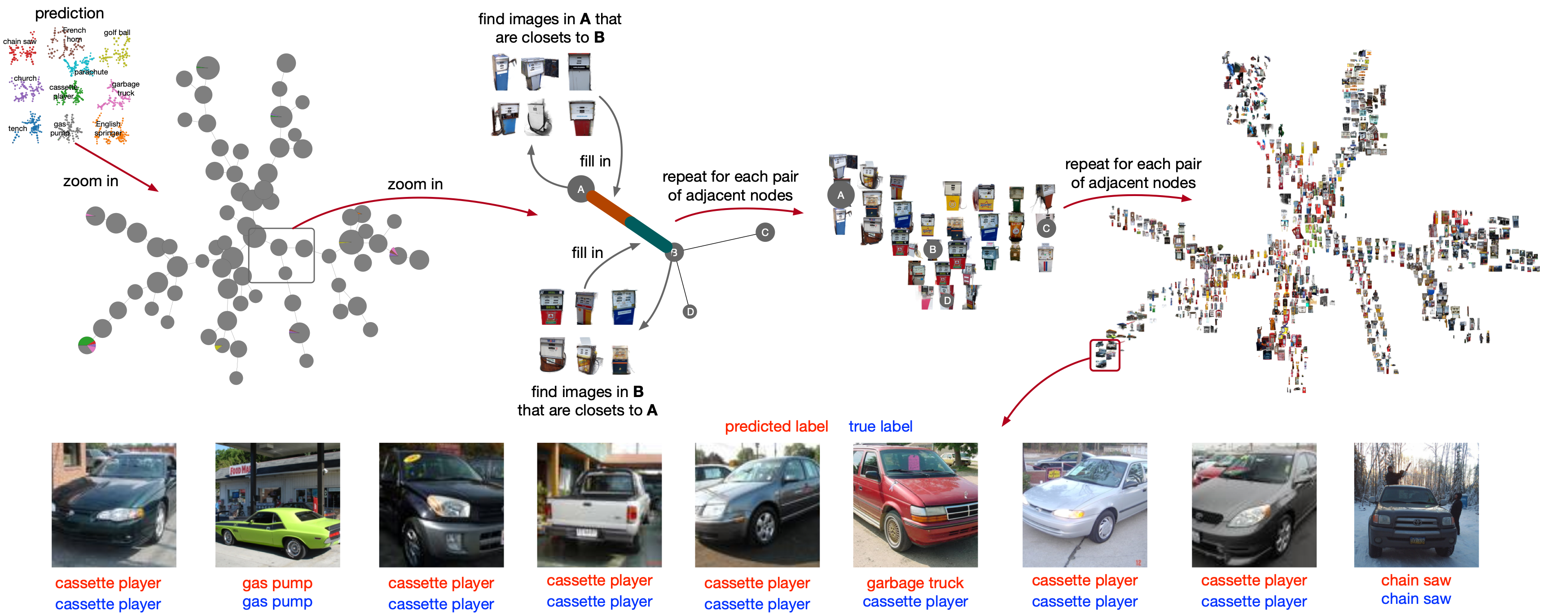}
  \caption{This figure demonstrates the procedure of embedding images on a Reeb net component. For each pair of adjacent nodes, we select images from one end that are closest to the other end and fill in those images in half of the edge and vice versa. Browsing around embedded images at different regions can help us quickly identify 7 ambiguous ``cassette player'' images that are really just ``cars''.}
  \label{gas_pump_detailed}
\end{figure}

We provide more details on how we embed images on a Reeb net component to get Figure 3 of the main text. For each pair of adjacent Reeb net nodes, for each image in one end, we measure its smallest distance in the projected Reeb net to some node in the other end. Note some images can be duplicated in two ends, in such case, we consider the distance to be zero. If two images have the same distance, we include the one with larger degree in the projected Reeb net. Then we fill in the closest images to one half of the edge and vice versa. A simple demo can be found in Figure~\ref{gas_pump_detailed}. We also apply a background removal algorithm~\cite{qin2020u2} for each image we embed. After embedding selected images, we can then easily browse around different regions of the component to understand the model's behaviour of predicting ``gas pumps''. Then we can simply select a few Reeb net nodes at different places and check them in detail by listing all images it contains to look for the most common patterns. Eventually, this can help us quickly identify 7 ambiguous ``cassette player'' images that are really just ``cars''.

\subsection{Statistical validation} 
\label{sec:validate-cassette-tape-vs-car}
Firstly, we verify that GTDA is stable in detecting those 7 confusing ``cassette player'' images as shown in Figure 3 of the main text. We randomly train 100 models in the same way as described before and check the visualization using each of these 100 models. On average, only 1.3 of these 7 images are predicted wrong, which means simply iterating through all the prediction errors is not enough. We define that this labeling issue can be detected in a visualization if the following criteria can be met:
\begin{itemize}
\item All or most of these 7 images are in the same component
\item Some neighbors of these images are from a different class
\item These images are well localized in the component with small pairwise path length
\end{itemize}
In our results, we find the visualization from all 100 models can meet these criteria. More specifically, for 74 models, all 7 images can meet these 3 criteria. In the other 26 models, for 22 of them, 6 images can meet all 3 criteria, for 2 models, 5 images can meet and for the rest 2 models, 4 images can meet. Also the maximum pairwise path length for images meeting the criteria is 4 (for most models, this maximum length is 2). Secondly, we verify that a random group of 7 images will be very unlikely to satisfy these criteria. We pick one of the 100 models and randomly sample 7 images from each Reeb net component. We cannot find any randomly sampled group in 10000 Monte Carlo experiments that can satisfy these criteria simultaneously.

\subsection{Comparing to influence functions}
\begin{figure}[tp]
  \centering
    \includegraphics[width=\linewidth]{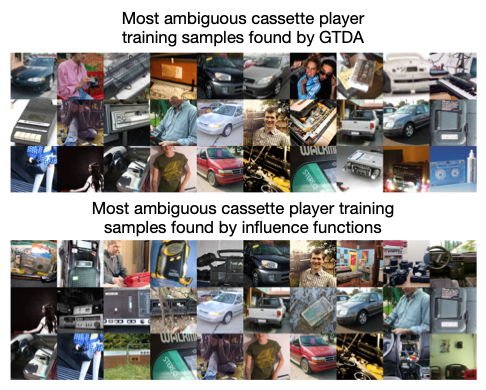}
  \caption{This figure compares the top 30 most confusing training images of ``cassette player'' from influence functions~\cite{koh2017understanding} or GTDA. Both method can find some common training images that are indeed ambiguous. However, it will take influence functions almost 4 hours to compute influence for all 12,894 training images while GTDA only takes about 1 minute to process the entire dataset.}
  \label{influence}
\end{figure}

Influence functions~\cite{koh2017understanding} is a framework recently proposed to extract the most influential training samples on any specific testing sample. It can also be used to find adversarial or mislabeled training data. We used an existing implementation of influence functions from \url{https://github.com/nimarb/pytorch_influence_functions} to find ambiguous training samples of Imagenette. The biggest issue of influence functions is scalability. Computing influence for all 12,894 images will take almost 4 hours while our GTDA framework only takes about 1 minute to process the entire dataset. \Cref{influence} compares the top 30 most confusing training images of ``cassette player'' from influence functions or GTDA. For GTDA, we directly take top 30 images with the largest estimated errors using \Cref{error_est}. Both methods find training images that indeed look confusing. However, another advantage of GTDA is we get more insights by grouping these ambiguous training images based on their locations in the visualization and checking nearby images in the visualization. For instance, we can conclude from \Cref{gas_pump_detailed} that some ``cassette player'' images can be confused with ``gas pump'' or ``chain saw'' images with cars in them.

\subsection{Understanding model generalization on other labels} 
\begin{figure}[tp]
  \centering
    \includegraphics[width=\linewidth]{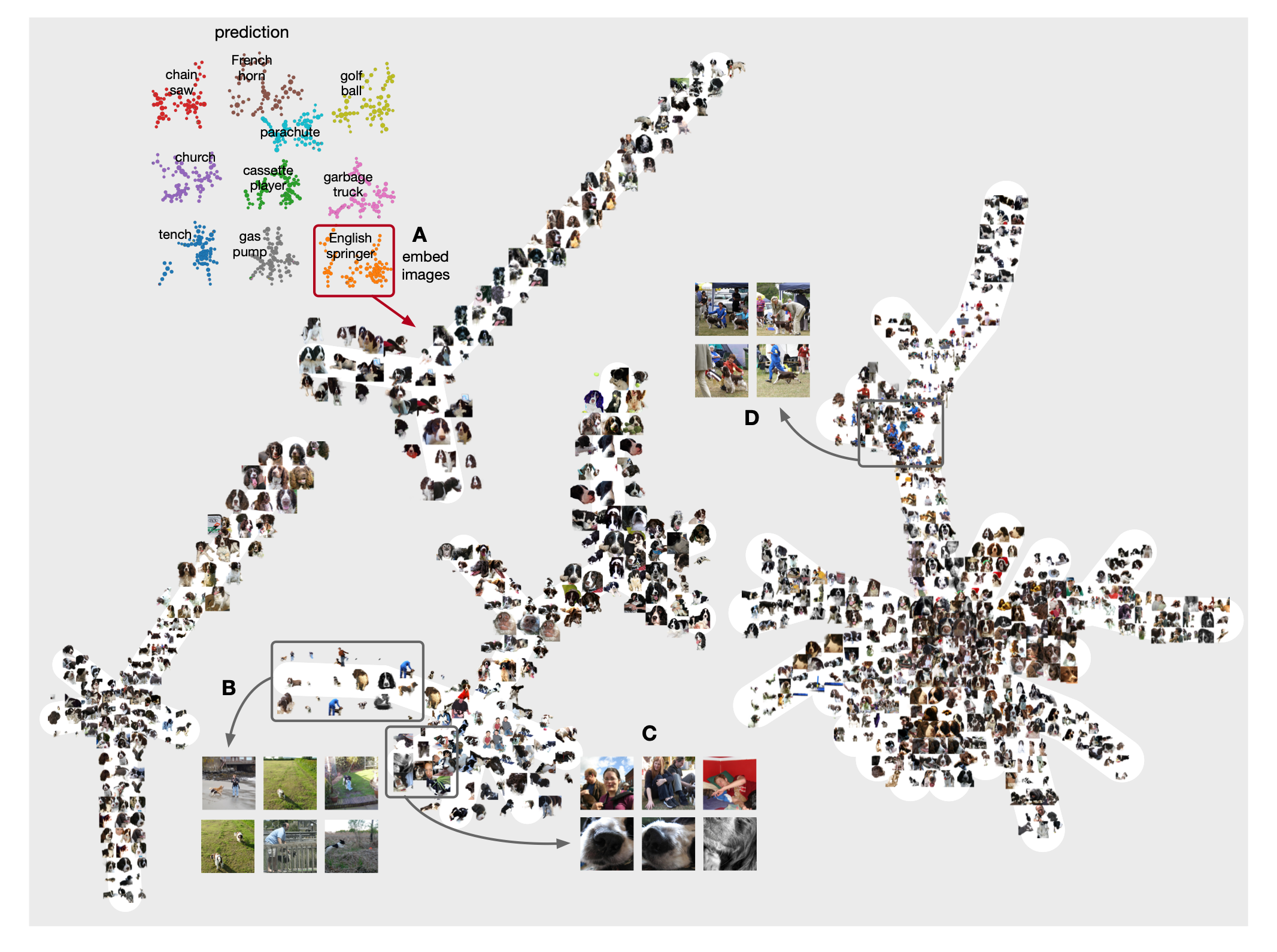}
  \caption{We embed images on components that are mostly ``English Springer'' predicitons (A). While most ``English Springer'' images are easy to classify, we also find some groups where the background information is dominant in (B) and (D) or the images are ambiguous (C). Consider zooming in to see the micropictures.}
  \label{english_springer}
\end{figure}

\begin{figure}[tp]
  \centering
    \includegraphics[width=\linewidth]{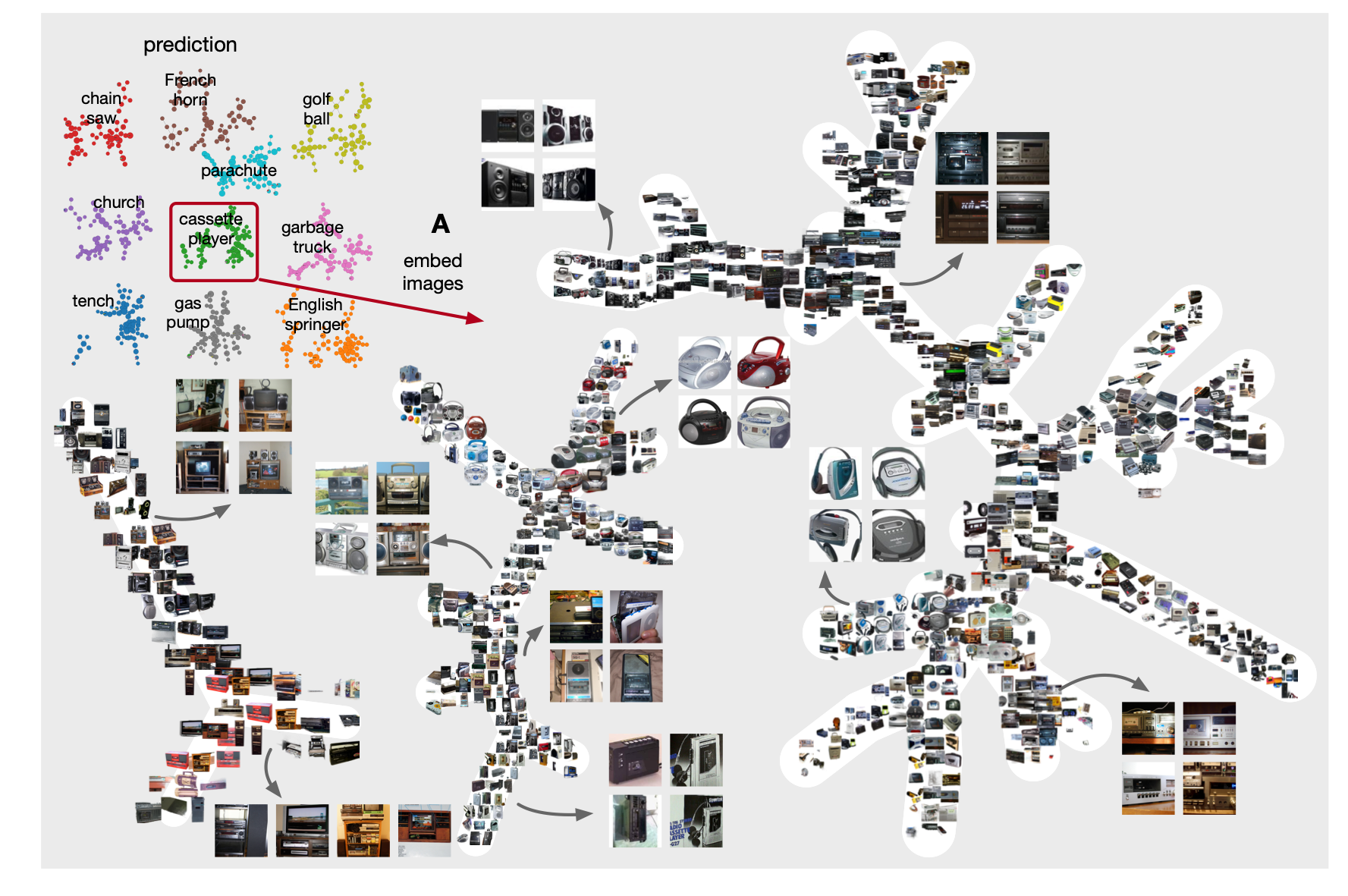}
  \caption{By embedding images on ``cassette player'' components (A) can help us find ``cassette player'' in various shapes.}
  \label{cassette_player}
\end{figure}

\begin{figure}[tp]
  \centering
    \includegraphics[width=\linewidth]{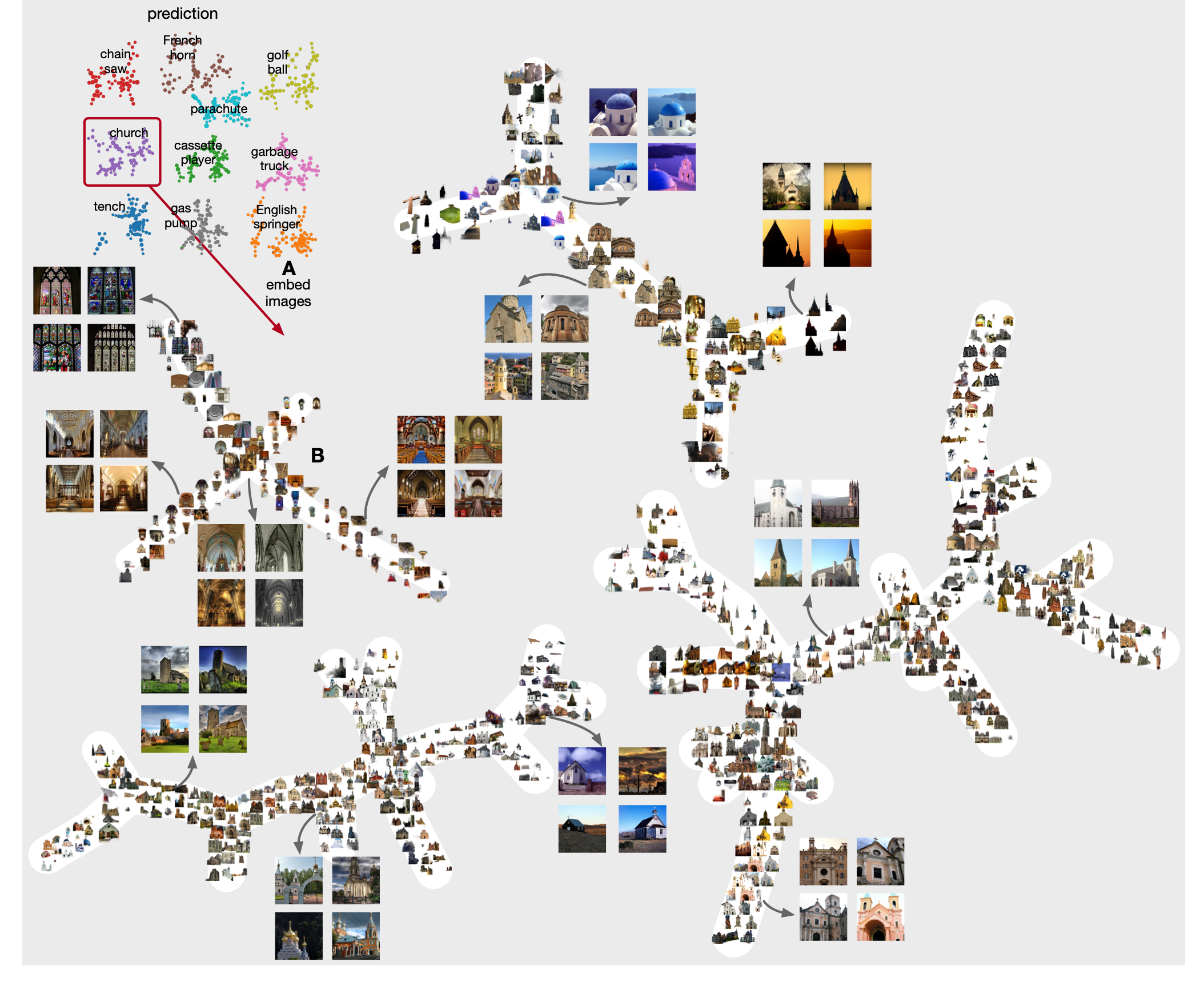}
  \caption{By embedding images on ``church'' components (A), we find one component has images that depicts the inside decorations of church (B) while the other components are images showing different outside landscapes of church.}
  \label{church}
\end{figure}

\begin{figure}[h!]
  \centering
    \includegraphics[width=\linewidth]{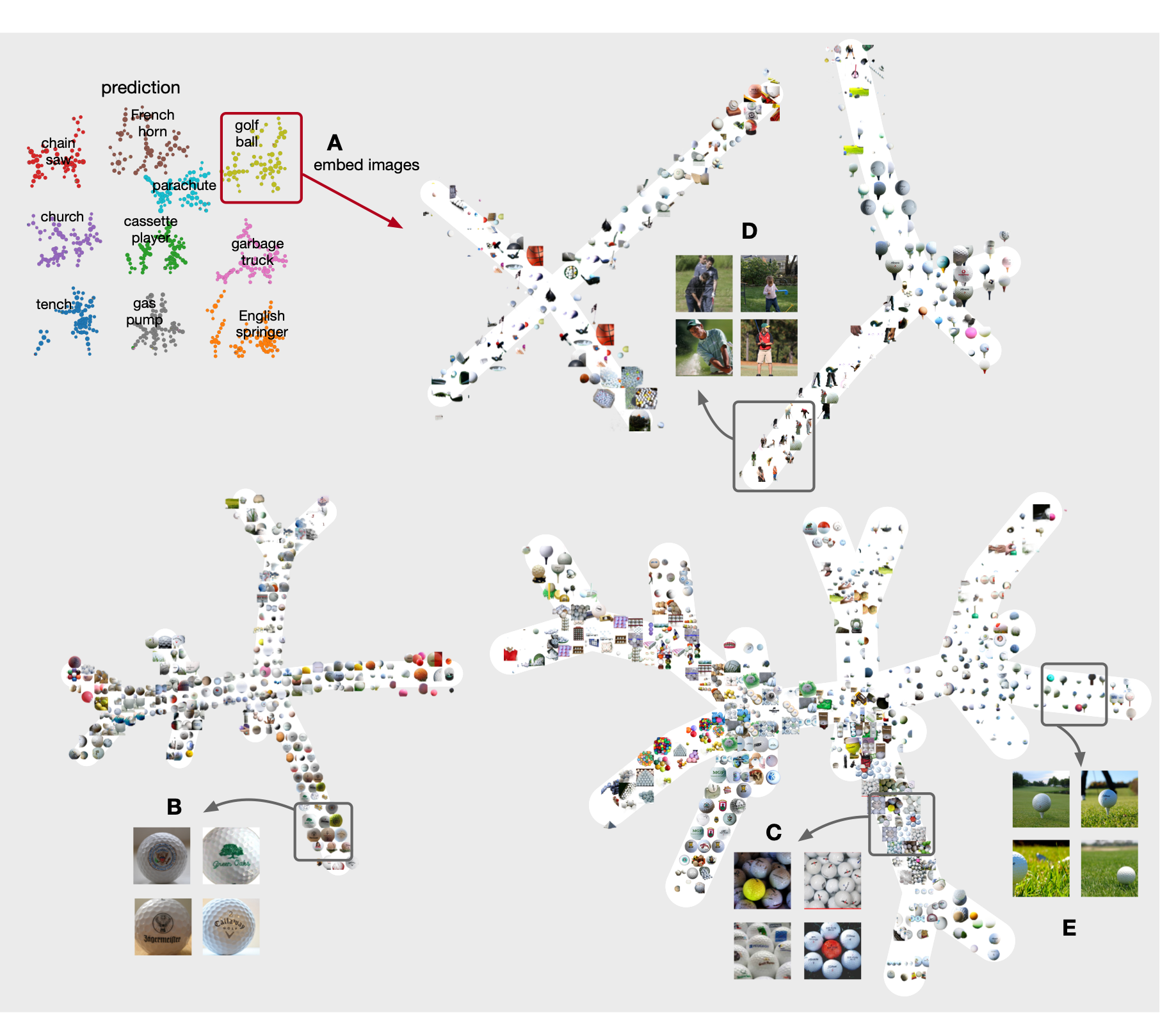}
  \caption{We embed images on ``golf ball'' components (A). We can find images with only one large golf ball (B), or images with lots of small golf balls (C), or images where a person is playing golf ball (D), or images with a golf ball placed on grass (E).}
  \label{golf_ball}
\end{figure}

\begin{figure}[h!]
  \centering
    \includegraphics[width=\linewidth]{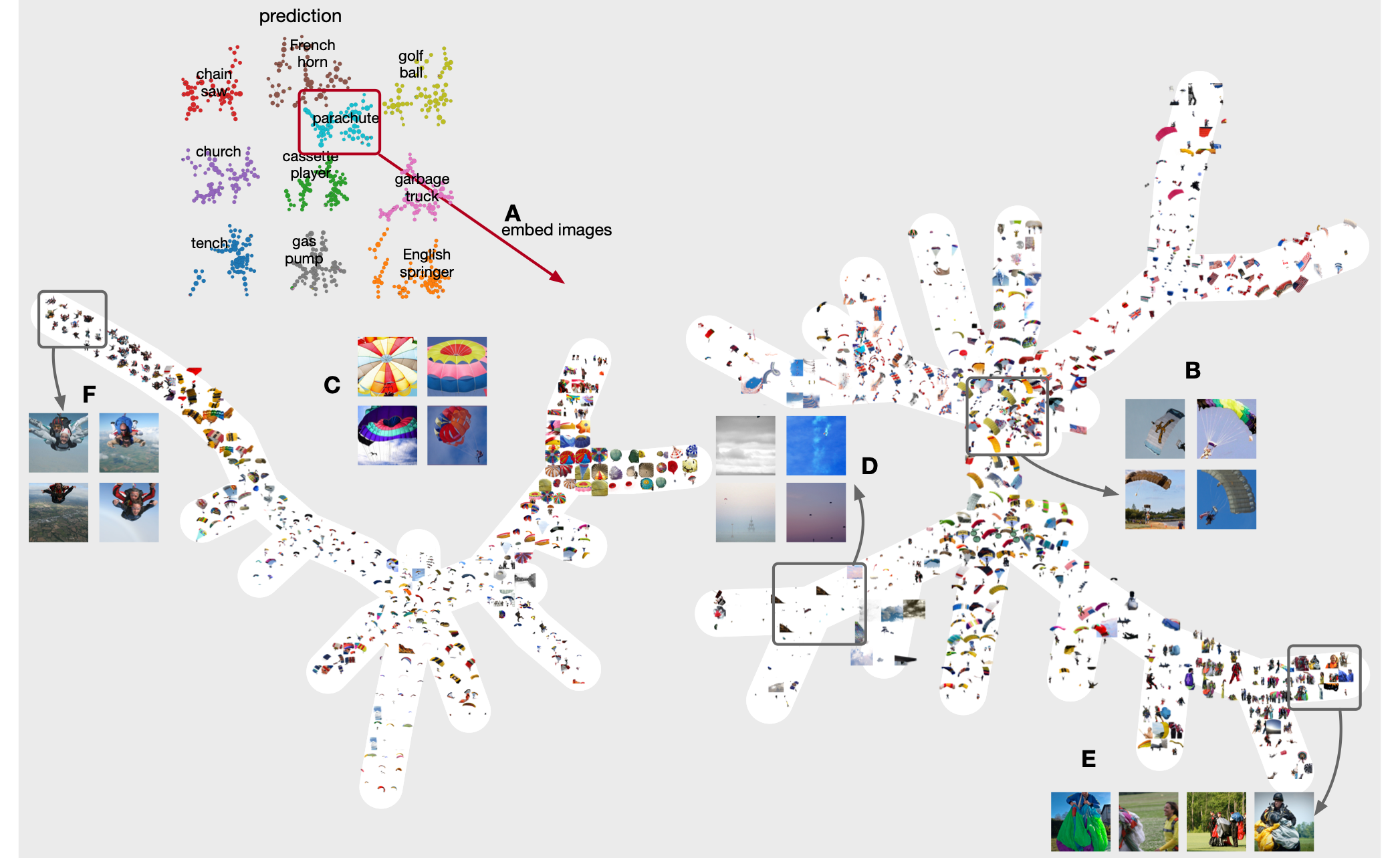}
  \caption{We embed images on ``parachute'' components (A). We can mainly see parachutes in two different shapes (B and C). Some images are ambiguous as they are really just ``sky'' (D). We also find images where a person is standing on the ground wearing a parachute (E) or a person that jumps into the sky (F).}
  \label{parachute}
\end{figure}

Other than the detailed analysis for ``gas pump'' component, we provide similar figures (\Cref{english_springer,cassette_player,church,golf_ball,parachute}) for components of other labels. We embed images on each component in the same way as above. GTDA can always find groups of images with different visual features. For instance, it can find ``church'' images that are either the inside decorations of a church or the outside landscapes in Figure~\ref{church}. It can also find images that are ambiguous like group (C) in \Cref{english_springer} or group (D) in Figure~\ref{parachute}. All these results can help us understand how the model is utilizing different features of an image to make the prediction and when it might make mistakes.


\subsection{Comparing to a Reeb net from original TDA framework}
\label{sec:standard-tda}
\begin{figure}[h!]
  \centering
    \includegraphics[width=\linewidth]{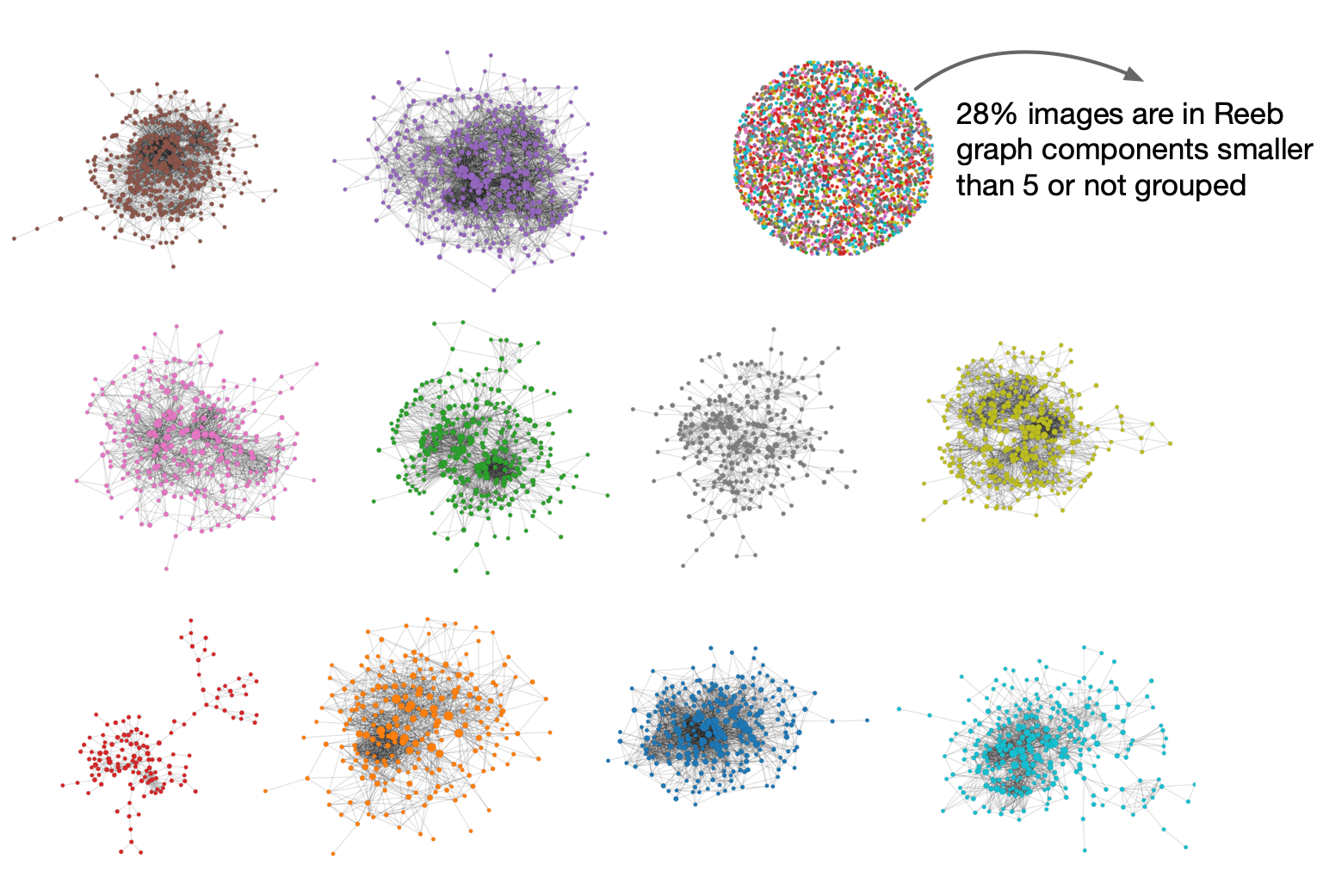}
  \caption{Reeb net on the 10 easy classes of ImageNet created by the original TDA framework. TDA is direclty applied to the ResNet image embedding matrix without transforming into KNN graph. Unlike GTDA visualization, we cannot find any obvious subgroups other than 10 major components representing 10 classes or the labeling issues of some ``cassette player'' images. Moreover, no information can be extracted at all for around 28\% images as they are either in some very small Reeb net components or simply considered as noise by the clustering scheme.}
  \label{tda_img_pred}
  \label{fig:tda_img_pred}
\end{figure}

Since the original format of the image representations is an embedding matrix, we get another Reeb net from the original TDA framework (i.e. \emph{mapper}) without transforming the embedding matrix into a KNN graph. The embedding matrix is still PCA reduced to 128, whitened and $\ell_2$ normalized. We also use the prediciton lens without softmax as the softmax function will make lens highly skewed, i.e. most lens will be close to 0 or 1. We split each len into 10 bins with 10\% overlap. Then we apply density based spatial clustering~\cite{ester1996density} for samples in each bin so that we don't need to select the number of clusters. This clustering scheme will consider some samples as noise and not clustering them. We set the maximum distance between two points to be in the same cluster as 3. The Reeb net is shown in \Cref{tda_img_pred}, which doesn't show any obvious subgroups other than 10 major components representing 10 classes or any labeling issues previously discovered by GTDA. We also find that no information can be extracted at all for around 28\% images as they are either in some very small Reeb net components or simply considered as noise by the clustering scheme.

%% file: exp_image_comparison.tex
\label{sec:comparison}
In this section, we apply GTDA framework on the entire ImageNet dataset with 1000 classes from 2012~\cite{ILSVRC15} to compare performance between state of the art CNN models and historical models in any individual class. The results in the later sections show that GTDA can highlight which subgroups inside a class the more advanced models can have improved performance. It also shows how models predict when the image itself has confusing labels.
\subsection{Dataset and CNN models}
We use the training and validation images of entire ImageNet dataset with 1000 classes that was released in 2012~\cite{ILSVRC15}. We use 3 different CNN models for comparison, AlexNet, ResNet-50 and VOLO. AlexNet is one of the historical CNN models, with around 60\% top-1 testing accuracy. ResNet is one of the most widely used CNN models nowadays with a better performance of about 75\% top-1 testing accuracy. Finally, VOLO is one of the state of art CNN models that achieves about 87\% top-1 testing accuracy without using any additional training data. Then, for each CNN model, we extract the prediction matrix and the image embeddings. For AlexNet and ResNet, the image embeddings are the outputs of the layer before final prediction layer. Similar to the previous section, we replace the average pooling by max pooling in the last convolutional layer. For VOLO, we directly used the dedicated feature forwarding function to get image embeddings. Similar to previous sections, all image embeddings are PCA reduced to 128 with whitening and normalization. For GTDA parameters, we set $K=25$, $d=0.001$, $r=0.01$, $s_1=5$, $s_2=5$, $\alpha=0.5$ and $S=10$. We use 10 steps of iterations for GTDA error estimation.
\subsection{Building graphs and initial results of GTDA} 
We first compare AlexNet and ResNet. To do so, we build a 5-NN graph using the image embeddings of ResNet only. Then we concatenate the prediction matrix of AlexNet and ResNet to get 2,000 lens. GTDA framework is then applied using the same set of parameters as \Cref{sec:imagenette}. Similarly, to compare ResNet and VOLO, we build a 5-NN graph using the image embeddings of VOLO and concatenate the prediction matrix of ResNet and VOLO. In \Cref{tab:imagenet_comparison}, we provide some initial statistics on the final Reeb nets. We can see that despite the Reeb net has tens of thousands of nodes, the maximum Reeb component size is just a few hundred of nodes, which guarantees that we can easily visualize any component of the Reeb net.

\begin{table}[tp]
\centering
\begin{tabular}{ccc}
\toprule
 & AlexNet v.s. ResNet & ResNet v.s. VOLO \\
\midrule
original graph nodes & 1,331,167 & 1,331,167 \\
original graph edges & 5,954,900 & 5,805,714 \\
Reeb nodes & 63,239 & 68,354\\
Reeb edges & 59,881  & 64,360\\
Reeb components & 3,395 & 4,046\\
max Reeb component size & 169 & 79\\
max Reeb node size & 330 & 643\\
average Reeb components for each class & 3.5 & 4.0\\
\bottomrule
\end{tabular}
\caption{Statistics on Reeb nets. Reeb node size is the number of samples represented in a Reeb net node. Average Reeb components for each class is the average number of Reeb net components where the most frequent predicted label (by one of the two models) is that class. The maximum Reeb component just has a few hundred of nodes, which guarantees that any component of the Reeb net can be easily visualized and analyzed.}
\label{tab:imagenet_comparison}
\end{table}

\subsection{Highlighting subgroups where advanced models perform better}
\begin{figure}[tp]
    \centering
    \includegraphics[width=\linewidth]{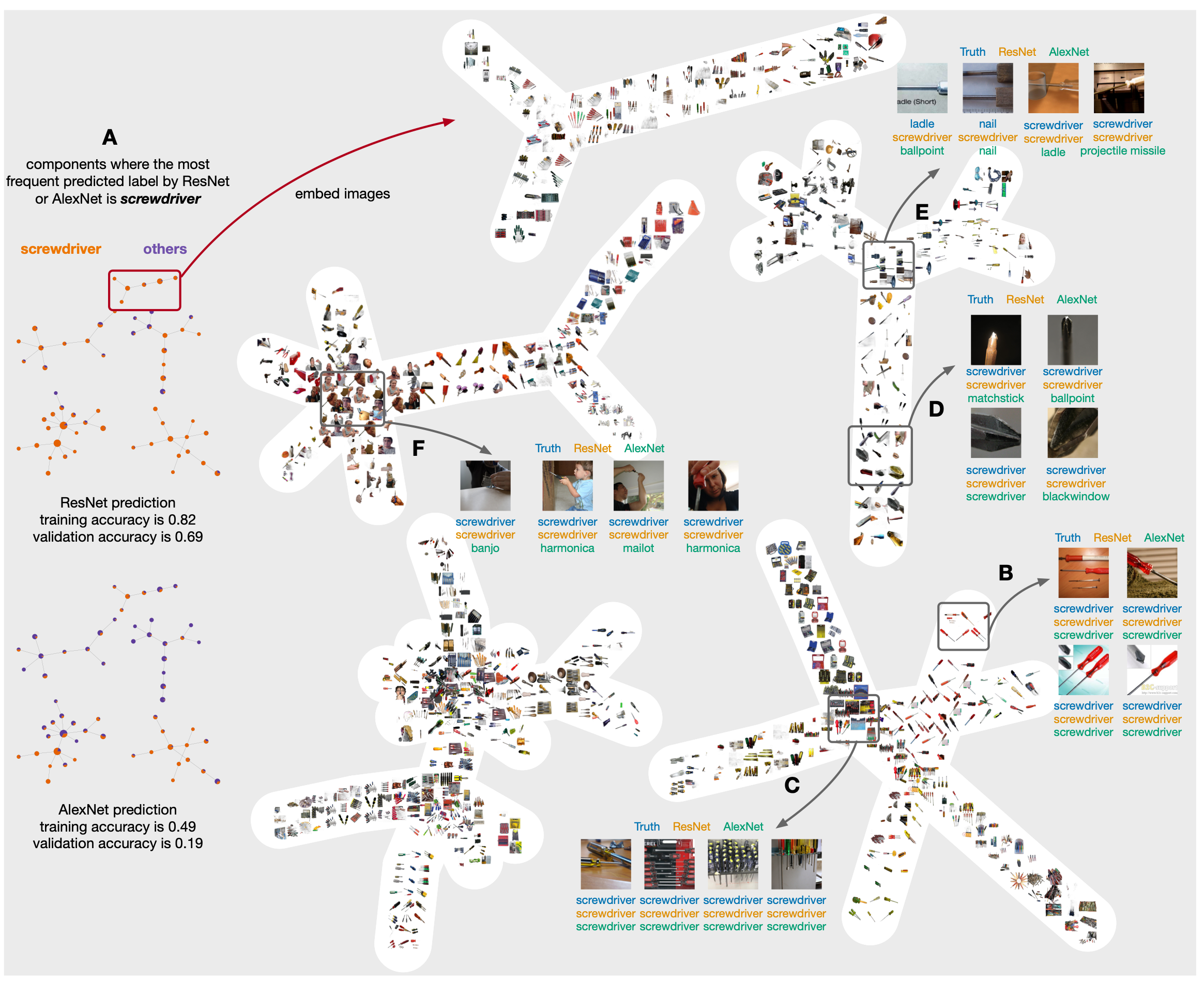}
\caption{In this figure, we analyze the prediction of ``screwdriver'' from both ResNet and AlexNet. We can see AlexNet can only predict ``screwdriver'' with high accuracy if both handle and the tip are clearly visible in the image (see B and C). Otherwise, if only the tip (D) or a small part of the handle (E) is shown or the image is about a person using a screwdriver (F), AlexNet will likely fail while ResNet still maintains high accuracy.}
\label{screwdriver}
\end{figure}

\begin{figure}[tp]
    \centering
    \includegraphics[width=\linewidth]{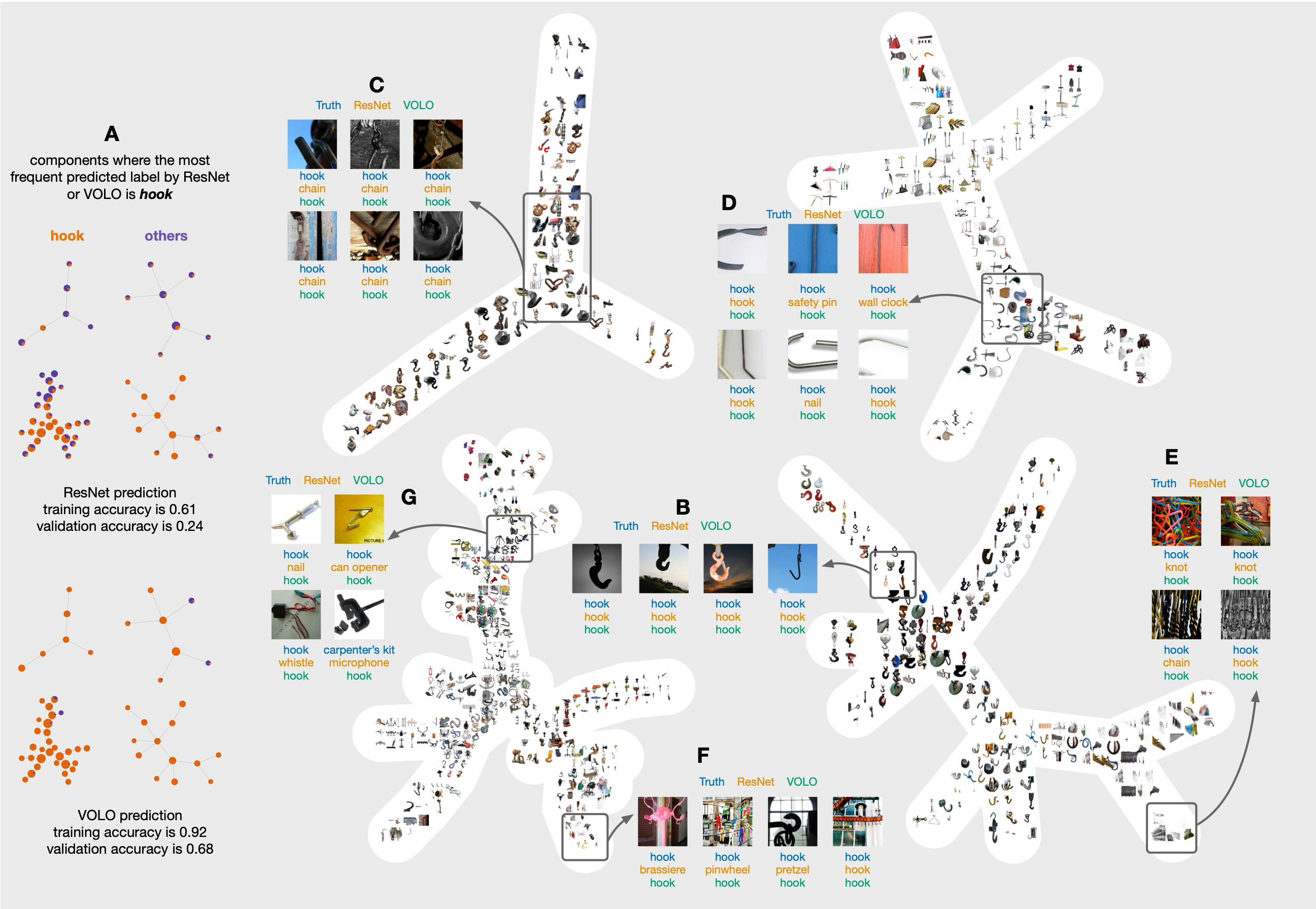}
\caption{In this figure, we analyze the prediction of ``hook'' from both ResNet and VOLO. VOLO has much higher training and validation accuracy on this class than ResNet (A). We first find subgroups of images that shows a single hook where both model have high accuracy (B). Then we find ResNet model often prefers to predict chain instead of hook if they are both present in the image (C). ResNet model also has difficulty predicting hook if only part of the hook is shown (D), or the shape of the hook is not common (G) and (F), or there are lots of hooks in the image (E).}
\label{hook}
\end{figure}

Figure~\ref{screwdriver} shows the results on one class, ``screwdriver'', from GTDA when comparing AlexNet with ResNet. AlexNet and ResNet have huge difference in terms of training or validation accuracy as shown in (A) of Figure~\ref{screwdriver}. By embedding images on top of each component, we can find different subgroups inside the ``screwdriver'' class, where some groups like (B) or (C) can be predicted with high accuracy by both models, while for some other groups like (D), (E) or (F), only ResNet can maintain the high accuracy. By showing some example images from each group, we can see that in general, AlexNet can only find the screwdriver if both the handle and the tip are clear enough in the image. Showing only some part of the screwdriver or having a slightly complex background will likely cause AlexNet to fail. Similarly, \Cref{hook} compares with prediciton of ResNet and VOLO on the ``hook'' class. We first find subgroups of images that shows a single hook where both model have high accuracy in group (B). Then we find ResNet model often prefers to predict chain instead of hook if they are both present in the image from group (C). ResNet model also has difficulty predicting hook if only part of the hook is shown (D), or the shape of the hook is not common (G) and (F), or there are lots of hooks in the image (E).

\subsection{Understanding different models' predictions}
\begin{figure}[tp]
    \centering
    \includegraphics[width=\linewidth]{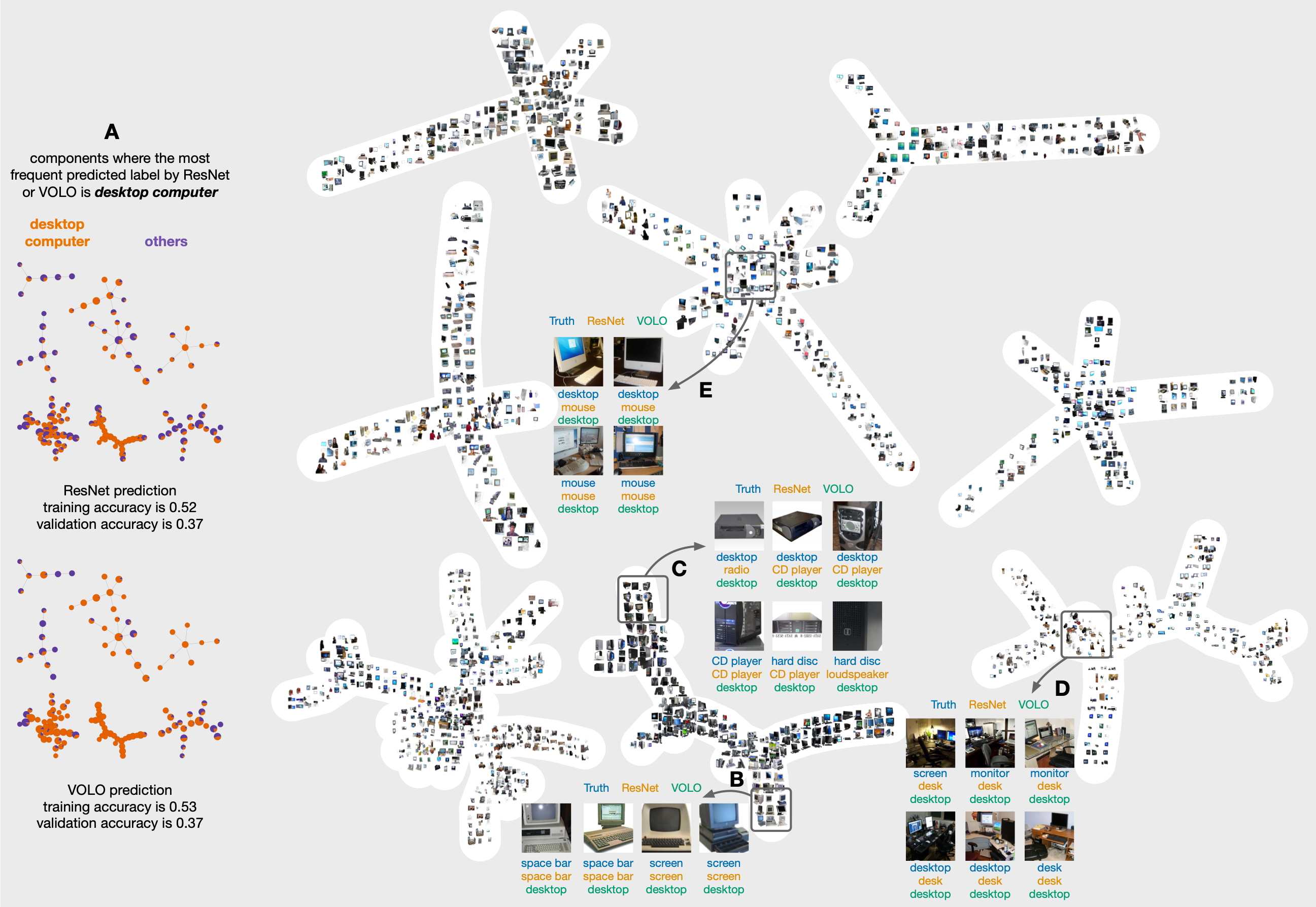}
\caption{In this figure, we analyze the prediction of ``desktop computer'' from both ResNet and VOLO. In (A), we show all components GTDA has found where ``desktop computer'' is the most frequent predictions. ResNet and VOLO show very close training and validation accuracy on these components. By embedding images on them, we can first find subgroups of images that look confusing. For instance, some images in (B) have labels like ``space bar'' or ``screen'' despite they are just old fashioned desktop computers. Images in (C) show some ``CD player'' or ``hard disc'' that look very similar to PC chasis. Images in (D) have desk, desktop computer and monitor at the same time. And some images in (E) are labeled as ``mouse'' even if they also contain a monitor or a keyboard. We can also notice how ResNet and VOLO handle these confusing images differently. Overall, VOLO's predcition on ``desktop computer'' is more robust and less affected by other objects in the image or similar objects from other classes.}
\label{desktop}
\end{figure}

In Figure~\ref{desktop}, we compare the the predictions between ResNet and VOLO on ``desktop computer''. Both models have very similar training or validation accuracy on this class. But they make mistakes in different places. We highlight a few subgroups where we can see lots of difference in predicted labels. These subgroups contain images that are indeed confusing. For instance, images in group (D) clearly have a desk, a monitor and a desktop computer at the same time. We can see VOLO tends to predict all these confusing images as ``desktop computer'', even though the true labels for some of those images are different. This suggests the VOLO prediction of ``desktop computer'' is more robust, while the ResNet prediction is more likely to be affected by other objects in the image.

%% file: exp_mutation.tex
\section{Understanding Malignant Gene Mutation Predictions}
In this section, we apply our method to inspect model predictions of gene sequence variants effects. A gene sequence variant means that some part of the DNA sequence for this gene is mutated compared with the reference. Modifications include single nucleotide variation, deletion, duplication, etc. We study a model proposed to predict whether such variant is harmful or not~\cite{avsec2021effective}. In the following section, we will provide details on the model and the dataset we use. Then we will show that the model's prediction is highly correlated with both gene variants coordinates as well as mutation types. We also discover abnormal places that could imply unreliable labels.
\subsection{Dataset and model} 
The model we use is recently proposed to predict gene expression from DNA sequence by integrating long-range interactions~\cite{avsec2021effective}. In this model, a consecutive DNA sequence of 196,608bp is used to predict 5,313 human genome tracks. For each gene variant, we follow the same steps as proposed by~\cite{avsec2021effective} to compute its embedding. First, we extract the reference and alternate DNA sequences from homo sapiens (human) genome assembly, either hg19 or hg38 as specified by the gene variant record. This gives a 393,216bp long DNA sequence with the centered on the VCF position (Variant Call Format). Note that for the alternate sequence, the gene variant is applied first before extracting the modified sequence. Then, we directly use the pretrained model from~\cite{avsec2021effective} to make predictions on the reference and alternate sequences. This model will aggregate the center 114,688bp into 128-bp bins of length 896. The prediction for each 128bp bin is a 5,313 vector, where each element represents the predicted gene expression in one of the 5,313 genome tracks for the human genome (including 2,131 transcription factor chromatin immunoprecipitation and sequencing tracks, 1,860 histone modification tracks, 684 DNase-seq or ATAC-seq tracks and 638 CAGE tracks). The prediction vector of the 4 128bp bins located in the center is then summed together to get a prediction vector for the reference or alternate sequence. After that, the elements in each prediction vector corresponding to the CAGE tracks is $\log(1+x)$ transformed. Finally, we compute the difference of preprocessed prediction vectors between reference and alternate sequences as the final embedding for the gene variant. In total, we get a 23,376-by-5,313 embedding matrix for 23,376 gene variant records. Then, a linear classifier will be trained on this 5,313 difference vector to predict variants effects. The original paper uses the training and testing datasets from CAGI5 competition~\cite{shigaki2019integration}, where a Lasso regression is trained to predict a label of -1 (significant downregulating effect), 0 (very little to no effect on expression) or +1 (significant upregulating effect). We were not able to download the dataset from the official CAGI5 competition website. Therefore, we use similar procedure to predict harmful (label 1) vs non-harmful (label 0) gene mutations from ClinVar. We download gene variants experiments from the official ClinVar website~\cite{landrum2018clinvar}. We choose all experiments that are targeting \textbf{BRCA1} as it is one of the genes with the most number of experiments and part of the protein it encodes has known 3D structures (i.e.~\textbf{1JNX}). Gene variants without a valid VCF (variant call format) position are removed. As for the labels, we directly use the ``ClinSigSimple'' field as the label of each gene variant record. An integer 1 means at least one current record indicates ``Likely pathogenic'' or ``Pathogenic'', but doesn't necessarily mean this record includes assertion criteria or evidence. An integer 0 means there are no current records of ``Likely pathogenic'' or ``Pathogenic''. An integer -1 means no clinic significance and is replaced by label 0 in our experiments. And we use a logistic regression with L1 penalty since this is a binary prediction. We include 23,376 gene variants where 50\% of them are used as training, and the other 50\% are used as testing. To build the graph for GTDA, the embedding matrix is PCA reduced to 128 dimensions with PCA whitening and then each row is $\ell_2$ normalized. A 5-NN graph is constructed on the preprocessed embedding matrix with cosine similarity. This 5-NN graph has some small components smaller than the threshold set by $s_1$ and $s_2$. As a result, 338 out of 23,376 gene variants ($\sim$ 1.4\%) are not included in the final Reeb net; this is not expected to impact the results.  We use 2 prediction lens and the first 2 PCA lens of the embedding matrix for GTDA analysis. For GTDA parameters, we set $K=30$, $d=0$, $r=0.05$, $s_1=5$, $s_2=5$, $\alpha=0.5$ and $S=10$. We use 20 iterations for GTDA error estimation.
\subsection{Validating the GTDA visualization}
\begin{figure}[t]
    \centering
\includegraphics[width=\linewidth]{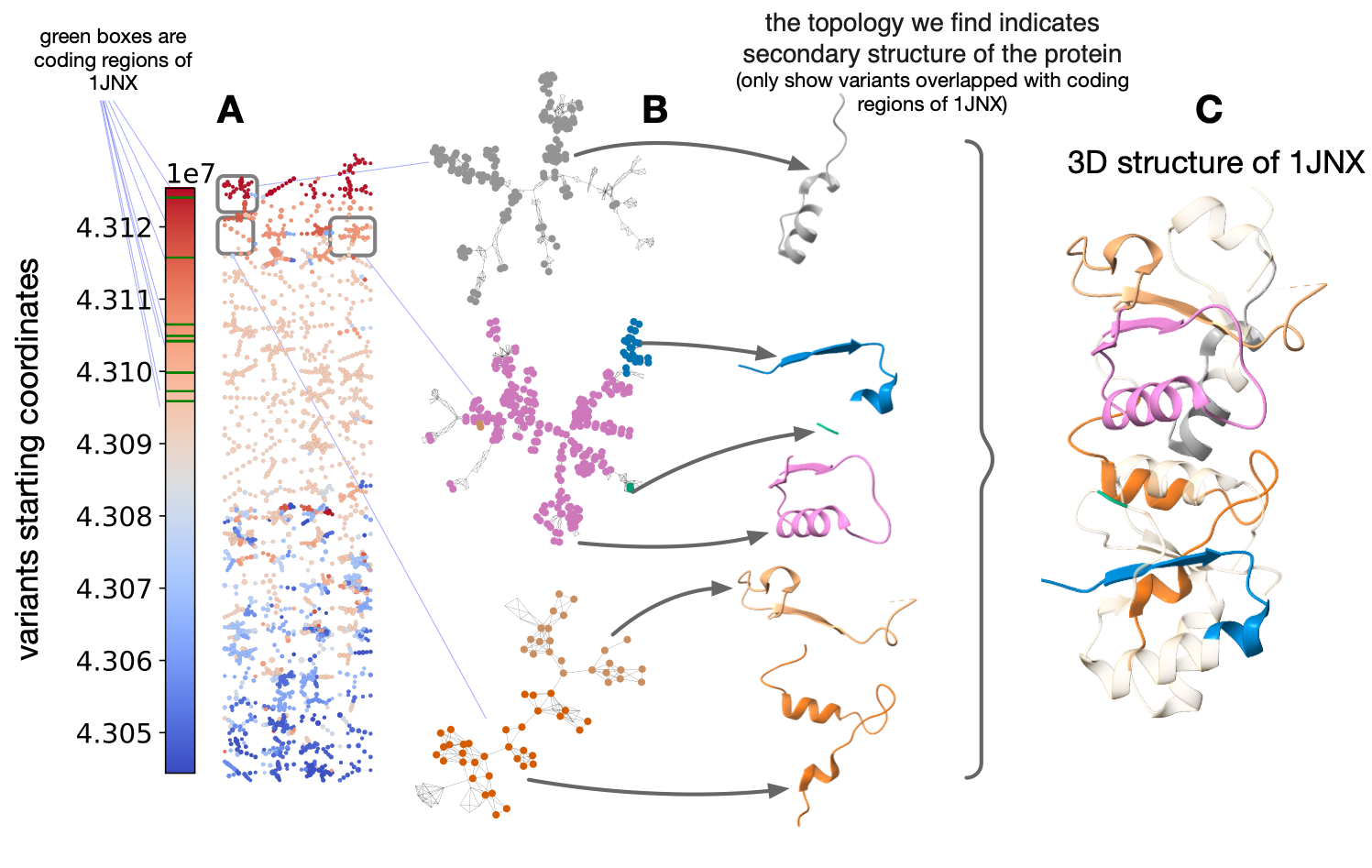}
\caption{(A) shows components found by GTDA, where each node is colored by median hg38 coordinates of mutation starting positions. Different components are ordered by the averaged median coordinates in a zig-zag fashion from lower right to upper left. We zoom in a few components where the gene variants have the highest overlap ratio with the coding regions of \textbf{1JNX} (B). Different node colors are assigned based on which consecutive protein coding region they overlap with. Nodes for gene variants not in the coding regions of 1JNX are not plotted. We can find that different secondary structures of the crystal of \textbf{1JNX} (C) are also well separated in the GTDA visualization.}
\label{gene_location}
\end{figure}

The visualization we get from this dataset is shown in \Cref{gene_location}. The first finding is that different components in this visualization are strongly related to different regions of the DNA sequence. Such a result is not surprising because this model aims to predict gene expressions from a long range of DNA sequence while most gene variants will only change one or two base pairs. Therefore, it is expected that gene variants close to each other in coordinates will also get similar embeddings. To further validate whether this visualization can capture finer 3D protein structures, we check the crystal structure of the BRCT repeat region (PDB id is \textbf{1JNX}), also shown in plot (C) of \Cref{gene_location}. In total, \textbf{BRCA1} encodes a protein with 1863 amino acids. And \textbf{1JNX} covers amino acids from 1646 to 1849. In the color bar of \Cref{gene_location}, we mark the protein coding regions (exons) of \textbf{1JNX} in green. In (B) of \Cref{gene_location}, we check a few components in detail that contains gene mutation locations overlapped with the green area. Each green area represents an exon. Different node colors are assigned based on which exon they overlap with. \footnote{4 out of 2756 mutation samples overlap with more than 1 exons of 1JNX, we color those based on the first exon they overlap with.} We can find that different local structures of this crystal are also very well localized in our visualization. All these findings suggest that the model's embedding space has a strong correlation with VCF (variant call format) positions of gene variants and GTDA can capture such property successfully.
\paragraph{Statistical validation} We conduct 10000 random experiments to see if such strong location sensitivity can be found in a random graph. In each experiment, we shuffle the embeddings and rebuild the KNN graph. The PCA lenses and prediction lenses are kept the same. Then we run GTDA on each of the 10000 random graphs. We consider a random graph to shows location sensitivity if in the results of GTDA, one component has more than 40 mutation samples that overlap with exons and more than half of them are overlapped with the same exon. We were not able to find any random graph that can pass this criteria in these experiments.

\subsection{Estimating and correcting prediction errors} 
\begin{figure}[tp]
    \centering
\includegraphics[width=\linewidth]{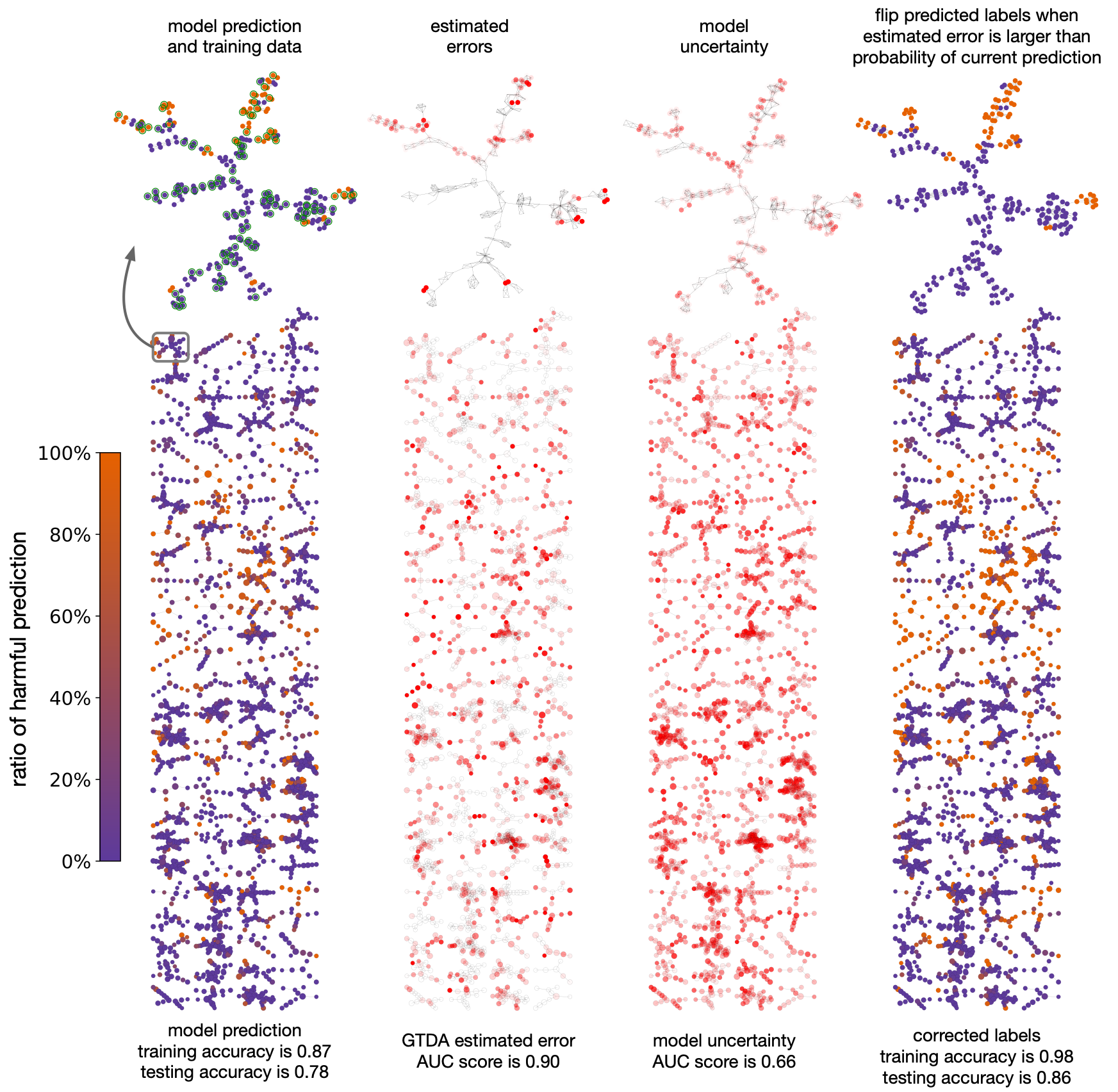}
\caption{In the top part, we zoom in a component and mark training data using green circles. Then we show GTDA estimated errors and model uncertainty on this component. We flip predicted labels if the estimated error is larger than the prediction probability. In the lower part, we can see GTDA error estimation has much better overall AUC score and the corrected labels have higher training and testing accuracy.}
\label{gene_correction}
\end{figure}

We apply \Cref{error_est} to estimate errors of model prediction \Cref{gene_correction}. Overall, GTDA estimated errors (after normalizing to 0 to 1) achieve an AUC score of 0.90. In comparison, using model uncertainty gives an AUC score of 0.66. Since this is a binary classification, we can also flip predicted labels if they are more likely to be errors. Instead of setting a single threshold, we flip predicted labels when the estimated errors are larger than the probability of the current prediction. The corrected labels can improve training accuracy from 0.87 to 0.98 and testing accuracy from 0.78 to 0.86.

\subsection{Extracting insights about mutation types and single nucleotide variants}\label{sec:types} 
\begin{figure}[t]
\centering
\includegraphics[width=\linewidth]{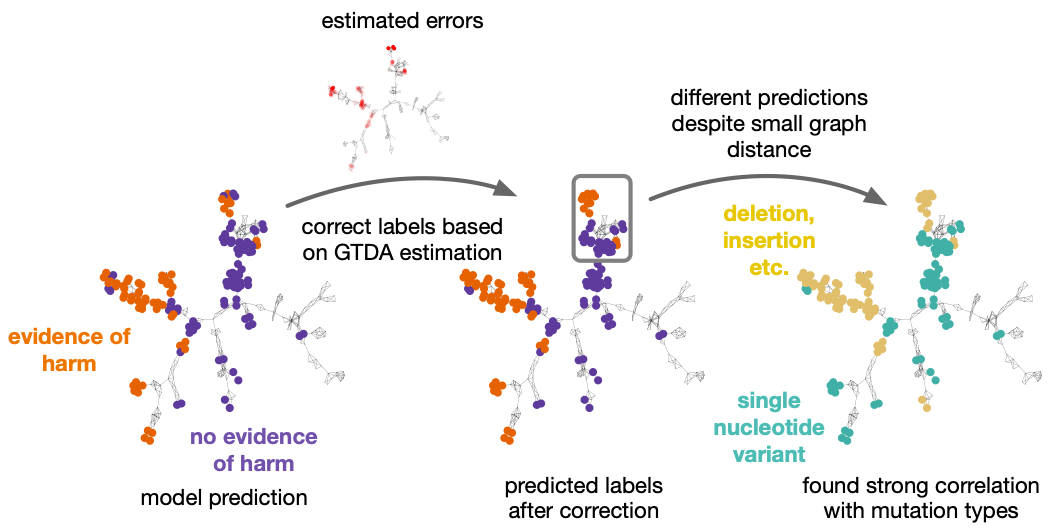}
\caption{We zoom in one component GTDA finds and only show mutation records that happen in the protein coding regions (non-coding regions are not shown as colored dots, but do impact the Reeb net structure). After correcting prediction based on GTDA error estimation, we still see records that happen in a small region of the protein but still get different predictions. By checking these records, such difference can be well explained by different mutation types.}
\label{mutation_type}
\end{figure}

As we explore model predictions for gene mutations happening inside protein encoding regions, i.e., green boxes in \Cref{gene_location}, we find different predicted labels for mutations that target a small area of the protein structure. One such example is \Cref{mutation_type}, where records in the grey box happen in a small region of the protein structure with around 17 amino acids. So there should be other aspects that help the model make different predictions. By checking the actual mutation record, we find the non-harmful mutations are all single nucleotide variant (SNV), while harmful mutations are other types of mutations including deletion, insertion or duplication. This makes sense as the latter types will not only affect the current amino acid, but also the subsequent amino acids and hence cause more substantial changes to the final protein structure. 

Overall, we find for gene mutations that are predicted harmful (after GTDA correction) and target gene encoding regions, 70\% of them are mutations like deletion, insertion or duplication. 
For gene mutations that are predicted as non-harmful and target gene encoding regions, only 6\% are mutations like deletion, insertion or duplication. 
When including gene mutations outside protein encoding regions as well, 72\% of harmful predictions are mutations like deletion, insertion or duplication, while that number drops to 5\% for non-harmful predictions. 

We assess the statistical significance of the relationship between single nucleotide variants (SNV) and harmful predictions for each component GTDA identifies in \Cref{contigency}. The associated Chi-square $p$-values highlight a few components where this association is missing, such as component 100 with 34 non-harmful non-SNV results throughout the entire BRCA1 structure (coding and non-coding regions), with a $p$ value of $0.22$. This suggests a difference in behavior for this component in comparison with the remainder of the components. Other large $p$-values include the nearby components 99 and 101, along with component 3, 26. Overall, this highlights another way these GTDA results could be used. 

\input{"figures/mutation/stats.tex"}

\subsection{Incorrect GTDA error estimation implies unreliable labels}
When we compared the GTDA error estimation with true errors, we found a few places where the GTDA estimate is entirely wrong. 

\begin{figure}[tp]
    \centering
    \includegraphics[width=\linewidth]{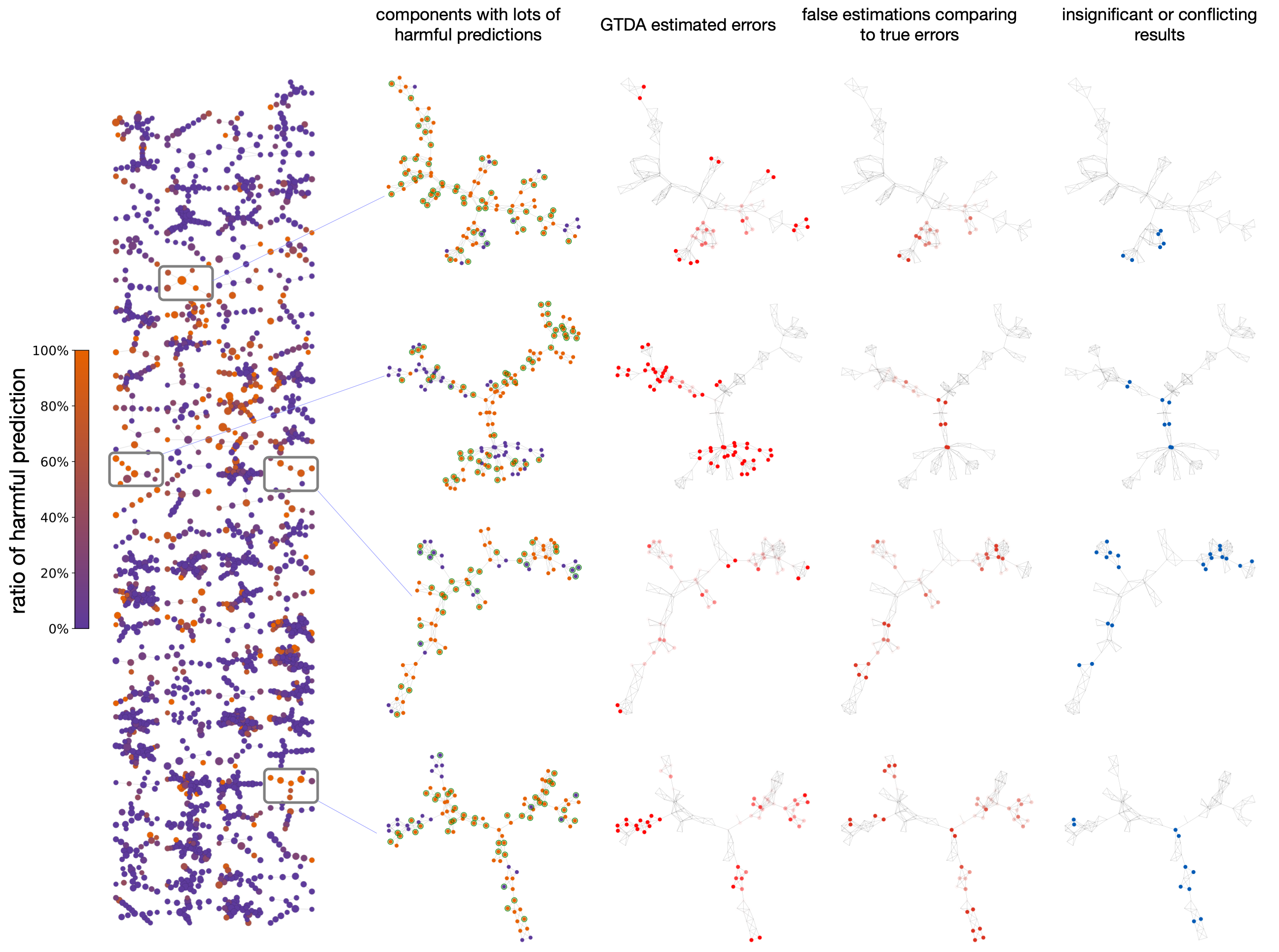}
\caption{Checking false error estimations of GTDA in some components reveals that they are likely to be caused by variants experiments with insignificant or conflicting results.}
\label{insignificance}
\end{figure}

To understand this abnormality, in \Cref{insignificance} we zoom in a few components and use green circles to mark training and validation data. We show the GTDA estimated errors as well as the false estimations when comparing to the true errors. We can see a few false error estimates in each of these components. And on checking those false estimations, we find they are either testing experiments with insignificant or conflicting results or affected by nearby insignificant training experiments.

To understand this effect across all components found by GTDA, we use the difference between the true presence of an error and our estimate.  For instance, if GTDA estimation on whether a prediction is wrong is 0.3 and the prediction is indeed wrong based on its true label, such difference will be 1 minus 0.3. 
 In total, we can find 2,031 GTDA error estimations where such difference is larger than 0.5. These are spread over 771 Reeb nodes. Since an error estimation being wrong can be due to either its own label being unreliable or  training samples nearby have unreliable labels,
 we study how many of those $771$ Reeb nodes have at least 1 insignificant or conflicting samples (either training or testing sample). We find 662 of them (81\%) have at least one problematic label.  Consequently, the intuition from Figure 24 would hold across much of the dataset. 

\input{exp_mutation_comparison.tex}

%% file: figures/mutation/stats.tex
\begin{table}[tp]
\tiny 
\centering
\begin{tabular}{@{}ccccccccccc@{}}
\toprule
& \multicolumn{4}{c}{Prediction and Type (coding regions of 1JNX)} & & \multicolumn{4}{c}{Prediction and Type (all)} & \\
\cmidrule{2-5} 
\cmidrule{7-10}
& Harmful & Harmful & non-Harm & non-Harm & \shortstack{Chi-square \\p-value} & Harmful & Harmful & non-Harm & non-Harm & \shortstack{Chi-square \\p-value}\\
\cmidrule(lr){2-3} 
\cmidrule(lr){4-5} 
\cmidrule(lr){7-8} 
\cmidrule(lr){9-10}
Component & SNV & non-SNV & SNV & non-SNV &  & SNV & non-SNV & SNV & non-SNV  \\
\midrule
0&11&6&83&4&5.1e-04&13&6&167&8&1.2e-04\\ \arrayrulecolor{lightgray} \hline
2&0&0&0&0&-&17&264&49&16&1.2e-36\\ \hline 
3&1&3&99&5&1.8e-05&12&3&230&9&2.5e-02\\ \hline 
4&0&0&0&0&-&13&14&181&4&1.2e-16\\ \hline 
6&0&10&10&2&-&24&38&298&20&2.7e-27\\ \hline 
7&0&0&0&0&-&6&42&152&24&1.5e-22\\ \hline 
8&0&4&0&0&-&16&82&96&22&6.2e-21\\ \hline 
9&0&0&0&0&-&6&23&44&11&4.9e-07\\\hline 
10&0&0&0&0&-&17&102&129&22&1.0e-30\\ \hline 
14&0&2&2&0&-&13&31&297&19&7.6e-30\\ \hline 
15&6&2&40&0&-&25&146&437&24&8.7e-90\\ \hline 
16&0&0&0&0&-&6&155&136&13&3.9e-53\\ \hline 
17&5&14&49&4&6.5e-08&55&93&485&23&2.7e-59\\ \hline 
18&0&0&2&0&-&9&7&115&3&7.5e-08\\ \hline 
19&0&8&0&2&-&36&70&422&32&1.0e-44\\ \hline 
21&0&6&64&2&-&20&31&376&17&4.7e-33\\ \hline 
22&12&16&102&2&4.2e-13&32&20&188&6&1.1e-12\\ \hline 
25&0&0&0&0&-&15&17&19&3&7.7e-03\\ \hline 
26&0&0&0&0&-&34&4&256&12&2.4e-01\\ \hline 
28&0&1&0&1&-&29&42&63&12&1.7e-07\\ \hline 
29&3&6&193&18&8.1e-07&9&9&339&31&1.3e-07\\ \hline 
30&0&0&0&0&-&19&57&93&9&6.4e-19\\ \hline 
31&0&0&0&0&-&16&18&68&8&4.3e-06\\ \hline 
32&0&4&2&0&-&5&23&51&5&1.0e-10\\ \hline 
33&10&55&64&11&5.5e-16&16&55&204&23&1.1e-28\\ \hline 
34&16&18&32&0&-&32&70&66&12&3.5e-12\\ \hline 
36&2&4&250&6&1.8e-12&8&11&314&23&3.2e-12\\ \hline 
37&0&0&0&0&-&40&76&26&14&1.5e-03\\ \hline 
38&0&0&0&0&-&21&37&137&19&8.6e-14\\ \hline 
39&0&0&0&0&-&14&198&24&14&3.4e-18\\ \hline 
40&0&2&6&0&-&50&81&488&13&1.5e-63\\ \hline 
41&0&0&2&0&-&16&14&158&6&1.1e-11\\ \hline 
44&0&0&0&0&-&27&29&423&17&4.2e-30\\ \hline 
46&0&16&30&2&-&19&36&51&10&2.0e-07\\ \hline 
48&0&4&20&4&-&30&14&220&16&3.1e-06\\ \hline 
50&0&0&62&0&-&4&36&262&10&2.2e-45\\ \hline 
52&0&4&18&2&-&60&67&830&79&5.7e-40\\ \hline 
53&0&0&0&0&-&15&27&303&25&2.7e-22\\ \hline 
54&0&0&10&0&-&19&27&365&13&2.5e-32\\ \hline 
55&0&0&0&0&-&21&19&109&3&2.8e-11\\ \hline 
56&0&0&0&0&-&5&34&29&4&9.4e-10\\ \hline 
57&0&0&0&0&-&40&18&78&6&4.5e-04\\ \hline 
58&2&2&0&0&-&25&30&157&10&2.3e-15\\ \hline 
59&0&0&0&0&-&17&44&163&6&6.6e-28\\ \hline 
60&6&0&36&2&-&6&7&130&13&2.8e-05\\ \hline 
62&2&33&12&7&1.9e-05&12&69&218&35&8.1e-33\\ \hline 
64&2&25&114&7&5.0e-22&6&25&192&17&4.4e-22\\ \hline 
66&0&4&24&0&-&27&23&165&9&1.9e-12\\ \hline 
67&0&6&0&0&-&9&30&111&4&1.0e-20\\ \hline 
68&21&13&87&9&3.3e-04&76&108&570&92&3.8e-36\\ \hline 
69&2&3&78&7&4.4e-03&4&11&314&35&8.6e-12\\ \hline 
70&0&0&0&0&-&40&48&6&6&1.0e+00\\ \hline 
71&0&0&0&0&-&17&8&269&18&4.4e-05\\ \hline 
72&0&0&12&0&-&6&5&318&21&1.7e-05\\ \hline 
73&0&0&0&0&-&12&9&142&3&3.1e-10\\ \hline 
74&0&0&0&0&-&19&11&119&15&1.5e-03\\ \hline 
77&6&20&14&4&1.1e-03&13&33&213&11&6.0e-28\\ \hline 
78&0&0&2&0&-&3&8&181&6&4.1e-16\\ \hline 
79&10&4&52&0&-&33&10&203&6&3.3e-06\\ \hline 
81&0&0&0&0&-&9&57&95&15&9.5e-21\\ \hline 
82&0&0&0&0&-&8&4&212&6&1.5e-05\\ \hline 
85&0&1&38&1&-&14&61&496&29&5.3e-65\\ \hline 
88&0&0&0&0&-&3&34&269&16&6.8e-41\\ \hline 
90&2&4&76&2&4.4e-07&10&4&162&4&1.0e-04\\ \hline 
99&0&0&8&0&-&6&3&100&7&3.3e-02\\ \hline 
100&0&0&2&0&-&4&6&64&34&2.2e-01\\ \hline 
101&0&2&2&0&-&6&4&60&6&2.8e-02\\ \hline 
102&0&0&0&0&-&7&9&109&5&5.3e-09\\ \hline 
Overall&148&344&2114&122&2.9e-259&1506&3986&16208&1338&0.0e+00\\
\arrayrulecolor{black}
\bottomrule
\end{tabular}
\caption{For each component in the Reeb networks, 2 contingency tables are computed, where the left table only considers variants in the coding regions of 1JNX and the right table considers all variants. Only components where each cell of the right table has a count 3 or higher are included. Chi-square p-values are computed for tables where each cell has a count larger than 0.}
\label{contigency}
\end{table}

%% file: exp_mutation_comparison.tex
\subsection{Comparison with other methods}
\label{sec:mutation_comparison}
\begin{figure}[p]
  \centering
    \includegraphics[width=\linewidth]{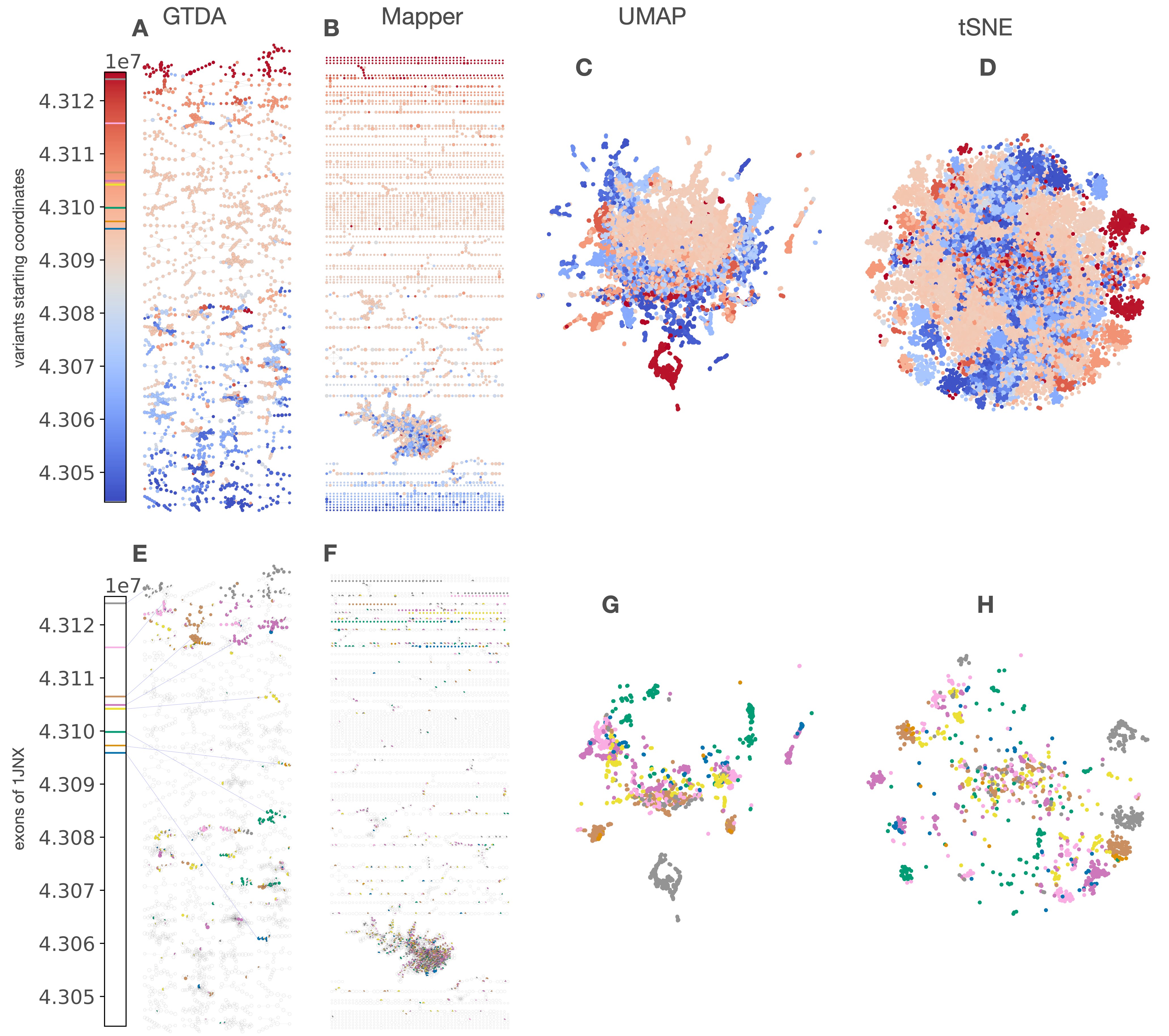}
   \caption{The topological simplification identified by GTDA is highly correlated with DNA variant starting location (A). Alternative global visualizations such as the simplification from Mapper (B), or dimensionality reduction techniques UMAP (C) and tSNE (D) show significantly less sensitivity to the variant location ($p < 0.001$ in a KS test, see  Table~\ref{tab:ks_mutation}). Likewise, the GTDA results strongly localize the exons of the 1JNX structure within the BRCA1 gene (E). This is also significantly better than Mapper (F), UMAP (G), and tSNE (H) ($p < 0.001$, see Table~\ref{tab:ks_mutation}). These results demonstrate both how the Enformer model is sensitive to these aspects of gene expression and also that GTDA makes them possible to inspect.  }
\label{fig:enformer-dimred}
\end{figure}

We visually compared the visualization of GTDA with other methods including Mapper, UMAP and tSNE in \Cref{fig:enformer-dimred}. We can find that the visualization of GTDA clearly shows the location sensitivity of mutation samples in the DNA sequence that is not quite obvious in the visualizations of other methods. 

\begin{figure}[t]
    \centering
    \includegraphics[width=\linewidth]{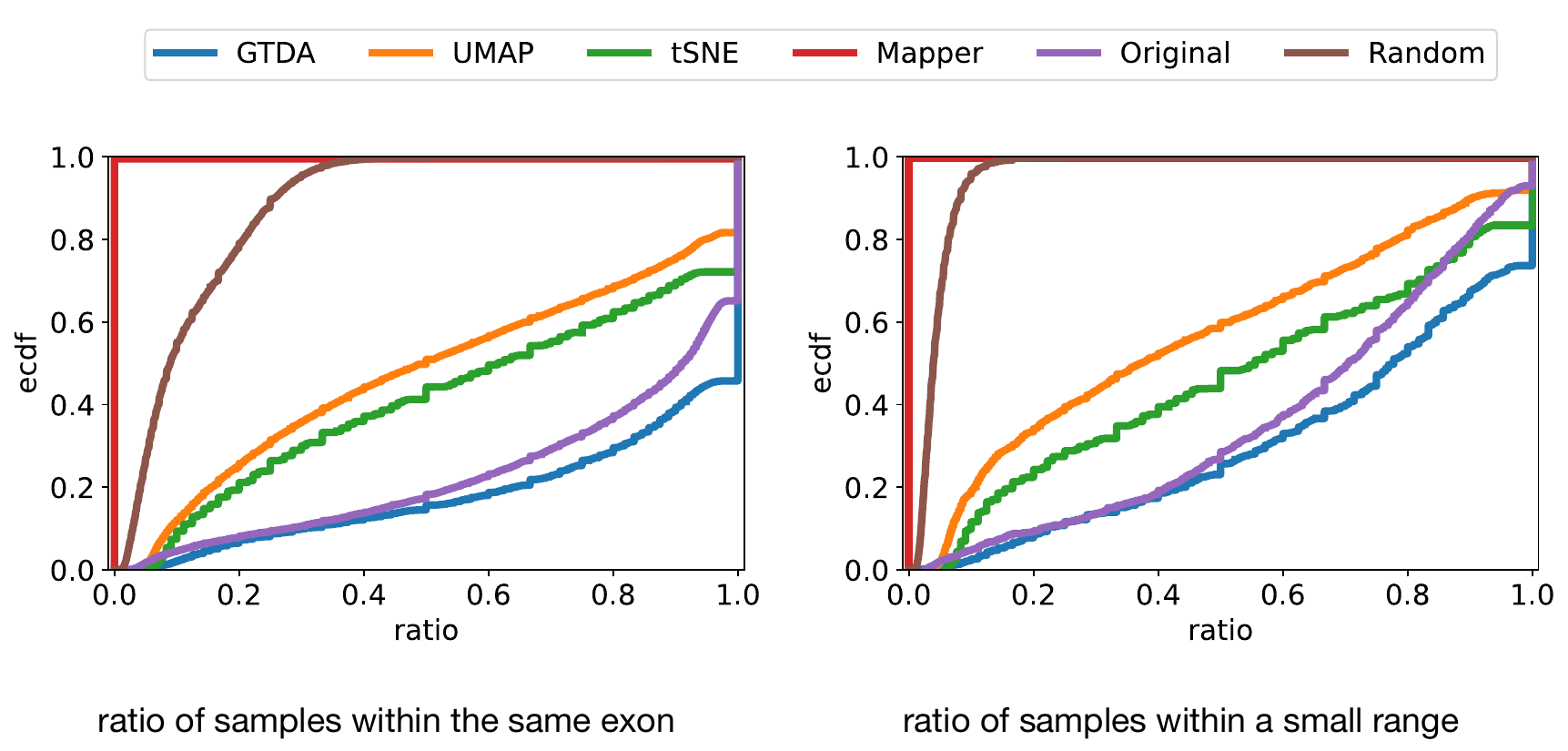}
    \caption{Overall GTDA performs the best on both metrics, while the other methods are not clearly better or even worse than the original graph. This suggests (1) the strong location sensitivity of mutation samples indeed exist in the original graph (2) GTDA can not only preserve and enhance such location sensitivity, but also visualize such property easily.}
  \label{fig:hist_comparison_mutation}
  \end{figure}

In the following of this section, we quantify this visual advantage of GTDA. We first convert UMAP and tSNE visualizations into graphs by building a 5-NN graph on top of the 2 dimensional embedding. For GTDA and Mapper, we project each Reeb net node using Algorithm \Ref{Reeb_projection} to get the corresponding graphs. We also add the original KNN-graph that has been used as input to GTDA and 100 random graphs by shuffling edges for comparison. Then we design the following metrics:
\begin{itemize}
    \item ratio of samples within the same exon: in this metric, for each mutation sample that overlaps with an exon, we search the neighbors within 3 hops on each graph and compute the ratio of mutation samples that overlap with the same exon. Note that we only consider exons that encodes 1JNX.
    \item ratio of samples within a small range: in this metric, for each mutation sample, we search the neighbors within 3 hops on each graph and compute the ratio of mutation samples whose mutation starting coordinates are within 1000 base pairs of the starting coordinate of the selected mutation sample.
\end{itemize}
We also consider the corresponding ratio to be zero if the number of neighbors within 3 hops is smaller than 5. This is because the visualization of Mapper has too many single nodes or tiny components which could result in better metrics despite the visualization itself is much worse. In \Cref{fig:hist_comparison_mutation}, we compare the empirical cumulative distribution function of the ratio distributions. For each of the 100 random graphs, we compute the ecdf and report the average of the 100 ecdf curves. We can first notice that comparing to random graphs, the ratio in both metrics is much higher in the original graph, which means mutation samples are indeed significantly localized in the original graph. Also, GTDA performs the best on both metrics., which can be verified by the Kolmogorov–Smirnov test in \Cref{tab:ks_mutation}.

  \begin{table}[tp]
    \centering
    \begin{tabular}{cccc}
    \toprule
     & GTDA vs tSNE & GTDA vs UMAP & GTDA vs Mapper \\
    \cmidrule(lr){2-2}
    \cmidrule(lr){3-3}
    \cmidrule(lr){4-4}
    ratio within the same exon &  0.23 $\quad p < 10^{-10}$ & 0.35 $\quad p < 10^{-10}$ & 0.99 $\quad p < 10^{-10}$\\
    ratio within a small range &  0.33 $\quad p < 10^{-10}$ & 0.40 $\quad p < 10^{-10}$ & 0.97 $\quad p < 10^{-10}$\\
    \bottomrule
    \end{tabular}
    \caption{The ks statistics and p-value of the one tailed Kolmogorov–Smirnov test. The null assumption is that the ecdf of GTDA is larger than the ecdf of other methods at all locations. The $p$-values were extremely small or numerically 0 in floating point, which we report as less than $10^{-10}$}
    \label{tab:ks_mutation}
\end{table}

\begin{figure}[h!]
    \centering
    \includegraphics[width=\linewidth]{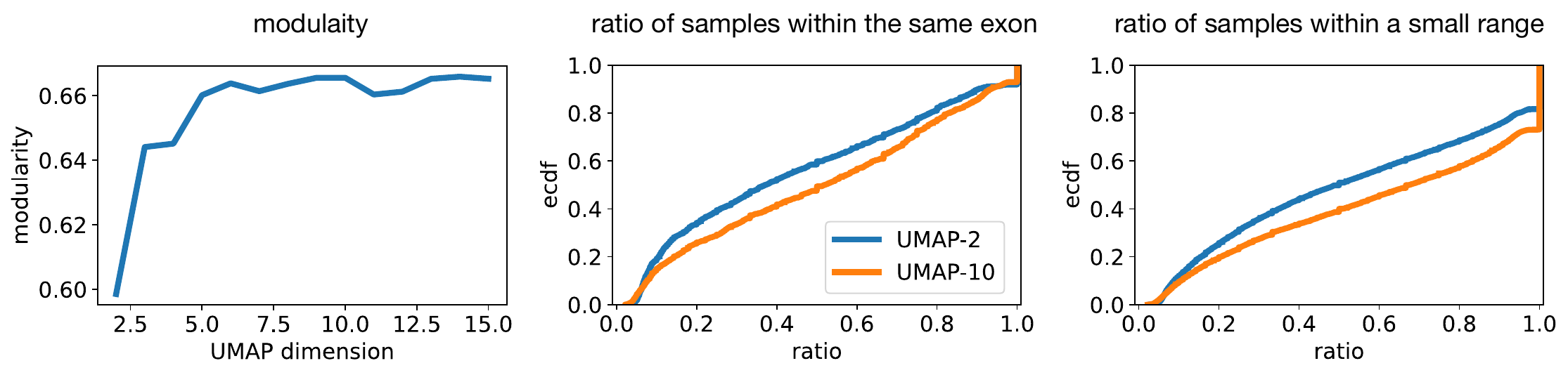}
    \caption{In the first plot, we compute the modularity by considering mutation samples that overlap with the same exon as a community. This plot shows that the modularity increases as we use more dimensions in the output of UMAP, which suggests these communities are less mixed in the corresponding KNN graph of UMAP embeddings in higher dimensions. In the second and third plots, we compare the two metrics we designed between UMAP embedding in 2 dimensions and UMAP embedding in 10 dimensions. Higher dimension also performs better on both metrics. More specifically, in one tailed Kolmogorov–Smirnov test, the ks statistics and p-value are 0.11 and $3.0^{-16}$ for ratio within the same exon and such numbers become 0.11 and $9.8^{-132}$ for ratio within a small range, showing that UMAP-10 is better localized than UMAP-2.}
  \label{fig:umap_comparison}
  \end{figure}

One key advantage of topological based methods like GTDA is that it visualize the embedding space by directly simplifying it without reducing the dimensions too much. In the case of mutation dataset, the KNN graph that will be used as the input of GTDA is constructed on a 128 PCA reduced embedding space. In comparison, tSNE or UMAP try to project the original embedding space on only 2 dimensions, which could cause huge information loss. We find that using more dimensions in UMAP can better preserve the location sensitivity. To show this, we increase the embedding dimensions on UMAP and rebuild the KNN graph using only mutation samples that overlap with exons. To study how mutation samples from different exons will be localized, we consider samples within each exon as a community and compute the graph modularity~\cite{Newman2004-community}. A higher modularity score means these communities are better localized in the graph. As we can see in \Cref{fig:umap_comparison}, the modularity score increases as we use more dimensions until after 10 dimensions. UMAP embedding in higher dimension also performs better on the two metrics we designed. However, using more than 2 dimensions on these types of methods will make the subsequent visualization difficult or impossible. We were not able to replicate the same experiment on tSNE as the running time of tSNE becomes extremely long when setting the dimension larger than 2. 

%% file: exp_chest.tex
\section{Inspecting chest X-ray images}
\label{sec:chest-xray}
\label{sec:xray-images}
In this section, we apply our GTDA framework to inspect the prediction of disease on 112,120 images of chest X-rays~\cite{wang2017chestx}. Each X-ray image might be either normal or indicating one or more diseases. Our results show that GTDA is very useful to help radiologists detect images with incorrect normal and abnormal labels.
\subsection{Dataset and model}
The NIH ChestX-ray14 dataset we use comprises 112,120 de-identified frontal-view X-ray images of 30,805 unique patients~\cite{wang2017chestx}. Among these images, 86,524 images are used as training or validation and the others are used as testing. Images are split at the patient level, which means images belonging to the same patient will be put in the same group. Among the 86,524 images, we randomly choose 20\% patients and use their associated images as validation data while images for the other patients are used as training data. In the original dataset, a text mining approach is used on the associated radiological reports to find the existence of 14 possible diseases and one image can have multiple disease labels. As a result, it is expected that many of the labels assigned are incorrect. In some other studies of these data,  expert labels are solicited for 810 selected testing images from multiple experienced radiologists~\cite{nabulsi2021deep}. 

The model we use GTDA to study is called CheXNet~\cite{rajpurkar2017chexnet} which is a 121-layer Dense Convolutional Network (DenseNet)~\cite{huang2017densely}. When applying our GTDA framework, we first reduce the 14 disease predictions to a simple normal (label 0) vs abnormal (label 1) prediction. To do so, we first take a row wise maximum to reduce the prediction matrix for 14 disease into a vector $v$ with values ranging 0 to 1. Then we consider each individual value as a threshold and generate predicted labels by treating values larger than this threshold as 1 or 0 otherwise. Then we compute the F1 score using the union of training and validation data. The threshold that gives the largest F1 score will be kept, denoted as $t$. Similar procedures have been used in other papers that predict ontological annotations~\cite{clark2013information, kulmanov2020deepgoplus}. Finally, we transform each value of $v$ using $v_i=\text{min}(1,0.5 v_i/t)$. The transformed $v$ also ranges from $0$ to $1$ and is considered as the probability of being abnormal. As a result, the row wise maximum column index of the new prediction matrix $\mP=[1-v,v]$ will give the same largest F1 score. Other than the abnormal vs normal lens, we also include the original disease prediction matrix as the extra lenses. This process gives 16 lenses in total. For GTDA parameters, we set $K=50$, $d=0$, $r=0.005$, $s_1=5$, $s_2=5$, $\alpha=0.5$ and $S=10$. We use 10  iterations for GTDA error estimation.

\subsection{GTDA finds incorrect normal vs abnormal labels}
Out of the 810 images in the test set with expert labels, 222 images have incorrect normal vs abnormal labels. Our goal is to use the GTDA visualization to find images in this set (i.e.~those that are more likely to an incorrect label). The procedure of finding those images is similar to find insignificant or conflicting gene mutation experiments from the previous section. 

We first use GTDA to estimate prediction errors. The estimation is normalized to a number between 0 and 1. Then we use the original testing labels (i.e.~without the correction from experts) to find which of these error estimates are wrong. We can then sort the test samples in the order of descending absolute difference between estimated error and true error. 

\begin{figure}[tp]
    \centering
    \includegraphics[width=\linewidth]{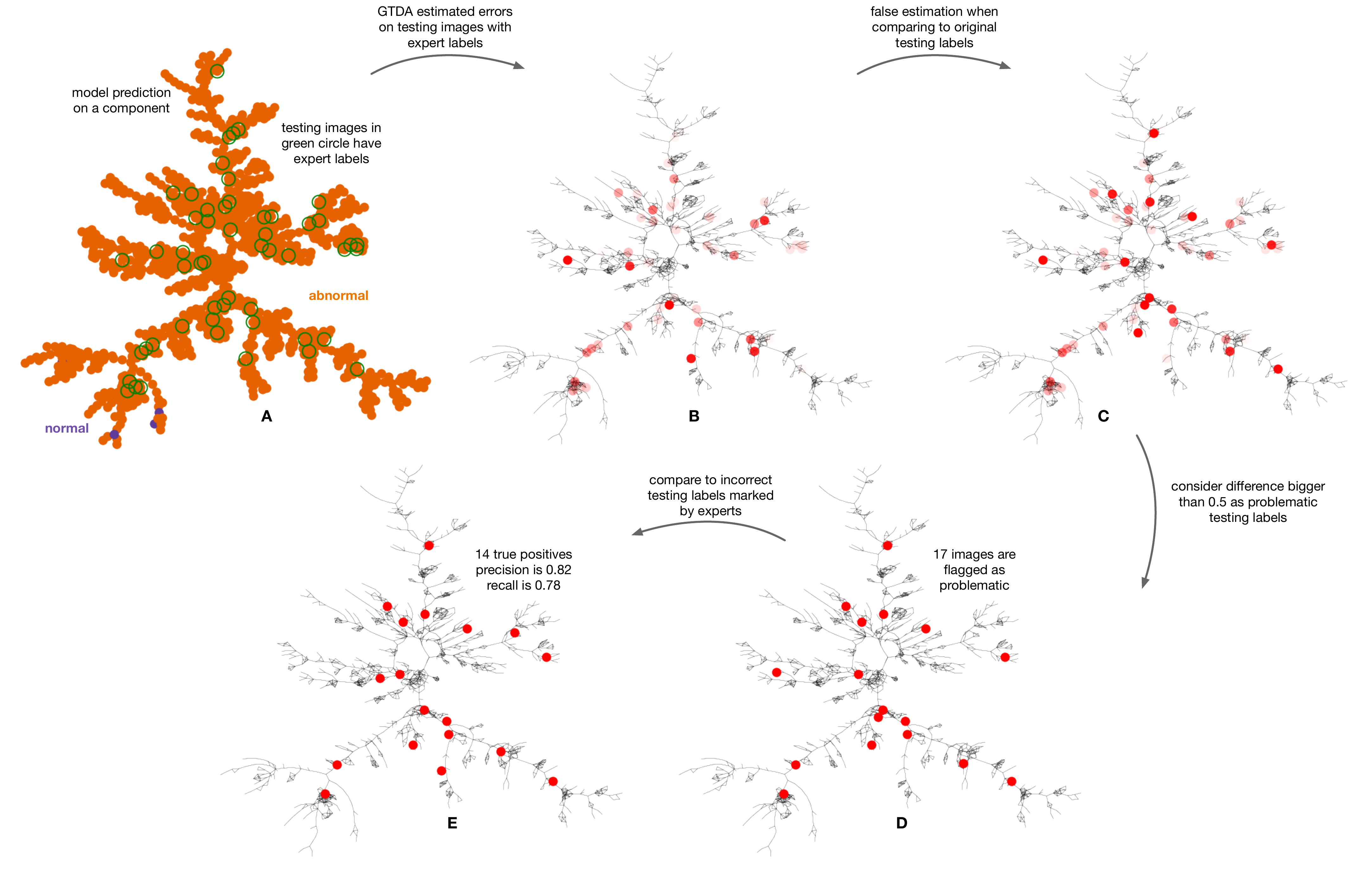}
\caption{We give a demonstration on how to use GTDA results to find which testing labels are likely to be problematic. We first zoom in a component found by GTDA and use green circles to mark testing images where we have expert labels (A). Then we use GTDA to estimate prediction errors on circled images (B). Comparing GTDA estimation with original testing labels can identify a few places with false estimations (C). We consider these false estimations are due to problematic testing labels and do a simple thresholding of 0.5, which flags 17 problematic testing labels in this component (D). Comparing to expert labels can find 14 true positives with a precision of 0.82 and a recall of 0.78 (E).}
\label{chest_xray_demo}
\end{figure}

For simplicity, images in the test set where such differences are larger than 0.5 are considered to have incorrect labels. A demonstration on this process can be found in \Cref{chest_xray_demo}. Overall, out of the 810 testing images with expert labels, GTDA highlights 265 images are likely to have incorrect normal vs abnormal labels and 138 of them are confirmed by the expert labels, which gives a precision of 0.52 and a recall of 0.62. As a comparison, randomly sampling 265 images for experts to check can only find around 73 images with incorrect labels in average. More detailed results on each component are shown in \Cref{tab:chest}. By testing multiple thresholds instead of 0.5, we get an AUC score of 0.75. As a comparison, using self confidence~\cite{northcutt2021confidentlearning} gives an overall AUC score of 0.60.
\input{figures/chest_xray/stats.tex}

%% file: figures/chest_xray/stats.tex
\begin{table}[h!]
\centering
\footnotesize
\begin{tabular}{cccccc}
\toprule
Type&\shortstack{Expert Labels\\ in Component}&\shortstack{Incorrect \\by Experts}&\shortstack{Flagged as \\Problematic}&Precision&Recall\\
\midrule
Single Component&53&18&17&0.82&0.78\\
Single Component&10&5&5&1.0&1.0\\
Single Component&9&5&4&0.25&0.2\\
Single Component&19&4&7&0.57&1.0\\
Single Component&9&4&5&0.8&1.0\\
Single Component&10&4&3&0.33&0.25\\
Single Component&7&4&2&1.0&0.5\\
Single Component&8&4&5&0.6&0.75\\
Single Component&14&4&4&1.0&1.0\\
Single Component&4&4&2&1.0&0.5\\
Single Component&7&4&3&0.33&0.25\\
Single Component&10&3&2&0.0&0.0\\
Single Component&6&3&1&0.0&0.0\\
Single Component&4&3&2&0.5&0.33\\
Single Component&6&3&3&0.33&0.33\\
Single Component&3&3&2&1.0&0.67\\
Single Component&5&3&3&1.0&1.0\\
Single Component&5&3&2&0.5&0.33\\
Single Component&8&3&5&0.4&0.67\\
Single Component&7&3&4&0.5&0.67\\
Single Component&19&3&8&0.25&0.67\\
Single Component&9&3&8&0.38&1.0\\
Single Component&8&3&3&0.33&0.33\\
Single Component&8&3&4&0.5&0.67\\
\midrule 
Components with 2 incorrect labels&135&56&50&0.74&0.66\\
Components with 1 incorrect label&219&67&78&0.5&0.58\\
Components with 0 incorrect label&208&0&33&0.0&NaN\\
Overall&810&222&265&0.52&0.62\\
\bottomrule
\end{tabular}
\caption{Detailed precision and recall on different components when using GTDA to find likely incorrect testing labels of ChestX-ray14 dataset. Components are ordered by decreasing number of incorrect labels identified by experts they contain. Results for components with less than 3 incorrect labels are reported together.}
\label{tab:chest}
\end{table}

%% file: parameters.tex
\section{Parameter selection of GTDA}\label{sec:params}
\label{sec:sensitivity}
In this section, we will discuss how to select parameters for our GTDA framework, especially the component size threshold and overlapping ratio in \Cref{GTDA}. Currently, we manually focus the Reeb net's structure by varying these parameters.  It remains an open question on how one might automatically select parameters for our GTDA framework as proposed for other TDA frameworks~\cite{carriere2018statistical}. Although GTDA has 8 parameters (Table~\ref{tab:params}), the two most important are the component size threshold and the overlapping ratio. 

\subsection{Selecting component size threshold} 
\begin{figure}[tp]
    \centering
    \includegraphics[width=\linewidth]{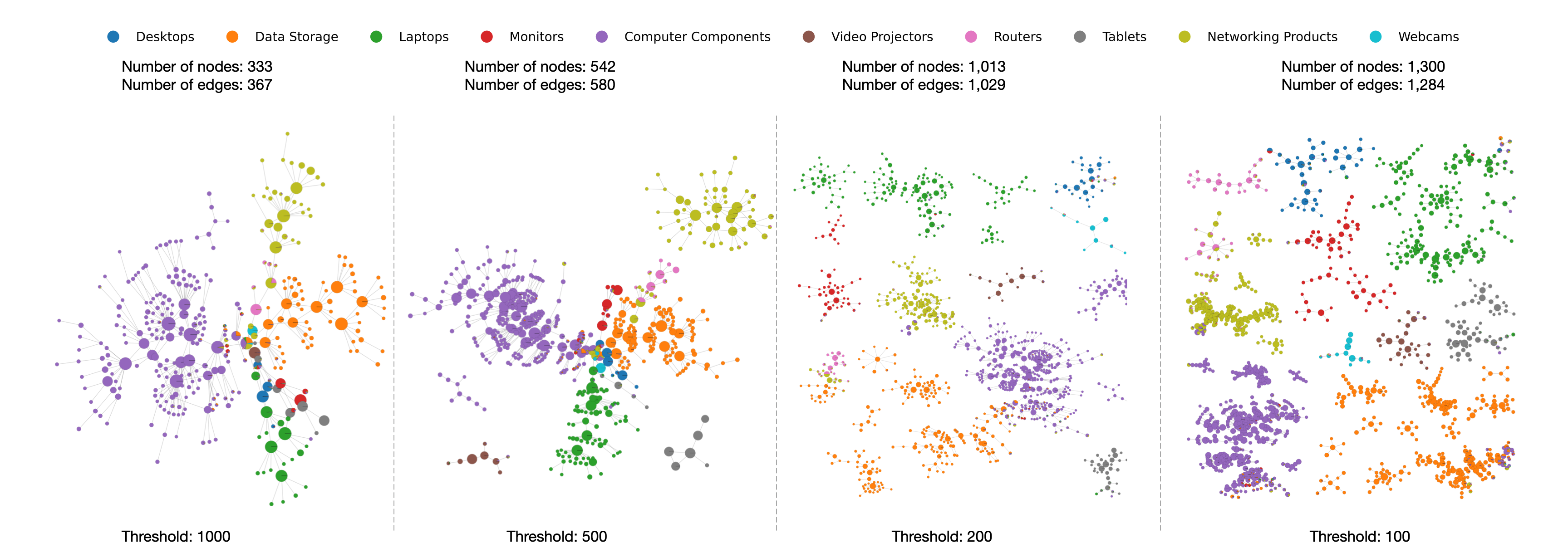}
\caption{We show different GTDA visualizations as we vary the component size threshold. The overlapping ratio is fixed as 1\%. Using a large threshold will cause different classes to be mixed together and the structure of small class like ``Routers'' or ``Webcams'' will be over simplified. As we gradually reduce the thresholds, the number of nodes and edges in the visualization will increase as well and different classes will be separated into several components. The results look similar between 100 and 200, which suggests GTDA structure are stable with respect to small change in parameters.}
\label{vary_size_thd}
\end{figure}

Recall that the component size threshold is the smallest component where we stop splitting.  Choosing a good component size threshold depends on the dataset we want to analyze. If the threshold is too small, we will end up with too many nodes to make the subsequent visualization and analysis difficult. On the other hand, if the threshold is too large, the topological structure of some small classes might be over simplified and components from different classes can be mixed. \Cref{vary_size_thd} shows how the Reeb net will change as we vary component size threshold. In general, we start from a larger component size threshold and then check the Reeb net we get, especially the size of the largest Reeb net component as well as whether different classes are mixed. If we have a component that is too large to be easily visualized or different classes are clearly mixed, we reduce this value. If the class sizes are highly skewed, we usually choose the threshold based on the smallest class. In this case, we the lower bound on the absolute difference of the lens parameter becomes useful. This is to avoid oversplitting class with large size, i.e., if the difference is smaller than the lower bound, we stop splitting as well. 

We find the results are stable to the choices and in particular, for uses to find clues of possible predicting errors or labeling issues from the visualization. As we can see in \Cref{vary_size_thd}, choosing a threshold between 100 and 200 or choosing an overlapping ratio between 0.5\% and 1.5\% can all show the ambiguity in ``Networking Products'' v.s.~``Routers'' and some part of ``Data Storage'' v.s.~``Computer Components'' allowing human insight into the predictions. 

\subsection{Select overlapping ratio} 
\begin{figure}[tp]
    \centering
    \includegraphics[width=\linewidth]{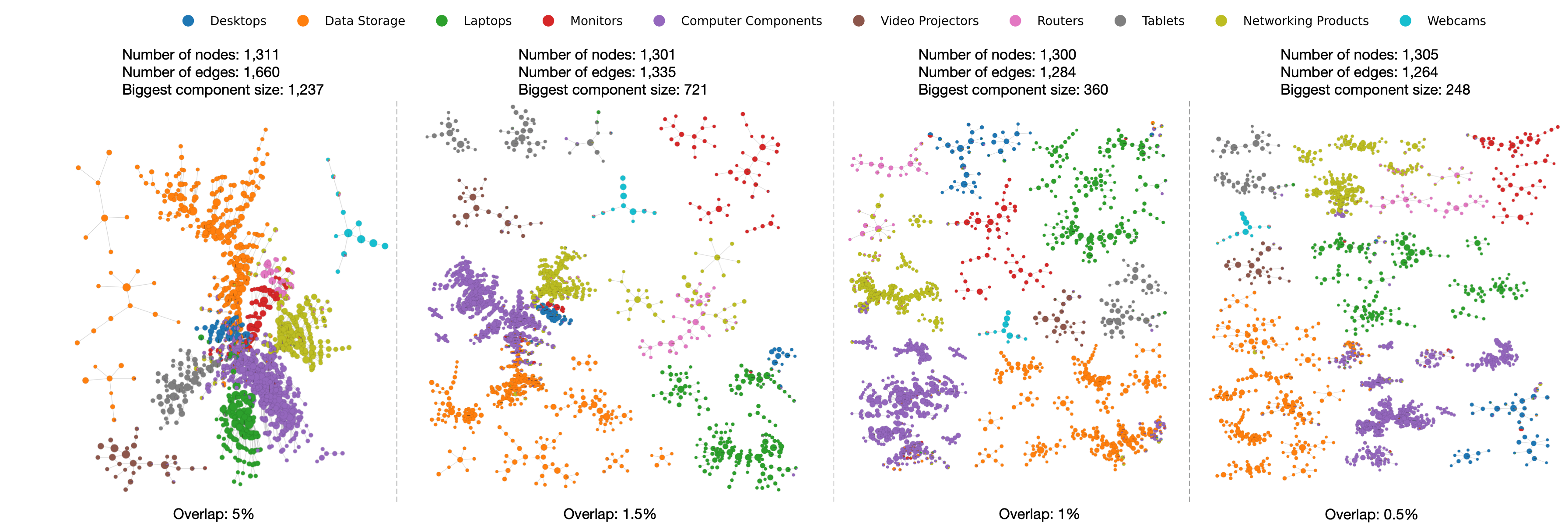}
\caption{We show different GTDA visualizations as we vary the overlapping ratio. The component size threshold is fixed as 100. Using a large overlapping ratio will cause different classes to be mixed together and some components too large to be properly visualized. As we gradually reduce the overlapping ratio, different classes will be separated into several components with each one easier to be plotted. Similar ambiguity in ``Networking Products'' v.s. ``Routers'' and some part of ``Data Storage'' v.s. ``Computer Components'' can be observed for overlapping ratio between 0.5\% and 1.5\%.}
\label{vary_overlap}
\end{figure}

The selection of overlapping ratio is similar to select component size threshold, we can start from a larger ratio like 10\% and then check the Reeb net to see if there is any component that is too large or too mixed. If so, we need to gradually reduce the ratio until every component can be properly visualized by a simple layout algorithm like spring layout~\cite{tutte1963draw} or Kamada Kawai algorithm~\cite{kamada1989algorithm}. \Cref{vary_overlap} shows different Reeb nets as we vary overlapping ratio.

\subsection{Notes on other parameters}
Other than component size threshold and overlapping ratio, \Cref{GTDA} has several other parameters. Two important ones are the smallest node size and the smallest component size. In our experiments, we can get consistently good visualizations by requiring the size of any Reeb net node or Reeb net component larger than 5. 

%% file: scaling.tex
\section{Performance and scaling}

Our GTDA framework scales to predictions with thousands of classes and millions of datapoints. We only split along the lens with the maximum difference at each iteration, which can be easily recomputed in linear time in the data or even more efficiently updated. After each split, we immediately check all the connected components we have found, which can be done in $O(N+M)$ where $N$ is the number of nodes and $M$ is the number of edges. 

It is difficult to estimate how many splitting iterations are needed.
Assuming we have $L$ lenses, initially the min-max difference across all lenses is 1 and the overlapping ratio is 0, then we will need at most $L$ iterations before the largest min-max difference across all components is reduced to 0.5, which means at most $O(tL)$ iterations are needed to reduce such difference below $2^{-t}$. If after a sufficient number of iterations, we still see large components with size bigger than $K$, it means new lenses are needed to further distinguish those nodes or a lower bound on the difference is needed to stop the splitting early.

Another step is to find out which pairs of components have overlap. This can be easily done in the original \emph{mapper} algorithm by checking the adjacent bins of each bin. In our GTDA framework, we first build a bipartite graph with all component indices on one side, all samples on the other side and connecting each component index to all samples it includes. Then identifying the overlapping components is equivalent to find 2-hop neighbors of each component index, which can also be done in $O(N+M)$. Finally, for the merging step, since the size of each super node or the size of Reeb net component will be at least doubled, it needs at most $O(M(\text{log}(s_1s_2)))$ time. Also note that, many steps of our GTDA algorithm can be easily parallelized. In our code, we mainly parallelize the merging steps using 10 cores, which has already given reasonable running time on graphs with millions of nodes and edges. 

Detailed running time for all datasets we have tested can be found in the \cref{tab:running_time}. All running time are reported on a server with 2 AMD EPYC 7532 processors (128 cores in total), 512 GB memory and one A100 GPU.

\begin{table}[h!]
\scriptsize
\centering
\begin{tabular}{@{}cccccccc@{}}
\toprule
dataset & nodes & edges & classes & lens & \makecell{predicting \& \\ embedding (s)} & preprocessing (s) & GTDA time (s) \\

\midrule
\arrayrulecolor{lightgray}
Swiss Roll & 1,000 & 3,501 & 3 & 3 & 0.003 & 0.3 & 1\\
\midrule 
Amazon Computers & 39,747 & 399,410 & 10 & 10 & 0.17 & 7 & 10\\
\midrule 
Subset of ImageNet & 13,394 & 51,520 & 10 & 10 & 27 & 5 & 7\\
\midrule 
\makecell{ImageNet-1k \\ \footnotesize (ResNet vs~AlexNet)} & 1,331,167 & 5,954,900 & 1,000 & 2,000 & 2,379 & 717 & 26,036\\
\midrule 
\makecell{ImageNet-1k \\ \footnotesize (VOLO vs~ResNet)} & 1,331,167 & 5,805,714 & 1,000 & 2,000 & 13,426 & 617 & 18,894\\
\midrule 
BRCA1 Gene Variants & 23,376 & 83,096 & 2 & 4 & 18,583 & 21 & 3\\
\midrule 
Chest X-rays & 112,120 & 431,893 & 2 & 16 & 821 & 35 & 26\\
\arrayrulecolor{black}
\bottomrule
\end{tabular}
\caption{Statistics on datasets and running time in seconds. Predicting and embedding represents the time used to generate prediction and extract embedding for all samples from a trained model. Preprocessing time includes PCA, normalization as well as building a KNN graph if the original dataset is not in graph format. GTDA time is the time to compute Reeb network given the input graph and the lens.}
\label{tab:running_time}
\end{table}

%% file: exp_tsne_umap.tex
\section{Comparing to tSNE and UMAP}
\label{sec:tsne-umap}
\begin{figure}[tp]
    \centering
    \includegraphics[width=\linewidth]{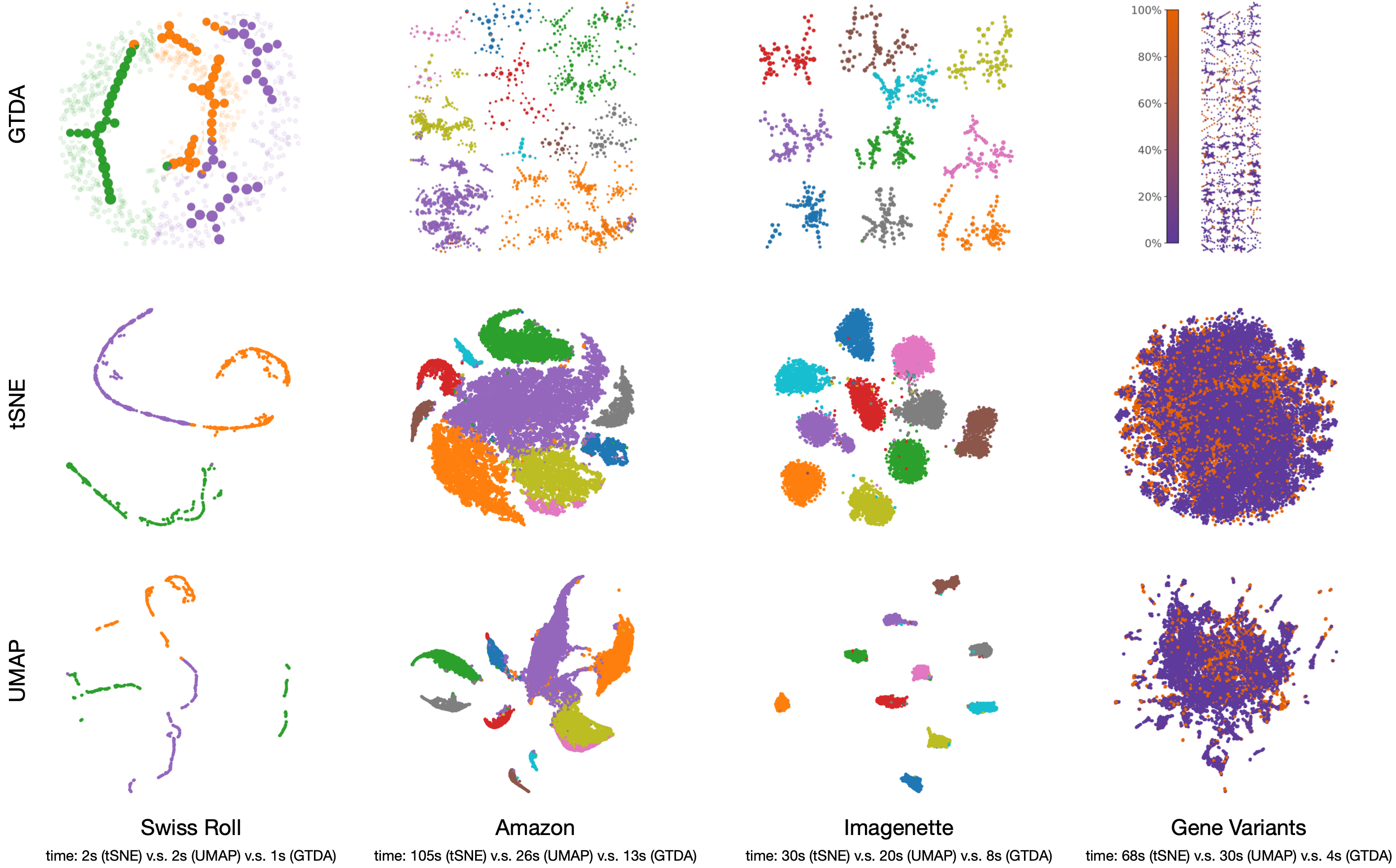}
\caption{Comparing the results of the dimension reduction techniques tSNE and UMAP on 4 datasets to the topological Reeb net structure from GTDA shows similarities and differences among summary pictures generated by these methods.  The graph created by GTDA permits many types of analysis not clearly possible with tSNE and UMAP output. 
For running time comparison, since we also need to extract model embeddings and predictions just like GTDA, we exclude such time and only report the time of the actual execution of tSNE or UMAP or GTDA (including Kamada-Kawai).}
\label{fig:tsne_umap}
\end{figure}

The goals of the Reeb net analysis from GTDA are distinct from the goals of dimension reduction techniques such as tSNE and UMAP. We seek the topological information identified by the Reeb net. The Reeb net is both useful for generating pictures or maps of the data as well as the algorithmic error estimate. We use the Kamada-Kawai~\cite{kamada1989algorithm} method to compute a visualization of the Reeb net, which does have many similarities with summary pictures from tSNE and UMAP. 
We compare here GTDA results with visualization from tSNE~\cite{tSNE} and UMAP~\cite{Becht2018,umap} on all 4 datasets of the main text. For tSNE, we directly use the implementation from \emph{scikit-learn}. For UMAP, we use the implementation from \url{https://umap-learn.readthedocs.io}. The inputs to tSNE and UMAP are the concatenation of neural model embedding and prediction probability. We keep all parameters as default except setting the number of final dimension as 2. The results are shown in \Cref{fig:tsne_umap}. 

These pictures support different uses and purposes. Reeb nets from GTDA offer a number of compelling advantages as described throughout the main text and supplement. Among others, note that GTDA is faster than tSNE (2 to 15 times faster) and UMAP (2 to 8 times faster) in all 4 datasets. It also scales easily to datasets with millions of datapoints.
